\documentclass{article}

\usepackage[final]{corl_2026} % camera-ready ``final'' version（显示真实作者与单位、首页底部会议 notice）。
\makeatletter\renewcommand{\@conferencelocation}{Austin, Texas, USA}\makeatother
\usepackage{amsmath,amssymb}
\usepackage{graphicx}
\usepackage{wrapfig}
\usepackage{booktabs}
\usepackage{multirow}
\usepackage{xcolor}
\definecolor{mydarkblue}{rgb}{0,0.08,0.45}
\usepackage{colortbl}
\usepackage{subcaption}
\usepackage{algorithm}
\usepackage{algorithmic}
\usepackage{bm}

\usepackage{caption}
\newcommand{\ours}{G2G}

\newcommand{\SE}{\mathrm{SE}(3)}
\newcommand{\SO}{\mathrm{SO}(3)}
\newcommand{\R}{\mathbb{R}}
\newcommand{\pose}[2]{T_{#1 \leftarrow #2}}
\definecolor{cBest}{HTML}{0091AD}
\definecolor{cSecond}{HTML}{D2791E}
\newcommand{\best}[1]{\textcolor{cBest}{#1}}
\newcommand{\second}[1]{\textcolor{cSecond}{#1}}
\makeatletter
\AtBeginDocument{\@setfontsize\normalsize\@xpt\@xipt\normalsize}
\makeatother

\makeatletter
\newcommand{\firstpagenotes}[1]{%
  \begingroup
    \renewcommand{\thefootnote}{}%
    \long\def\@makefntext##1{\parindent\z@ \parskip 2\p@ \noindent ##1}%
    \ifdefined\H@@footnotetext \let\@footnotetext\H@@footnotetext \fi
    \footnotetext{#1}%
  \endgroup
}
\makeatother

\title{G2G: Exploiting Intra-Group Geometry\\for Inter-Group Pose Estimation}

\author{
  Yufei Wei$^{*\,1}$ \quad
  Shuhao Ye$^{*\,1}$ \quad
  Chenxiao Hu$^{1}$ \quad
  Yiyuan Pan$^{2}$ \quad
  Dongyu Feng$^{2}$ \\
  \bfseries Rong Xiong$^{1}$ \quad
  Yue Wang$^{1}$ \quad
  Yanmei Jiao$^{\dagger\,3}$ \\[4pt]
  \mdseries\small
  $^{1}$\,State Key Laboratory of Industrial Control Technology, Zhejiang University, Hangzhou, China \\
  \mdseries\small
  $^{2}$\,Zhejiang Humanoid Robot Innovation Center Co., Ltd., Ningbo, China \\
  \mdseries\small
  $^{3}$\,School of Information Science and Engineering, Hangzhou Normal University, Hangzhou, China \\
}

\begin{document}
\hypersetup{pdftitle={G2G: Exploiting Intra-Group Geometry for Inter-Group Pose Estimation},pdfauthor={Yufei Wei, Shuhao Ye, Chenxiao Hu, Yiyuan Pan, Dongyu Feng, Rong Xiong, Yue Wang, Yanmei Jiao}}
\maketitle
\firstpagenotes{%
  $^{*}$ Co-first authors. Yufei Wei formulated the Group-to-Group problem and led the project, carrying out the method design, implementation, data pipelines, experiments, and writing; Shuhao Ye ran the LoMa baseline and processed the ZJH ground truth. Full author contributions are listed in Appendix~\ref{app:contributions}.\par
  $^{\dagger}$ Corresponding author.%
}

% ==============================================================================
\begin{abstract}
% ==============================================================================
% Abstract (included by main.tex inside \begin{abstract}...\end{abstract})
% ==============================================================================

Recovering the relative 6-DoF pose between two image groups underlies cross-sequence relocalization and multi-camera rig odometry.
Each group carries known intra-group geometry from visual odometry or rig calibration, and pretrained multi-view backbones already fuse such geometry into visual features.
Yet current models treat all views as an unstructured set, leaving cross-group reasoning as the missing piece.
We introduce \ours{}, which keeps the foundation model entirely frozen and adds three lightweight trainable modules to bridge the two groups: a perceiver resampler, a cross-group bridge with merged self-attention, and a multi-frame pose head.
The trainable footprint totals about 32M parameters, under 6\% of the full model, and is supervised only by relative poses.
Across four datasets that span indoor and outdoor simulation, real-world cross-season capture, and zero-shot sim-to-real transfer, \ours{} attains state-of-the-art accuracy on both tasks, while trainable baselines are retrained with their original supervision.
Code and visualizations: \url{https://weiyufei0217.github.io/G2G/}.

\end{abstract}


% ==============================================================================
% ==============================================================================
% Section 1: Introduction
% ==============================================================================
\section{Introduction}
\label{sec:introduction}

Recovering the relative six-degree-of-freedom pose between two groups of images is a recurring task in robotic perception.
It underlies cross-sequence visual relocalization and multi-camera rig odometry, and the same primitive also supports loop closure in visual SLAM~\cite{campos2021orbslam3} and multi-session mapping for heterogeneous robot teams.
These applications share the same input structure: two image groups with known intra-group geometry, and an unknown rigid transformation between the two groups that has to be recovered.
% It underlies cross-sequence visual relocalization 要加参考文献 and multi-camera rig odometry 要加参考文献
% heterogeneous robot teams 要加参考文献

We refer to this primitive as \emph{Group-to-Group} (\ours{}) pose estimation.
The two groups arise from two complementary sources.
\emph{Temporal groups} are monocular sequence segments captured along different traversals of the same environment.
Their intra-group geometry is produced by visual odometry, visual SLAM, or a previously built SfM map.
\emph{Spatial groups} are simultaneous observations of a multi-camera rig.
Their intra-group geometry is given by the calibrated camera-to-body transforms.
In both cases the intra-group geometry is known but not noise-free, since odometry tends to drift along a trajectory and rig calibrations can degrade over time.
The two cases look superficially different, yet they reduce to the same input and output interface, which suggests that a single model can serve a broad class of downstream applications.
% Temporal groups 和 Spatial groups 的参考文献也要找一找

Classical pipelines based on local feature matching followed by essential-matrix recovery~\cite{detone2018superpoint, sarlin2020superglue, wang2024efficientloftr, edstedt2024roma, nordstrom2026loma} aggregate pairwise estimates into a group-level pose, but they often fail catastrophically in the presence of dynamic objects, seasonal variation, or scene changes between captures. They also rely on multiple inference passes and iterative optimization to converge.
Feed-forward multi-view models provide an alternative by learning geometric representations jointly across views.
DUSt3R~\cite{wang2024dust3r}, MASt3R~\cite{leroy2024mast3r}, and VGGT~\cite{wang2025vggt} follow this paradigm, yet they treat the inputs as an unstructured collection, encoding neither which images belong to the same group nor the intra-group geometry that the application can readily supply.
MapAnything~\cite{keetha2025mapanything} accepts optional camera priors expressed in a common reference frame. Conditioning it jointly on two independently posed groups would require the very inter-group transform that we seek to estimate.
This intra-group geometry is intrinsic to the applications above rather than imposed by our formulation, yet it is left largely unexploited by existing methods.

Exploiting this geometry does not require retraining the foundation model.
A backbone pretrained to fuse intrinsics and extrinsics within a single image group already grounds visual tokens in camera geometry.
The missing piece is cross-group reasoning.
We propose \ours{}, a method that keeps a pretrained multi-view foundation model entirely frozen and adds three lightweight modules on top to bridge the two groups.
The frozen backbone processes each group independently and produces geometry-enhanced patch features through its native fusion of camera pose and ray information.
A perceiver resampler then compresses these features into a small number of latent tokens per frame.
A cross-group bridge concatenates the latent tokens from both groups, augments them with frame, group, and anchor embeddings, and refines them through merged self-attention.
A pose head finally predicts every non-anchor frame's pose with respect to a shared reference frame in a single forward pass.
\ours{} is trained with relative poses as the only supervision signal. Depth maps, when available, are used during dataset construction to score visual overlap and select training pairs.
This freeze-and-bridge design preserves the pretrained knowledge of the foundation model, enables data-efficient training, and facilitates transfer from simulation to the real world.

Our main \textbf{contributions} are as follows.
(1) We introduce \emph{Group-to-Group} pose estimation as a unified problem. Cross-sequence relocalization and multi-camera rig odometry, which have so far been studied with separate pipelines, are cast as the same task and addressed by a shared architecture.
(2) We propose a freeze-and-bridge architecture. A frozen multi-view foundation model carries the intra-group geometric reasoning, while three lightweight trainable modules supply the missing cross-group reasoning. The trainable footprint adds approximately $32$M parameters, under six percent of the full system.
(3) We report state-of-the-art accuracy on four datasets and two downstream tasks while using only relative-pose supervision, whereas trainable baselines are retrained with their full original supervision, including dense or sparse depth where applicable.

% ==============================================================================
% ==============================================================================
% Section 2: Related Work
% ==============================================================================
\section{Related Work}
\label{sec:related_work}

\paragraph{Classical pipelines for cross-group geometry estimation.}
Classical approaches estimate cross-group geometry through multi-stage pipelines composed of feature extraction, matching, geometric verification, and iterative optimization.
In the cross-sequence setting, local features are extracted from each image and matched across views, after which the relative pose is recovered via essential-matrix decomposition or PnP with RANSAC.
Traditional systems employ handcrafted descriptors for this purpose, while recent learned matchers such as ELoFTR~\cite{wang2024efficientloftr} and RoMa~\cite{edstedt2025romav2} have substantially advanced matching quality and efficiency under large viewpoint and appearance changes.
For multi-camera rigs, the generalized camera model~\cite{pless2003using} enables motion estimation across non-overlapping fields of view~\cite{kazik2012real, heng2015self}. At a larger scale, multi-session and multi-robot SLAM~\cite{campos2021orbslam3, tian2022kimeramulti, lajoie2020door} and collaborative localization systems~\cite{jiao2026opennavmap, peterson2024roman} align independently constructed maps through place recognition and optimization.
All such approaches depend on feature-match quality or map-reconstruction fidelity and degrade under limited visual overlap.
Feed-forward multi-view models offer an alternative by reasoning over all frames jointly in a single pass.

\paragraph{Multi-view feed-forward foundation models.}
DUSt3R~\cite{wang2024dust3r} and MASt3R~\cite{leroy2024mast3r} established the feed-forward paradigm for multi-view 3D reconstruction by regressing dense 3D pointmaps from image pairs in a single forward pass, but both require $O(N^2)$ pairwise inference followed by iterative global alignment.
Subsequent work moves beyond pairwise processing toward multi-view architectures~\cite{yang2025fast3r, cabon2025must3r, liu2025slam3r} or scales to full SfM pipelines~\cite{duisterhof2025mastrsfm, elflein2025light3r}.
VGGT~\cite{wang2025vggt} predicts depth, camera poses, point maps, and point tracks through dedicated heads in a single alternating-attention transformer, finding that composing independently estimated depths and cameras yields more accurate 3D points than the point-map head alone.
These models treat all input views as an unstructured collection.
When multiple views come from the same capture session or share a calibrated rig, the known intra-group geometry is not exploited.
Several recent methods have begun to address this gap by incorporating geometric priors into the feed-forward pipeline.

\paragraph{Incorporating known geometry into feed-forward models.}
Reloc3R~\cite{dong2025reloc3r} specializes a DUSt3R-style backbone for pairwise relative-pose regression and recovers metric-scale absolute poses through motion averaging over multiple posed database images.
MapAnything~\cite{keetha2025mapanything} moves geometry into the backbone by accepting optional ray directions, camera poses, and depth as encoder-level conditioning.
A dedicated scale token resolves the metric-scale ambiguity that Reloc3R defers to a separate motion-averaging stage, and flexible input augmentation enables a single model to handle over twelve 3D reconstruction tasks.
Rig3R~\cite{li2026rig3r} introduces optional rig metadata, including camera identifiers, timestamps, and rig-relative raymaps, with metadata dropout for robustness to missing information.
Its dual-raymap output separates ego-motion from rig-internal structure and supports rig discovery from unordered images.
Despite these advances, all three methods operate on a single image collection or a database-query pipeline.
None directly outputs the rigid transform between two arbitrary image groups with known intra-group geometry.
\ours{} addresses this gap by keeping the foundation model entirely frozen and adding three lightweight trainable modules to bridge the two groups.
These modules total under six percent of the model parameters and predict the inter-group $\SE$ transform in a single forward pass.

% ==============================================================================
% ==============================================================================
% Section 3: Method
% ==============================================================================
\section{Method}
\label{sec:method}

We introduce the Group-to-Group formulation and present \ours{}, which keeps a pretrained multi-view foundation model entirely frozen and bridges the two groups with three lightweight trainable modules.
The foundation model fuses intra-group geometry into the visual features of each group.
The trainable modules then reason across the two groups and jointly predict every target frame's pose relative to a shared anchor in a single forward pass.
\ours{} takes RGB images with their intrinsics and the known intra-group extrinsics as input, and uses relative poses as the only supervision signal.

% ------------------------------------------------------------------------------
\subsection{Problem Formulation}
\label{sec:problem}

Consider two observation groups $\mathcal{G}_A = \{(\mathbf{I}_a^i,\, \mathbf{K}_a^i)\}_{i=0}^{N_A-1}$ and $\mathcal{G}_B = \{(\mathbf{I}_b^j,\, \mathbf{K}_b^j)\}_{j=0}^{N_B-1}$, each containing RGB images $\mathbf{I}$ paired with camera intrinsics $\mathbf{K}$.
The frame counts $N_A$ and $N_B$ may differ.
Each group is accompanied by a set of known intra-group extrinsics
\begin{equation}
  \mathbf{E}_A = \{\pose{A_0}{A_i}\}_{i=0}^{N_A-1}, \qquad
  \mathbf{E}_B = \{\pose{B_0}{B_j}\}_{j=0}^{N_B-1},
\end{equation}
expressed relative to each group's anchor frame.
We designate $A_0$ as the global reference and call the remaining frames the \emph{target frames}.
\ours{} realizes the mapping
\begin{equation}
  \ours{} \,:\; \big( (\mathcal{G}_A, \mathbf{E}_A),\; (\mathcal{G}_B, \mathbf{E}_B) \big)
            \;\longmapsto\; \big\{\pose{A_0}{A_i}\big\}_{i=1}^{N_A-1}
                            \cup \big\{\pose{A_0}{B_j}\big\}_{j=0}^{N_B-1}
                            \subset \SE,
  \label{eq:mapping}
\end{equation}
producing every target-frame pose in a single forward pass, from which any pairwise relative pose follows by composition.
The inter-group outputs $\{\pose{A_0}{B_j}\}$ align the two groups, while the intra-group outputs $\{\pose{A_0}{A_i}\}$ correct residual noise in $\mathbf{E}_A$ from odometry drift or miscalibration.

The formulation above subsumes the two instantiations motivated in Sec.~\ref{sec:introduction}, which differ only in how $\mathbf{E}$ is acquired.
Temporal groups obtain $\mathbf{E}$ from visual odometry, visual SLAM, or a pre-built SfM map.
Spatial groups read $\mathbf{E}$ from the calibrated camera-to-body transforms of a multi-camera rig.
Both cases use the same architecture and pose convention in Eq.~\eqref{eq:mapping}; for rig odometry, we initialize from the corresponding relocalization model and adapt it to spatial groups (Appendix~\ref{app:impl:training}).

\paragraph{Selecting observation groups.}
For relocalization, each group is a short temporal window; for rig odometry, each group contains simultaneous camera views at one timestamp.
When temporal windows are not supplied, an optional RGB covisibility predictor selects overlapping windows from longer sequences before pose estimation (Appendix~\ref{app:stage1}).
This separates finding a suitable observation pair from estimating its relative pose; the latter uses the known geometry within each selected group.

% ------------------------------------------------------------------------------
\subsection{Architecture}
\label{sec:architecture}

\begin{figure*}[t]
  \centering
  \includegraphics[width=\textwidth]{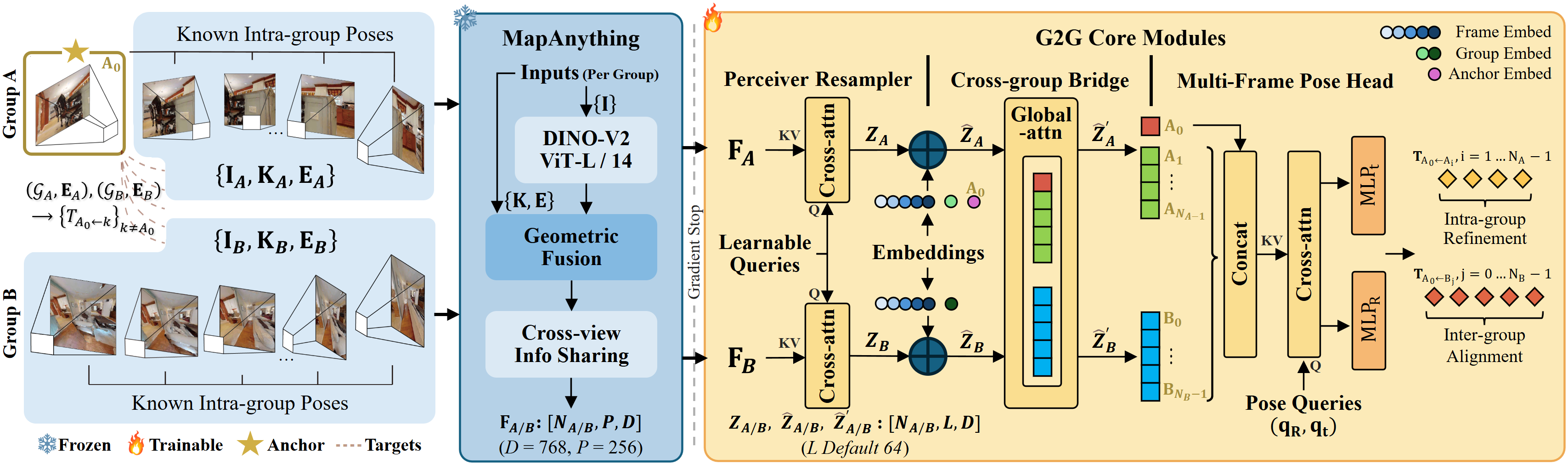}
  \vspace{-0.4cm}
  \caption{\textbf{\ours{} architecture.}
  A frozen MapAnything backbone independently encodes each group into geometry-enhanced patch features. Three lightweight trainable modules then reason across the two groups and regress every target pose relative to the anchor $A_0$ in a single forward pass.}
  \vspace{-0.4cm}
  \label{fig:pipeline}
\end{figure*}

\ours{} consists of a frozen multi-view foundation model and three lightweight trainable modules: a perceiver resampler, a cross-group bridge, and a multi-frame pose head, as illustrated in Fig.~\ref{fig:pipeline}.
The frozen backbone accounts for approximately 539M parameters, while the trainable modules total approximately 32M, under 6\% of the full model (Appendix~\ref{app:impl:params}).
A gradient stop separates the backbone from the trainable modules, so the backbone is never updated during training.

\paragraph{Foundation model.}
We adopt MapAnything~\cite{keetha2025mapanything} as a per-group feature extractor.
For each frame, the inputs to the backbone are the RGB image, the camera extrinsic decomposed into a quaternion and a translation, and per-pixel ray directions computed from the intrinsic $\mathbf{K}$.
Each group is encoded in its own local frame, with its anchor at identity.
This lets the backbone use both sets of intra-group extrinsics without knowing the transform between their coordinate systems; the bridge subsequently estimates that relationship.
Inside the backbone, a DINOv2 ViT-L/14 encoder~\cite{oquab2024dinov2} extracts patch tokens from each frame.
MapAnything's native geometric fusion then injects pose and ray encodings into the patch tokens, and an alternating-attention transformer~\cite{wang2025vggt} propagates information across views within the same group.
The output for each group is a tensor of geometry-enhanced patch features
\begin{equation}
  \mathbf{F}_g \in \R^{N_g \times P \times D}, \qquad g \in \{A, B\},
\end{equation}
where $P$ is the number of patches per frame and $D$ is the feature dimension.
Camera extrinsics enter the model exclusively at this stage.
The downstream trainable modules receive no extrinsic input, since the intra-group geometry is already encoded in $\mathbf{F}_g$.

\paragraph{Perceiver resampler.}
A perceiver resampler compresses the $P$ patch tokens per frame into a smaller set of $L$ latent tokens.
It consists of $L$ learnable query tokens and two cross-attention layers, and it processes every frame independently:
\begin{equation}
  \mathbf{Z}_g \in \R^{N_g \times L \times D}, \qquad g \in \{A, B\}.
\end{equation}
The compression reduces the sequence length entering the cross-group bridge; without it, the bridge would have to attend over a much longer concatenated sequence.
We use $L{=}64$; Appendix~\ref{app:ablation:efficiency} quantifies the accuracy and memory trade-off against retaining all $P{=}256$ patch tokens.

\paragraph{Cross-group bridge.}
The bridge applies standard self-attention to the concatenated tokens of both groups, allowing each token to exchange information within and across groups.
Three types of learnable embeddings organize the merged token sequence.
A \textit{frame embedding} $\mathbf{e}_\text{frame}$ encodes each frame's position within its group, corresponding to the frame index for temporal groups and the camera index for spatial groups.
It is shared between groups $A$ and $B$, since the positional semantics are analogous in either case.
After flattening each group's tokens, the two groups are concatenated into a single sequence.
A \textit{group embedding} $\mathbf{e}_\text{group}$ distinguishes $A$ tokens from $B$ tokens in the merged sequence.
An \textit{anchor embedding} $\mathbf{e}_\text{anchor}$ marks the $A_0$ tokens as the global reference frame.
Concretely, each latent token is augmented as
\begin{equation}
  \hat{\mathbf{Z}}_g^{\,i} \;=\; \mathbf{Z}_g^{\,i} \;+\; \mathbf{e}_\text{frame}^{\,i} \;+\; \mathbf{e}_\text{group}^{\,g} \;+\; \mathbb{1}[g{=}A,\, i{=}0]\,\mathbf{e}_\text{anchor},
  \label{eq:bridge_token}
\end{equation}
where the three embeddings broadcast across the $L$ latent tokens of frame $i$ in group $g$, and the indicator $\mathbb{1}[\cdot]$ activates the anchor term only on the $A_0$ tokens.
Two self-attention layers then transform the merged sequence $[\hat{\mathbf{Z}}_A;\,\hat{\mathbf{Z}}_B]$ into the bridged tokens $[\hat{\mathbf{Z}}'_A;\,\hat{\mathbf{Z}}'_B]$, so that cross-group exchange and intra-group context sharing emerge from the same unified attention.
These bridged tokens are split back into per-group, per-frame tokens for the pose head.
Appendix~\ref{app:ablation:varwin} evaluates a model trained with variable group sizes, including inference beyond its training range.

\paragraph{Pose head.}
The pose head predicts every target frame's pose relative to $A_0$ through a single cross-attention layer whose weights are shared across all target frames.
Two learnable pose queries $\mathbf{q}_R, \mathbf{q}_t \in \R^{1 \times D}$, one for rotation and one for translation, are also common to every target frame.
For each target frame $k$, a learnable identity embedding $\mathbf{e}_k$ is added to its bridged tokens $\hat{\mathbf{Z}}'_k$ to specify which pose is being predicted.
The key-value input is the concatenation $[\hat{\mathbf{Z}}'_{A_0};\, \hat{\mathbf{Z}}'_k + \mathbf{e}_k]$ along the token axis, where $\hat{\mathbf{Z}}'_{A_0}$ denotes the anchor's tokens after the bridge.
The pose queries cross-attend to this input to produce a rotation feature $\mathbf{f}_R^k$ and a translation feature $\mathbf{f}_t^k$. Two separate MLPs decode these features into a 6D continuous rotation~\cite{zhou2019continuity} and a 3D translation. The 6D output is then converted to a rotation matrix via Gram-Schmidt orthogonalization.
All target frames are batched into a single forward pass, producing the full prediction set in Eq.~\eqref{eq:mapping}.

\paragraph{Training objective.}
The pose estimator is supervised only by relative poses. For a target frame $k$, we combine a chordal rotation loss with an $\ell_1$ translation loss:
\begin{equation}
 \ell_k = 5d_R(\widehat{\mathbf R}_k,\mathbf R_k)
          +\tfrac{1}{3}\|\widehat{\mathbf t}_k-\mathbf t_k\|_1, \qquad
 \mathcal L = \frac{0.5}{N_A-1}\sum_{i=1}^{N_A-1}\ell_{A_i}+\frac{1}{N_B}\sum_{j=0}^{N_B-1}\ell_{B_j}.
 \label{eq:training_loss}
\end{equation}
Here hats denote predictions, unhatted quantities denote ground truth, and $d_R$ is the normalized chordal distance specified in Appendix~\ref{app:impl:loss}.
The larger weight on inter-group predictions prioritizes alignment, while intra-group supervision trains the model to refine noisy input geometry.
During training, we perturb input extrinsics while keeping the supervision poses clean.
A three-phase curriculum first samples high-overlap pairs, then progressively introduces lower-overlap pairs before learning-rate decay.
Depth is used offline to score overlap and construct training pairs, but is neither an input to the pose estimator nor a supervision target.
Appendices~\ref{app:impl} and~\ref{app:data:pipeline} give the training schedule, noise distribution, and pair-construction procedure.

% ==============================================================================
% ==============================================================================
% Section 4: Experiments
% ==============================================================================
\vspace{-0.1cm}
\section{Experiments}
\label{sec:experiments}

% ------------------------------------------------------------------------------
\vspace{-0.1cm}
\subsection{Experimental Setup}
\label{sec:setup}

\textbf{Tasks.}
We train and evaluate \ours{} on each of four datasets for the two downstream tasks of Sec.~\ref{sec:method}, comparing against classical and learned baselines for each task.
Task~1 is cross-sequence relocalization, and Task~2 is multi-camera rig odometry.

\textbf{Datasets.}
The four datasets together cover indoor simulation, outdoor simulation, real-world cross-time capture, and sim-to-real transfer.
HM3D~\cite{ramakrishnan2021habitat} provides 800 training and 100 validation indoor scenes rendered from Habitat-Matterport scans.
TartanGround~\cite{patel2025tartanground} adds 53 training and 10 validation outdoor ground-robot scenes.
NCLT~\cite{carlevaris2016university} consists of real campus traversals from a 5-camera Segway platform, with 8 training and 2 validation dates spread across seasons, so evaluation pairs span seasonal and illumination change.
Our self-collected ZJH dataset provides 22 training and 3 test indoor environments reconstructed with Gaussian Splatting, together with 3 real 4-camera rig sequences held out for zero-shot sim-to-real evaluation.
Appendix~\ref{app:data} details the splits, camera configurations, and training-pair construction.

\textbf{Baselines.}
For Task~1 we compare against LoMa~\cite{nordstrom2026loma}, CoViS-Net~\cite{blumenkamp2024covisnet}, VGGT~\cite{wang2025vggt}, Reloc3R~\cite{dong2025reloc3r}, and two MapAnything variants: MA-A is given only group~$A$'s extrinsics, and the oracle MA-AB additionally receives group~$B$'s extrinsics through the ground-truth inter-group transform; Task~2 swaps LoMa and CoViS-Net for the rig-aware Rig3R~\cite{li2026rig3r}.
Trainable baselines are retrained on each target dataset with their full original supervision (including dense or sparse depth where applicable), while \ours{} is supervised by relative poses alone; Appendix~\ref{app:baselines} details each input protocol, evaluation pipeline, and the pretrained-versus-finetuned comparison (Appendix~\ref{app:baselines:pretrained}). Within each setting, all methods receive the same image groups; camera priors enter according to each method's supported interface.

\textbf{Metrics and protocol.}
We report mean translation error $t$ in meters and mean rotation error $r$ in degrees.
Mean errors retain the effect of large failures. We also report translation-direction angular error (RTA), which is independent of translation scale, for NCLT and ZJH in Table~\ref{tab:task1}.
Appendix~\ref{app:metrics} provides median errors, threshold accuracies, and scale-free mAA to characterize the error distribution more fully.
We apply no pair-wise scale alignment to any method, with VGGT in Table~\ref{tab:task2} as the only exception: VGGT predicts up-to-scale geometry and does not support metric-scale extrinsic injection that would otherwise anchor the scale, so its raw $t$ is not directly comparable to the metric-scale baselines.
Since the main rig table has no column for RTA, we $\mathrm{Sim}(3)$-align VGGT's predictions per pair in Table~\ref{tab:task2} to keep its translation column comparable.
Table~\ref{tab:task1} still lists VGGT's raw $t$, and we judge VGGT by rotation and RTA throughout.
The main tables use clean ground-truth intra-group extrinsics. Appendix~\ref{app:ablation:noise} evaluates noisy extrinsics, and Appendix~\ref{app:metrics:overlap} breaks down accuracy by visual overlap.

% ------------------------------------------------------------------------------
\vspace{-0.1cm}
\subsection{Cross-Sequence Relocalization}
\label{sec:task1}

% ----- Tables 1 & 2 (auto-generated, merged into one float to keep them on the same page) -----
\begin{table*}[!t]
% 由 generate_paper_table1_latex.py 自动生成，请勿手动编辑数值
% 注意：此片段不含 \begin{table*}/\end{table*} 外壳，必须被外层 table* 环境包裹后再 \input。
\centering
\caption{\textbf{Cross-sequence relocalization on four datasets} (mean errors).
  \textcolor{cBest}{Cyan}/\textcolor{cSecond}{orange} mark the best/second-best non-oracle result; ZJH~{\scriptsize(S/R)} reports sim/real per cell.}
\label{tab:task1}
\vspace{-0.1cm}
\resizebox{\textwidth}{!}{%
\begin{tabular}{lcccccccccccc}
\toprule
  & \multicolumn{2}{c}{\textbf{HM3D}} & \multicolumn{2}{c}{\textbf{TartanGround}} & \multicolumn{4}{c}{\textbf{NCLT}} & \multicolumn{4}{c}{\textbf{ZJH {\scriptsize(S/R)}}} \\
\cmidrule(lr){2-3} \cmidrule(lr){4-5} \cmidrule(lr){6-9} \cmidrule(lr){10-13}
Method & $t${\scriptsize(m)}$\downarrow$ & $r${\scriptsize($^\circ$)}$\downarrow$ & $t${\scriptsize(m)}$\downarrow$ & $r${\scriptsize($^\circ$)}$\downarrow$ & $t${\scriptsize(m)}$\downarrow$ & RTA{\scriptsize($^\circ$)}$\downarrow$ & $r${\scriptsize($^\circ$)}$\downarrow$ & RRA{\scriptsize @5$^\circ$}$\uparrow$ & $t${\scriptsize(m)}$\downarrow$ & RTA{\scriptsize($^\circ$)}$\downarrow$ & $r${\scriptsize($^\circ$)}$\downarrow$ & RRA{\scriptsize @5$^\circ$}$\uparrow$ \\
\midrule
LoMa & 0.802 & 3.79 & \second{1.183} & 18.40 & 7.858 & 72.39 & 22.13 & 55.1 & 2.946/3.443 & 38.13/31.08 & 19.32/\second{5.75} & 65.4/\second{86.1} \\
CoViS-Net$^\dagger$ & 1.511 & 37.24 & 3.025 & 36.18 & \second{5.731} & \second{42.97} & 17.22 & 37.6 & 2.134/3.413 & 37.93/40.49 & 32.61/59.35 & 12.8/6.4 \\
VGGT$^\ddagger$ & 1.500 & 8.18 & 3.006 & 16.06 & 8.775 & 82.09 & 43.87 & 18.4 & 3.047/4.497 & 21.98/17.65 & 13.09/13.73 & 72.8/76.3 \\
Reloc3R$^\ddagger$ & \second{0.725} & \second{2.66} & 1.649 & 14.88 & 6.918 & 48.47 & \second{6.34} & \second{70.3} & 1.701/2.988 & 26.34/29.14 & \second{4.18}/7.75 & \second{89.9}/85.0 \\
\midrule
MA-A & 1.008 & 10.08 & 1.793 & \second{13.58} & 7.021 & 44.61 & 31.69 & 25.9 & \second{0.556}/\second{1.324} & \best{10.60}/\second{9.63} & 4.68/9.02 & 79.9/78.4 \\
\rowcolor{gray!8}
MA-AB~{\scriptsize(oracle)} & 0.087 & 1.00 & 0.081 & 0.59 & 0.208 & 0.89 & 0.82 & 99.9 & 0.061/0.066 & 4.87/1.17 & 0.89/0.65 & 100.0/100.0 \\
\midrule
\textbf{\ours{}~(Ours)} & \best{0.155} & \best{1.72} & \best{0.526} & \best{5.39} & \best{0.692} & \best{5.43} & \best{2.47} & \best{95.2} & \best{0.305}/\best{1.206} & \second{11.16}/\best{7.01} & \best{1.67}/\best{3.46} & \best{96.5}/\best{95.0} \\
\bottomrule
\end{tabular}%
}

\vspace{2pt}
\parbox{\textwidth}{\footnotesize\raggedright%
$^\dagger$\,retrained from scratch on target data; $^\ddagger$\,finetuned from pretrained checkpoints.}
\vspace{-6pt}
\vspace{0.6em}
% 强制 hyperref 给第二个 caption 独立 anchor，避免 \ref{tab:task2} 跳到 Table 1。
\begingroup
\renewcommand{\theHtable}{\thetable.merged}
% 由 generate_paper_table2_latex.py 自动生成，请勿手动编辑数值
% 注意：此片段不含 \begin{table*}/\end{table*} 外壳，必须被外层 table* 环境包裹后再 \input。
\centering
\caption{\textbf{Multi-camera rig odometry on six configurations} (mean errors).
  \textcolor{cBest}{Cyan}/\textcolor{cSecond}{orange} mark the best/second-best non-oracle; NCLT$_{\text{in/cr}}$ are intra-/cross-date rigs, ZJH~{\scriptsize(S/R)} reports sim/real per cell.}
\label{tab:task2}
\vspace{-0.1cm}
\resizebox{\textwidth}{!}{%
\begin{tabular}{lcccccccccccc}
\toprule
  & \multicolumn{2}{c}{\textbf{HM3D-8}} & \multicolumn{2}{c}{\textbf{HM3D-4}} & \multicolumn{2}{c}{\textbf{TartanGround}} & \multicolumn{2}{c}{\textbf{NCLT$_{\text{in}}$}} & \multicolumn{2}{c}{\textbf{NCLT$_{\text{cr}}$}} & \multicolumn{2}{c}{\textbf{ZJH {\scriptsize(S/R)}}} \\
\cmidrule(lr){2-3} \cmidrule(lr){4-5} \cmidrule(lr){6-7} \cmidrule(lr){8-9} \cmidrule(lr){10-11} \cmidrule(lr){12-13}
Method & $t${\scriptsize(m)}$\downarrow$ & $r${\scriptsize($^\circ$)}$\downarrow$ & $t${\scriptsize(m)}$\downarrow$ & $r${\scriptsize($^\circ$)}$\downarrow$ & $t${\scriptsize(m)}$\downarrow$ & $r${\scriptsize($^\circ$)}$\downarrow$ & $t${\scriptsize(m)}$\downarrow$ & $r${\scriptsize($^\circ$)}$\downarrow$ & $t${\scriptsize(m)}$\downarrow$ & $r${\scriptsize($^\circ$)}$\downarrow$ & $t${\scriptsize(m)}$\downarrow$ & $r${\scriptsize($^\circ$)}$\downarrow$ \\
\midrule
Rig3R$^\dagger$ & 0.635 & 11.16 & 0.421 & \second{5.54} & 12.617 & 37.86 & 1.573 & 4.71 & 1.851 & 6.14 & 1.599/2.504 & 24.36/34.57 \\
VGGT\textsuperscript{\ddag\,\S} & \second{0.589} & 17.71 & \second{0.228} & 12.50 & \second{0.560} & 10.67 & \second{0.191} & 3.29 & 1.438 & 35.09 & \second{0.273}/\best{0.423} & 8.15/10.31 \\
Reloc3R$^\ddagger$ & 0.984 & \second{11.10} & 0.586 & 8.05 & 1.129 & \second{5.09} & 0.688 & \second{3.08} & \second{0.752} & \second{4.47} & 0.735/0.963 & 6.89/8.71 \\
\midrule
MA-A & 1.432 & 20.34 & 0.931 & 15.55 & 1.607 & 8.91 & 0.486 & 4.20 & 1.707 & 13.02 & 0.391/0.707 & \second{3.26}/\second{7.00} \\
\rowcolor{gray!8}
MA-AB~{\scriptsize(oracle)} & 0.068 & 1.16 & 0.061 & 1.26 & 0.064 & 0.61 & 0.060 & 0.75 & 0.063 & 1.06 & 0.031/0.048 & 0.66/0.70 \\
\midrule
\textbf{\ours{}~(Ours)} & \best{0.402} & \best{7.37} & \best{0.218} & \best{4.48} & \best{0.311} & \best{1.63} & \best{0.078} & \best{1.02} & \best{0.172} & \best{1.97} & \best{0.131}/\second{0.516} & \best{1.35}/\best{2.82} \\
\bottomrule
\end{tabular}%
}

\vspace{2pt}
\parbox{\textwidth}{\footnotesize\raggedright%
$^\dagger$\,reproduced and trained with DUSt3R initialization; $^\ddagger$\,finetuned from pretrained checkpoints; \textsuperscript{\S}\,translation reported after per-pair optimal $\mathrm{Sim}(3)$ alignment to ground truth.}
\vspace{-2pt}

\vspace{-1.6em}
\endgroup
\end{table*}

Table~\ref{tab:task1} compares \ours{} with classical and learned baselines across the four datasets.
\ours{} achieves the best non-oracle translation and rotation on every dataset.
On HM3D and TartanGround, the two simulated splits, it reaches $0.16$\,m / $1.72^\circ$ and $0.53$\,m / $5.39^\circ$, both ahead of the strongest learned baseline Reloc3R at $0.72$\,m / $2.66^\circ$ and $1.65$\,m / $14.88^\circ$.
% The MA-AB oracle, which receives both groups' extrinsics in the same anchor frame, reaches $0.09$\,m and $0.08$\,m on these two splits, so the gap from \ours{} to MA-AB quantifies the inter-group pose that the method estimates rather than receives.
On ZJH \ours{} stays the strongest non-oracle method on both the simulated and real columns even though the trainable modules see only the simulated split during training: RRA@$5^\circ$ holds at $96.5\%$ on simulation and $95.0\%$ on the real captures, and mean rotation rises only mildly from $1.67^\circ$ to $3.46^\circ$.

NCLT exposes the role of intra-group geometry most clearly.
The dataset consists of repeated traversals of the same campus collected across many months, so two groups frequently align the same place under different seasons, illumination, and minor scene changes such as added or removed structures.
\ours{} keeps RTA at $5.43^\circ$, RRA@$5^\circ$ at $95.2\%$, and a mean translation of $0.69$\,m, while every baseline degrades to RTA above $40^\circ$ with translations of several meters.
The baselines differ in how they use the available geometry: VGGT estimates geometry from images alone, and Reloc3R introduces intra-group extrinsics only when aggregating pairwise poses after inference, so in both cases the cross-group estimate rests on image-to-image correspondence that seasonal change disrupts.
\ours{} instead conditions each group's features on its known geometry before reasoning across groups, so the alignment does not hinge on such correspondence alone.
Appendix~\ref{app:baselines:vggt} examines the cross-season setting in more detail; the controlled attention ablation in Sec.~\ref{sec:ablation} tests the role of cross-group exchange while retaining the geometric inputs.

Beyond accuracy, the single-pass formulation also brings an efficiency advantage.
LoMa and Reloc3R recover the group-to-group pose by aggregating all pairwise estimates between the two groups, twenty-five of them for two groups of five frames, while \ours{} produces every target pose in one forward pass. Including the frozen backbone, inference from raw images takes $49.79$\,ms on an RTX~4090 for $5{+}5$ frames, compared with $461.55$\,ms for Reloc3R; Appendix~\ref{app:ablation:efficiency} gives the measurement protocol and memory costs.

% ------------------------------------------------------------------------------
\vspace{-0.1cm}
\subsection{Multi-Camera Rig Odometry}
\label{sec:task2}

% Fig.2/3 在此声明（4.3 开头，第 7 页顶部附近）：此处页面剩余空间足够，图会浮到第 7 页顶部；
% 若声明在 4.4 开头（第 7 页 60% 处），剩余空间不够会被推到第 8 页。图在 4.4 中引用。
\begin{figure}[t]
  \centering
  \vspace{-0.3cm}
  \begin{minipage}[t]{0.385\textwidth}
    \centering
    \includegraphics[width=\linewidth]{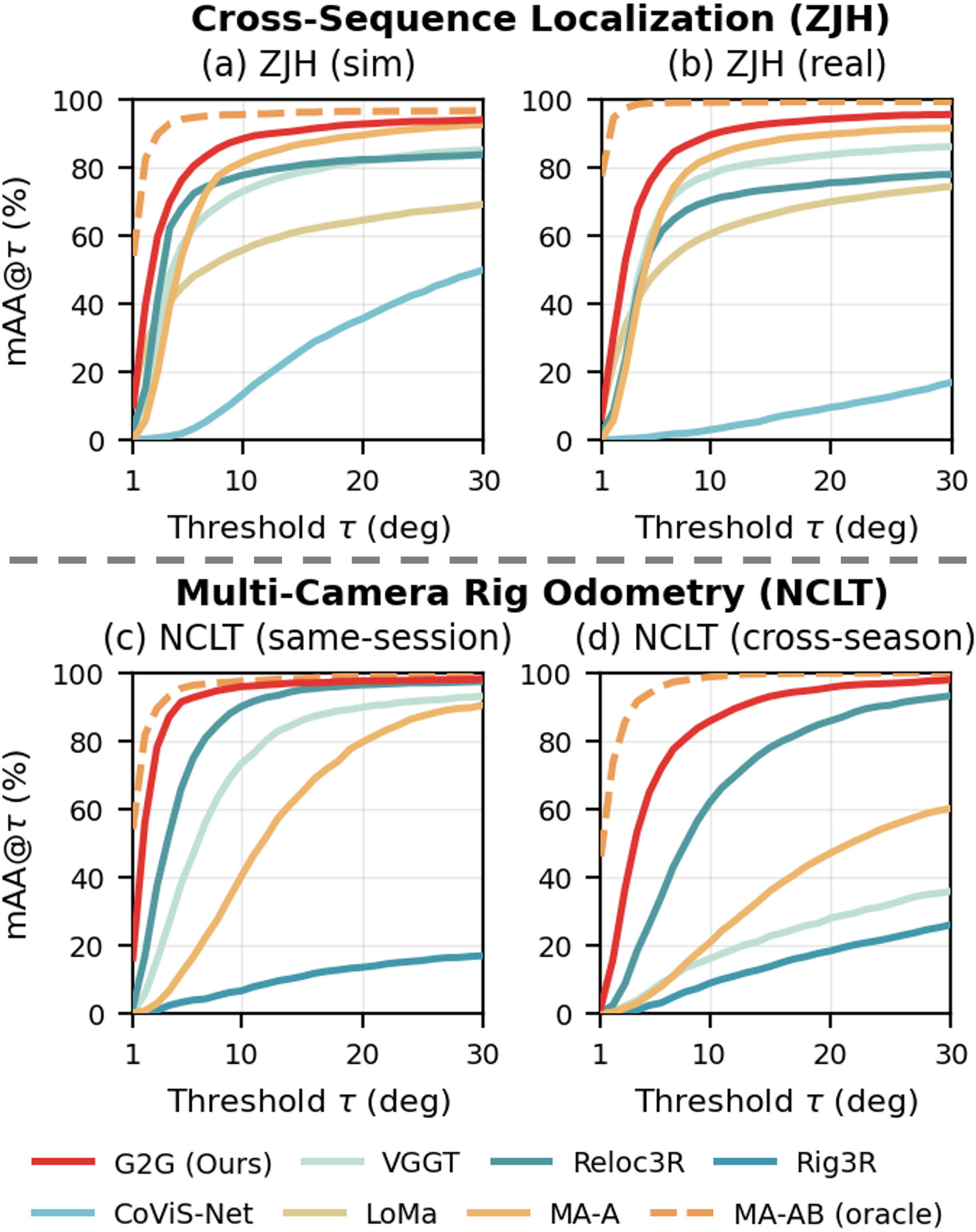}
    \caption{\textbf{Accuracy under increasing precision.}
    Each curve plots mAA@$\tau$ as the threshold $\tau$ sweeps from $1^\circ$ to $30^\circ$. A pair is correct when its rotation and scale-free translation-direction errors are both below $\tau$.}
    \label{fig:accuracy}
  \end{minipage}\hfill
  \begin{minipage}[t]{0.595\textwidth}
    \centering
    \includegraphics[width=\linewidth]{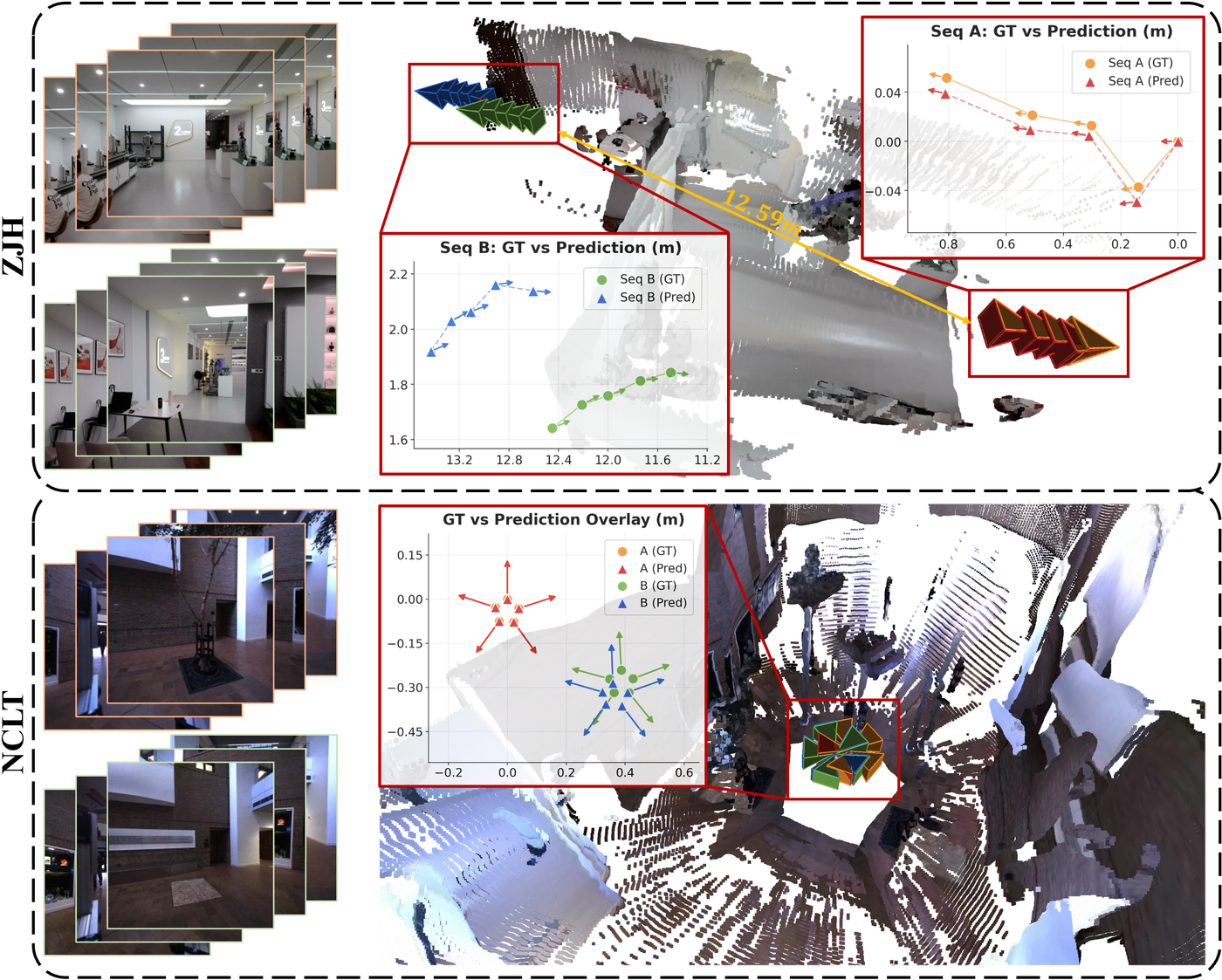}
    \caption{\textbf{Qualitative examples on real data.}
    (Top) A zero-shot sim-to-real ZJH localization pair in which the two groups image opposite sides of the same wall, leaving little direct visual overlap.
    (Bottom) A long-range cross-session NCLT rig pair captured in different seasons. Predicted poses overlay the ground-truth poses in the insets.}
    \label{fig:casestudy}
  \end{minipage}
  \vspace{-0.5cm}
\end{figure}

% ----- Table 2 (auto-generated) -----
Table~\ref{tab:task2} compares \ours{} with three rig-aware or multi-view baselines across six rig configurations.
\ours{} achieves the best non-oracle rotation and the best metric-scale translation on every configuration.
On the HM3D rigs it reaches $0.40$\,m / $7.37^\circ$ with eight cameras (inter-camera spacing $0$ to $3$\,m) and $0.22$\,m / $4.48^\circ$ with a random four-camera subset ($0$ to $1.5$\,m), against next-best learned baselines at $0.59$\,m / $11.10^\circ$ and $0.23$\,m / $5.54^\circ$.
On the NCLT same-session rig it improves further to $0.08$\,m / $1.02^\circ$, and on the cross-date rig it holds $0.17$\,m / $1.97^\circ$, while the next-best baselines reach only $0.19$\,m / $3.08^\circ$ and $0.75$\,m / $4.47^\circ$.
On ZJH the same lead carries over to both the simulated and real columns.
The simulated rig yields $0.13$\,m / $1.35^\circ$, and on the real rig, evaluated zero-shot from simulation, the model still produces the best rotation at $2.82^\circ$; its metric-scale translation of $0.52$\,m approaches VGGT's $0.42$\,m even though VGGT's column is reported after a per-pair optimal $\mathrm{Sim}(3)$ alignment to the ground truth.

Rig3R exhibits a larger generalization gap on TartanGround, where training covers $53$ scenes and evaluation uses $10$ disjoint scenes.
Its training and validation behavior is analyzed in Appendix~\ref{app:baselines:rig3r}.
\ours{} preserves the pretrained backbone and learns only the cross-group modules, retaining a common geometric representation across these scenes.
On the NCLT cross-date rig, \ours{} maintains a rotation error of $1.97^\circ$, compared with VGGT's $35.1^\circ$; the per-pair scale alignment applied to VGGT does not affect this rotation metric.

The pairwise evaluation above measures local motion accuracy.
Table~\ref{tab:controls} (right) additionally composes consecutive predictions over two held-out NCLT traversals of about $6$\,km each, with no loop closure or global bundle adjustment, and reports KITTI-style segment errors.
\ours{} reaches $6.1\%$ RTE and $3.5^\circ/100$\,m RRE, against $24.3\%$ and $11.0^\circ/100$\,m for the best non-oracle baseline Reloc3R, while the monocular systems VGGT-SLAM~\cite{maggio2025vggtslam}, LingBot-Map~\cite{chen2026lingbotmap}, and CUT3R~\cite{wang2025cut3r}, which build on geometry foundation models but consume raw images only, remain far less accurate.
Even the MA-AB oracle stays at $9.2\%$ after a ground-truth-assisted correction of its systematic $0.6^\circ$ yaw bias, and MA-A, which shares our frozen backbone, stops at $34.1\%$, so the gap is not explained by the backbone.
Appendix~\ref{app:odometry} gives the input, scale, and bias-correction protocols.

% ------------------------------------------------------------------------------
\vspace{-0.1cm}
\subsection{Robustness to Sim-to-Real and Seasonal Shifts}
\label{sec:transfer}
% \subsection{Case Study}
% \label{sec:case}

% \begin{wrapfigure}{R}{0.45\textwidth}
%   \centering
%   \vspace{-0.3cm}
%   \includegraphics[width=\linewidth]{figures/ferror_distribution_2.png}
%   \vspace{-0.2cm}
%   \caption{\textbf{Accuracy under increasing precision.}
%   Each curve plots mAA@$\tau$ as the threshold $\tau$ sweeps from $1^\circ$ to $30^\circ$. A pair counts as correct when its rotation error and its scale-free translation-direction error are both below $\tau$, so curves closer to the top-left corner are better.}
%   % Panels (a, b) cover cross-sequence localization on ZJH under simulation and real capture. Panels (c, d) cover multi-camera rig odometry on NCLT for same-session and cross-season pairs.
%   % \ours{} stays closest to the MA-AB oracle in every panel.
%   \vspace{-0.3cm}
%   \label{fig:accuracy}
% \end{wrapfigure}

% \begin{wrapfigure}{R}{0.5\textwidth}
%   \centering
%   \vspace{-0.3cm}
%   \includegraphics[width=\linewidth]{figures/case-study.png}
%   \vspace{-0.2cm}
%   \caption{\textbf{Qualitative examples on real data.}
%   (Top) A zero-shot sim-to-real ZJH localization pair in which the two groups image opposite sides of the same wall, leaving little direct visual overlap.
%   (Bottom) A long-range cross-session NCLT rig pair captured in different seasons.}
%   % Predicted poses overlay the ground-truth trajectories in the insets, and \ours{} aligns the bottom pair to within $1.12^\circ$ and $5.3$\,cm.
%   \vspace{-0.3cm}
%   \label{fig:casestudy}
% \end{wrapfigure}

Two settings stress generalization: ZJH transfers a sim-trained model to a real 4-camera rig, and the NCLT pairs cover the same campus across seasons.
On both, \ours{} stays the best non-oracle method on localization and the best metric-scale method on rig odometry, while the retrained baselines lose ground.
MA-A reuses the same frozen backbone but does not condition on both groups' local geometry. The attention ablation in Sec.~\ref{sec:ablation} further examines the contribution of cross-group reasoning.

Figure~\ref{fig:accuracy} reports the corresponding mAA@$\tau$ curves on both settings.
On ZJH the \ours{} curve barely moves from the simulated split to the real one, with mAA staying around $84$ on both panels, while CoViS-Net collapses to $7.0$ and the other learned baselines each lose several points.
On NCLT the same-session pairs are handled well by most methods, but the cross-season panel pulls them apart sharply: VGGT falls from $72.2$ to $21.0$ mAA and Reloc3R follows a similar trend, whereas \ours{} stays above $82$ and tracks the MA-AB oracle.
The \ours{} curve also rises steeply at small $\tau$ rather than catching up only at loose thresholds, so most of its predictions land in the high-precision regime rather than just in the correct direction.

% Table 3 在此声明：本段落起始于第 8 页，声明时页顶仍有余量，可与 Fig.2/3 同页置顶，不再被推到第 9 页。
% 由 generate_paper_table3_latex.py 自动生成，请勿手动编辑数值
% 数据源: aggregate_reloc_tables.build_ablation_sources + collect_results (clean, mean)
\begin{table*}[t]
\centering
\caption{\textbf{Ablation studies of \ours{} on five evaluation settings} (mean errors).
  Each column pair ablates one design choice; \textcolor{cBest}{cyan}/\textcolor{cSecond}{orange} mark the best/second-best $t$ and $r$ per row.}
\label{tab:ablation}
\vspace{-0.1cm}
\resizebox{\textwidth}{!}{%
\begin{tabular}{l>{\columncolor{gray!8}}cc>{\columncolor{gray!8}}cc>{\columncolor{gray!8}}cc>{\columncolor{gray!8}}cc>{\columncolor{gray!8}}cc>{\columncolor{gray!8}}cc}
\toprule
  & \multicolumn{2}{c}{\textbf{Default}} & \multicolumn{2}{c}{\textbf{w/o Resamp.}} & \multicolumn{2}{c}{\textbf{w/o Anchor}} & \multicolumn{2}{c}{\textbf{w/o Curric.}} & \multicolumn{2}{c}{\textbf{w/o Noise Aug.}} & \multicolumn{2}{c}{\textbf{$3{+}3$ Win.}} \\
\cmidrule(lr){2-3} \cmidrule(lr){4-5} \cmidrule(lr){6-7} \cmidrule(lr){8-9} \cmidrule(lr){10-11} \cmidrule(lr){12-13}
Setting & $t${\scriptsize(m)}$\downarrow$ & $r${\scriptsize($^\circ$)}$\downarrow$ & $t${\scriptsize(m)}$\downarrow$ & $r${\scriptsize($^\circ$)}$\downarrow$ & $t${\scriptsize(m)}$\downarrow$ & $r${\scriptsize($^\circ$)}$\downarrow$ & $t${\scriptsize(m)}$\downarrow$ & $r${\scriptsize($^\circ$)}$\downarrow$ & $t${\scriptsize(m)}$\downarrow$ & $r${\scriptsize($^\circ$)}$\downarrow$ & $t${\scriptsize(m)}$\downarrow$ & $r${\scriptsize($^\circ$)}$\downarrow$ \\
\midrule
HM3D & \second{0.155} & \second{1.72} & \best{0.123} & \best{1.47} & 0.171 & 1.98 & 0.171 & 1.83 & 0.177 & 2.03 & 0.244 & 3.05 \\
TartanGround & \second{0.526} & \second{5.39} & \best{0.441} & \best{4.46} & 0.618 & 5.92 & 0.541 & 5.93 & 0.531 & 5.72 & 0.711 & 10.31 \\
NCLT & \second{0.692} & 2.47 & \best{0.597} & \best{2.29} & 0.793 & 2.64 & 0.742 & 2.50 & 0.744 & \second{2.40} & 0.900 & 3.05 \\
ZJH {\scriptsize(sim)} & \second{0.305} & \best{1.67} & \best{0.249} & \second{1.73} & 0.369 & 2.19 & 0.352 & 1.74 & 0.428 & 2.79 & 0.434 & 2.22 \\
ZJH {\scriptsize(real)} & 1.206 & 3.46 & \second{1.118} & \best{2.81} & 1.122 & \second{3.17} & 1.143 & 3.79 & \best{1.020} & 4.51 & 1.170 & 4.54 \\
\bottomrule
\end{tabular}%
}
\vspace{-0.4cm}
\end{table*}

Figure~\ref{fig:casestudy} shows individual pairs, with additional examples across all four datasets in Appendix~\ref{app:qualitative}.
The top row contains a ZJH localization pair in which the two groups image the front and the back of the same wall, leaving almost no shared visual content.
Although \ours{} has never seen real data during training, the inset trajectories show its predicted poses tracking the ground truth on both groups.
The bottom row contains a long-range cross-session NCLT rig pair captured in different seasons, where surface appearance changes substantially between the two visits.
% \ours{} still brings the two groups into one frame, with a residual inter-group rotation of $1.12^\circ$ and a residual translation of $5.3$\,cm.

% ------------------------------------------------------------------------------
\vspace{-0.1cm}
\subsection{Ablation Studies}
\label{sec:ablation}

Table~\ref{tab:ablation} evaluates the anchor embedding, curriculum, extrinsic-noise augmentation, group size, and resampler across four datasets (five settings including both ZJH splits).
Removing the anchor embedding increases both errors on all simulated settings and NCLT, supporting its role in identifying the common reference frame.
The curriculum and noise augmentation generally improve accuracy, although the real ZJH split shows different translation and rotation trade-offs; Appendix~\ref{app:ablation:perchoice} discusses each design choice, Appendix~\ref{app:ablation:ranking} ranks their contributions, and Appendix~\ref{app:ablation:zjh} examines this transfer setting.

The resampler trades some accuracy for substantially lower training memory.
On HM3D, retaining all patch tokens reduces translation error from $0.155$ to $0.123$\,m, but increases measured training memory from $5.44$ to $21.34$\,GB at batch size $16$ (Appendix~\ref{app:ablation:efficiency}).
We therefore use $64$ latent tokens per frame as the default.
With noisy extrinsics, the default model's mean inter-group errors change by at most $0.015$\,m and $0.05^\circ$ under the tested perturbations, while its intra-group predictions refine the supplied geometry (Appendix~\ref{app:ablation:noise}).

Table~\ref{tab:controls} (left) retrains \ours{} on HM3D and NCLT with one factor changed at a time, keeping data and hyper-parameters fixed; training budgets are listed in the table footnote.
A block-diagonal mask that removes only the $A{\leftrightarrow}B$ exchange in the bridge, with all geometric inputs unchanged, raises the HM3D error from $0.155$\,m / $1.72^\circ$ to $0.562$\,m / $8.45^\circ$ and the NCLT error from $0.692$\,m / $2.47^\circ$ to $1.502$\,m / $3.76^\circ$, so the gain comes from cross-group reasoning rather than from the geometric priors alone.
Removing intrinsics degrades mainly translation, since the field of view must then be inferred from images and its error rescales the recovered metric geometry; the variant still outperforms Reloc3R on both datasets, so the advantage does not stem from access to $\mathbf{K}$.
A single model trained on variable group sizes $N\in[2,9]$ covers all evaluated sizes with one set of weights: at $N{=}3$ it is comparable to the fixed $3{+}3$ ablation, accuracy improves from $N{=}3$ to $N{=}7$ on both datasets, and NCLT keeps improving at $N{=}11$ beyond the training range, with end-to-end latency growing only from $41.8$ to $89.6$\,ms.
Finally, the same modules retrained on the frozen DA3~\cite{lin2025depthanything3} and $\pi^3$X~\cite{wang2025pi3} backbones both work on HM3D, and accuracy follows backbone capacity: the 1.36B-parameter $\pi^3$X reaches $0.151$\,m / $1.42^\circ$ and surpasses the 539M MapAnything version, whereas the 411M DA3-Large is less accurate, so \ours{} directly inherits improvements in the underlying foundation model (Appendices~\ref{app:ablation:conditioning}--\ref{app:ablation:backbones}).
% Table 4（camera-ready 新增）：左 = rebuttal Tab. II 受控变体（数值与附录 F.6–F.8 一致），
% 右 = rebuttal Tab. I 长时程里程计（数值与附录 G Table A16 一致）。
% 排版沿用 rebuttal 的做法：两块各自是独立的 booktabs 表，先各存进 savebox，
% 以右表为基准把左表等比缩放到严格同高，再把整行等比缩放到 \textwidth。
% 两表互不绑定行高（旧版塞进同一个 tabular 会让各自的 \cmidrule 叠出双行空隙、无关行被迫对齐）。
% caption 控制在 2 行，协议细节放表底 footnote；风格沿用 Table 1–3。
\newsavebox{\tabFourL}\newsavebox{\tabFourR}\newlength{\tabFourH}
\begin{table*}[t]
\centering
\caption{\textbf{Controlled variants (left) and long-horizon NCLT odometry (right)} (mean errors).
  \textcolor{cBest}{Cyan}/\textcolor{cSecond}{orange} mark the best/second-best non-oracle result; protocols are given in the footnote.}
\label{tab:controls}
\vspace{-0.1cm}
% ---------------- 左：受控变体 ----------------
\savebox{\tabFourL}{\footnotesize\setlength{\tabcolsep}{3pt}%
\begin{tabular}{llcccc}
\toprule
 & & \multicolumn{2}{c}{\textbf{HM3D}} & \multicolumn{2}{c}{\textbf{NCLT}} \\
\cmidrule(lr){3-4} \cmidrule(lr){5-6}
\multicolumn{2}{l}{Variant} & $t${\scriptsize(m)}$\downarrow$ & $r${\scriptsize($^\circ$)}$\downarrow$ & $t${\scriptsize(m)}$\downarrow$ & $r${\scriptsize($^\circ$)}$\downarrow$ \\
\midrule
\multicolumn{2}{l}{w/o intrinsics}        & 0.176 & 1.88 & 0.939 & 2.73 \\
\multicolumn{2}{l}{w/o cross-group attn.} & 0.562 & 8.45 & 1.502 & 3.76 \\
\midrule
\multirow{3}{*}{Var.\ win.$^\dagger$} & $N{=}3$  & 0.254 & 2.86 & 0.833 & 3.03 \\
 & $N{=}7$  & 0.193 & 1.91 & 0.732 & 2.66 \\
 & $N{=}11$ & 0.207 & 1.95 & \second{0.717} & \best{2.44} \\
\midrule
\multicolumn{2}{l}{DA3 backbone$^\dagger$}      & 0.263 & 2.64 & --- & --- \\
\multicolumn{2}{l}{$\pi^3$X backbone$^\dagger$} & \best{0.151} & \best{1.42} & --- & --- \\
\midrule
\multicolumn{2}{l}{\textbf{\ours{}~(default)}} & \second{0.155} & \second{1.72} & \best{0.692} & \second{2.47} \\
\bottomrule
\end{tabular}}
% ---------------- 右：长时程里程计 ----------------
\savebox{\tabFourR}{\footnotesize\setlength{\tabcolsep}{2.5pt}%
\begin{tabular}{llcc}
\toprule
 & & \multicolumn{2}{c}{\textbf{NCLT} {\scriptsize(open-loop)}} \\
\cmidrule(lr){3-4}
Method & Extrinsics used & RTE{\scriptsize(\%)}$\downarrow$ & RRE{\scriptsize($^\circ$/100m)}$\downarrow$ \\
\midrule
VGGT-SLAM       & none                 & 59.6 & 38.5 \\
LingBot-Map     & none                 & 45.3 & 20.7 \\
CUT3R (ft.)     & none                 & 43.2 & 19.1 \\
\midrule
Rig3R (retr.)   & rig metadata         & 48.8 & 24.3 \\
VGGT (ft.)      & scale via $A$ & 41.5 & 17.0 \\
Reloc3R (ft.)   & post-hoc agg. & \second{24.3} & \second{11.0} \\
MA-A            & group $A$ only       & 34.1 & 18.0 \\
\rowcolor{gray!8}
MA-AB~{\scriptsize(oracle)} & GT-aligned & 9.2 & 4.6 \\
\midrule
\textbf{\ours{}~(Ours)} & both groups   & \best{6.1} & \best{3.5} \\
\bottomrule
\end{tabular}}
% 以右表总高（height+depth）为基准，左表等比缩放到同高；整行再缩放到 \textwidth
\setlength{\tabFourH}{\ht\tabFourR}\addtolength{\tabFourH}{\dp\tabFourR}
\resizebox{\textwidth}{!}{%
  \resizebox*{!}{\tabFourH}{\usebox{\tabFourL}}\hspace{8pt}\usebox{\tabFourR}}

\vspace{2pt}
\parbox{\textwidth}{\footnotesize\raggedright%
Left: each variant retrains \ours{} with one factor changed; $^\dagger$\,two-thirds of the default training budget. The variable-window model is trained on $N\in[2,9]$, so $N{=}11$ lies beyond its training range; the alternative backbones are trained on HM3D only.
Right: open-loop composition over two held-out $\approx$6\,km traversals with KITTI segment errors and no loop closure; the first three rows are monocular, the rest take the multi-camera rig; column~2 states how each method uses the known intra-group extrinsics. (ft.)\,finetuned from the released checkpoint; (retr.)\,trained with DUSt3R initialization; VGGT-SLAM and LingBot-Map use released weights. MA-A/MA-AB include the ground-truth-assisted bias correction of Appendix~\ref{app:odometry}.}
\vspace{-0.4cm}
\end{table*}

% ==============================================================================
% ==============================================================================
% Section 5: Conclusion
% ==============================================================================
\section{Conclusion}
\label{sec:conclusion}

We introduced Group-to-Group pose estimation, a unified formulation that casts cross-sequence relocalization and multi-camera rig odometry as the same task with a shared input-output interface.
We presented \ours{}, which keeps a multi-view foundation model entirely frozen and adds three lightweight modules on top: a perceiver resampler, a cross-group bridge with merged self-attention, and a multi-frame pose head.
The frozen backbone fuses the known intra-group geometry into each group's visual features, and the lightweight modules supply the cross-group reasoning that the backbone lacks, producing every target pose in a single forward pass.
Across four datasets that span indoor and outdoor simulation, real-world cross-season capture, and zero-shot sim-to-real transfer, \ours{} attains the best non-oracle accuracy on both tasks while training only about 32M parameters and using relative poses as its only supervision signal.
The advantage is most pronounced on cross-season pairs and on the real 4-camera rig, two regimes in which methods that rely on cross-group visual correspondence break down.

\paragraph{Limitations.}
\ours{} treats intra-group geometry as a required input. Its accuracy therefore degrades when this geometry is severely corrupted or entirely absent, even though noise augmentation keeps the method robust within the perturbation range we evaluate.
The trainable modules inherit the perceptual capacity of the frozen backbone and require a backbone that can incorporate camera priors.
Our variable-window experiments cover up to eleven frames per group; substantially larger groups and dynamic intra-group geometry remain to be studied.
The long-horizon evaluation composes local estimates without loop closure, so drift can still accumulate.
Future work includes integrating \ours{} into a full SLAM system and training a single model jointly across datasets and tasks.

% ==============================================================================
% The acknowledgments are automatically included only in the final version.
\acknowledgments{This work was supported by the National Natural Science Foundation of China under Grants 62503144, 62522317, and 62373322.}

% ==============================================================================
% no \bibliographystyle is required -- corl_2026.sty auto-loads corlabbrvnat.
\bibliography{references}

@inproceedings{wang2024dust3r,
  title={Dust3r: Geometric 3d vision made easy},
  author={Wang, Shuzhe and Leroy, Vincent and Cabon, Yohann and Chidlovskii, Boris and Revaud, Jerome},
  booktitle={Proceedings of the IEEE/CVF conference on computer vision and pattern recognition},
  pages={20697--20709},
  year={2024}
}

@inproceedings{leroy2024mast3r,
  title={Grounding image matching in 3d with mast3r},
  author={Leroy, Vincent and Cabon, Yohann and Revaud, J{\'e}r{\^o}me},
  booktitle={European conference on computer vision},
  pages={71--91},
  year={2024},
  organization={Springer}
}

@inproceedings{wang2025vggt,
  title={Vggt: Visual geometry grounded transformer},
  author={Wang, Jianyuan and Chen, Minghao and Karaev, Nikita and Vedaldi, Andrea and Rupprecht, Christian and Novotny, David},
  booktitle={Proceedings of the Computer Vision and Pattern Recognition Conference},
  pages={5294--5306},
  year={2025}
}

@article{keetha2025mapanything,
  title={Mapanything: Universal feed-forward metric 3d reconstruction},
  author={Keetha, Nikhil and M{\"u}ller, Norman and Sch{\"o}nberger, Johannes and Porzi, Lorenzo and Zhang, Yuchen and Fischer, Tobias and Knapitsch, Arno and Zauss, Duncan and Weber, Ethan and Antunes, Nelson and others},
  journal={arXiv preprint arXiv:2509.13414},
  year={2025}
}

@inproceedings{dong2025reloc3r,
  title={Reloc3r: Large-scale training of relative camera pose regression for generalizable, fast, and accurate visual localization},
  author={Dong, Siyan and Wang, Shuzhe and Liu, Shaohui and Cai, Lulu and Fan, Qingnan and Kannala, Juho and Yang, Yanchao},
  booktitle={Proceedings of the Computer Vision and Pattern Recognition Conference},
  pages={16739--16752},
  year={2025}
}

@inproceedings{yang2025fast3r,
  title={Fast3r: Towards 3d reconstruction of 1000+ images in one forward pass},
  author={Yang, Jianing and Sax, Alexander and Liang, Kevin J and Henaff, Mikael and Tang, Hao and Cao, Ang and Chai, Joyce and Meier, Franziska and Feiszli, Matt},
  booktitle={Proceedings of the Computer Vision and Pattern Recognition Conference},
  pages={21924--21935},
  year={2025}
}

@inproceedings{elflein2025light3r,
  title={Light3r-sfm: Towards feed-forward structure-from-motion},
  author={Elflein, Sven and Zhou, Qunjie and Leal-Taix{\'e}, Laura},
  booktitle={Proceedings of the IEEE/CVF Conference on Computer Vision and Pattern Recognition},
  pages={16774--16784},
  year={2025}
}

@inproceedings{duisterhof2025mastrsfm,
  title={Mast3r-sfm: a fully-integrated solution for unconstrained structure-from-motion},
  author={Duisterhof, Bardienus Pieter and Zust, Lojze and Weinzaepfel, Philippe and Leroy, Vincent and Cabon, Yohann and Revaud, Jerome},
  booktitle={2025 International Conference on 3D Vision (3DV)},
  pages={1--10},
  year={2025},
  organization={IEEE}
}

@inproceedings{liu2025slam3r,
  title={Slam3r: Real-time dense scene reconstruction from monocular rgb videos},
  author={Liu, Yuzheng and Dong, Siyan and Wang, Shuzhe and Yin, Yingda and Yang, Yanchao and Fan, Qingnan and Chen, Baoquan},
  booktitle={Proceedings of the Computer Vision and Pattern Recognition Conference},
  pages={16651--16662},
  year={2025}
}

@inproceedings{cabon2025must3r,
  title={Must3r: Multi-view network for stereo 3d reconstruction},
  author={Cabon, Yohann and Stoffl, Lucas and Antsfeld, Leonid and Csurka, Gabriela and Chidlovskii, Boris and Revaud, Jerome and Leroy, Vincent},
  booktitle={Proceedings of the IEEE/CVF Conference on Computer Vision and Pattern Recognition},
  pages={1050--1060},
  year={2025}
}

@inproceedings{blumenkamp2024covisnet,
  title={CoViS-Net: A Cooperative Visual Spatial Foundation Model for Multi-Robot Applications},
  author={Blumenkamp, Jan and Morad, Steven and Gielis, Jennifer and Prorok, Amanda},
  booktitle={Conference on Robot Learning},
  pages={3780--3808},
  year={2025},
  organization={PMLR}
}

@article{jiao2026opennavmap,
  title={OpenNavMap: Structure-Free Topometric Mapping via Large-Scale Collaborative Localization},
  author={Jiao, Jianhao and Liu, Changkun and Yu, Jingwen and Liu, Boyi and Zhang, Qianyi and Wang, Yue and Kanoulas, Dimitrios},
  journal={arXiv preprint arXiv:2601.12291},
  year={2026}
}

@article{tian2022kimeramulti,
  title={Kimera-multi: Robust, distributed, dense metric-semantic slam for multi-robot systems},
  author={Tian, Yulun and Chang, Yun and Arias, Fernando Herrera and Nieto-Granda, Carlos and How, Jonathan P and Carlone, Luca},
  journal={IEEE transactions on robotics},
  volume={38},
  number={4},
  year={2022},
  publisher={IEEE}
}

@article{campos2021orbslam3,
  title={Orb-slam3: An accurate open-source library for visual, visual--inertial, and multimap slam},
  author={Campos, Carlos and Elvira, Richard and Rodr{\'\i}guez, Juan J G{\'o}mez and Montiel, Jos{\'e} MM and Tard{\'o}s, Juan D},
  journal={IEEE transactions on robotics},
  volume={37},
  number={6},
  pages={1874--1890},
  year={2021},
  publisher={IEEE}
}

@article{li2026rig3r,
  title={Rig3R: Rig-Aware Conditioning and Discovery for 3D Reconstruction},
  author={Li, Samuel and Kachana, Pujith and Chidananda, Prajwal and Nair, Saurabh and Furukawa, Yasutaka and Brown, Matthew A},
  journal={Advances in Neural Information Processing Systems},
  volume={38},
  pages={24139--24163},
  year={2026}
}

@article{oquab2024dinov2,
  title={Dinov2: Learning robust visual features without supervision},
  author={Oquab, Maxime and Darcet, Timoth{\'e}e and Moutakanni, Th{\'e}o and Vo, Huy and Szafraniec, Marc and Khalidov, Vasil and Fernandez, Pierre and Haziza, Daniel and Massa, Francisco and El-Nouby, Alaaeldin and others},
  journal={arXiv preprint arXiv:2304.07193},
  year={2023}
}

@inproceedings{detone2018superpoint,
  title={Superpoint: Self-supervised interest point detection and description},
  author={DeTone, Daniel and Malisiewicz, Tomasz and Rabinovich, Andrew},
  booktitle={Proceedings of the IEEE conference on computer vision and pattern recognition workshops},
  pages={224--236},
  year={2018}
}

@inproceedings{sarlin2020superglue,
  title={Superglue: Learning feature matching with graph neural networks},
  author={Sarlin, Paul-Edouard and DeTone, Daniel and Malisiewicz, Tomasz and Rabinovich, Andrew},
  booktitle={Proceedings of the IEEE/CVF conference on computer vision and pattern recognition},
  pages={4938--4947},
  year={2020}
}

@inproceedings{wang2024efficientloftr,
  title={Efficient LoFTR: Semi-dense local feature matching with sparse-like speed},
  author={Wang, Yifan and He, Xingyi and Peng, Sida and Tan, Dongli and Zhou, Xiaowei},
  booktitle={Proceedings of the IEEE/CVF conference on computer vision and pattern recognition},
  pages={21666--21675},
  year={2024}
}

@inproceedings{edstedt2024roma,
  title={Roma: Robust dense feature matching},
  author={Edstedt, Johan and Sun, Qiyu and B{\"o}kman, Georg and Wadenb{\"a}ck, M{\aa}rten and Felsberg, Michael},
  booktitle={Proceedings of the IEEE/CVF conference on computer vision and pattern recognition},
  pages={19790--19800},
  year={2024}
}

@article{edstedt2025romav2,
  title={RoMa v2: Harder Better Faster Denser Feature Matching},
  author={Edstedt, Johan and Nordstr{\"o}m, David and Zhang, Yushan and B{\"o}kman, Georg and Astermark, Jonathan and Larsson, Viktor and Heyden, Anders and Kahl, Fredrik and Wadenb{\"a}ck, M{\aa}rten and Felsberg, Michael},
  journal={arXiv preprint arXiv:2511.15706},
  year={2025}
}

@article{nordstrom2026loma,
  title={LoMa: Local Feature Matching Revisited},
  author={Nordstr{\"o}m, David and Edstedt, Johan and B{\"o}kman, Georg and Astermark, Jonathan and Heyden, Anders and Larsson, Viktor and Wadenb{\"a}ck, M{\aa}rten and Felsberg, Michael and Kahl, Fredrik},
  journal={arXiv preprint arXiv:2604.04931},
  year={2026}
}

@inproceedings{zhou2019continuity,
  title={On the continuity of rotation representations in neural networks},
  author={Zhou, Yi and Barnes, Connelly and Lu, Jingwan and Yang, Jimei and Li, Hao},
  booktitle={Proceedings of the IEEE/CVF conference on computer vision and pattern recognition},
  pages={5745--5753},
  year={2019}
}

@article{peterson2024roman,
  title={Roman: Open-set object map alignment for robust view-invariant global localization},
  author={Peterson, Mason B and Jia, Yixuan and Tian, Yulun and Thomas, Annika and How, Jonathan P},
  journal={arXiv preprint arXiv:2410.08262},
  year={2024}
}

@inproceedings{pless2003using,
  title={Using many cameras as one},
  author={Pless, Robert},
  booktitle={2003 IEEE Computer Society Conference on Computer Vision and Pattern Recognition, 2003. Proceedings.},
  volume={2},
  pages={II--587},
  year={2003},
  organization={IEEE}
}

@article{heng2015self,
  title={Self-calibration and visual slam with a multi-camera system on a micro aerial vehicle},
  author={Heng, Lionel and Lee, Gim Hee and Pollefeys, Marc},
  journal={Autonomous robots},
  volume={39},
  number={3},
  pages={259--277},
  year={2015},
  publisher={Springer}
}

@inproceedings{kazik2012real,
  title={Real-time 6d stereo visual odometry with non-overlapping fields of view},
  author={Kazik, Tim and Kneip, Laurent and Nikolic, Janosch and Pollefeys, Marc and Siegwart, Roland},
  booktitle={2012 IEEE Conference on computer vision and pattern recognition},
  pages={1529--1536},
  year={2012},
  organization={IEEE}
}

@article{lajoie2020door,
  title={DOOR-SLAM: Distributed, online, and outlier resilient SLAM for robotic teams},
  author={Lajoie, Pierre-Yves and Ramtoula, Benjamin and Chang, Yun and Carlone, Luca and Beltrame, Giovanni},
  journal={IEEE Robotics and Automation Letters},
  volume={5},
  number={2},
  pages={1656--1663},
  year={2020},
  publisher={IEEE}
}

@article{ramakrishnan2021habitat,
  title={Habitat-matterport 3d dataset (hm3d): 1000 large-scale 3d environments for embodied ai},
  author={Ramakrishnan, Santhosh K and Gokaslan, Aaron and Wijmans, Erik and Maksymets, Oleksandr and Clegg, Alex and Turner, John and Undersander, Eric and Galuba, Wojciech and Westbury, Andrew and Chang, Angel X and others},
  journal={arXiv preprint arXiv:2109.08238},
  year={2021}
}

@article{carlevaris2016university,
  title={University of Michigan North Campus long-term vision and lidar dataset},
  author={Carlevaris-Bianco, Nicholas and Ushani, Arash K and Eustice, Ryan M},
  journal={The International Journal of Robotics Research},
  volume={35},
  number={9},
  pages={1023--1035},
  year={2016},
  publisher={Sage Publications Sage UK: London, England}
}

@inproceedings{patel2025tartanground,
  title={Tartanground: A large-scale dataset for ground robot perception and navigation},
  author={Patel, Manthan and Yang, Fan and Qiu, Yuheng and Cadena, Cesar and Scherer, Sebastian and Hutter, Marco and Wang, Wenshan},
  booktitle={2025 IEEE/RSJ International Conference on Intelligent Robots and Systems (IROS)},
  pages={20524--20531},
  year={2025},
  organization={IEEE}
}

@article{lin2025depthanything3,
 title={Depth Anything 3: Recovering the Visual Space from Any Views},
 author={Lin, Haotong and Chen, Sili and Liew, Junhao and Chen, Donny Y. and Li, Zhenyu and Shi, Guang and Feng, Jiashi and Kang, Bingyi},
 journal={arXiv preprint arXiv:2511.10647}, year={2025}
}

@article{wang2025pi3,
 title={{\(\pi^3\)}: Permutation-Equivariant Visual Geometry Learning},
 author={Wang, Yifan and Zhou, Jianjun and Zhu, Haoyi and Chang, Wenzheng and Zhou, Yang and Li, Zizun and Chen, Junyi and Pang, Jiangmiao and Shen, Chunhua and He, Tong},
 journal={arXiv preprint arXiv:2507.13347}, year={2025}
}

@article{wang2025cut3r,
 title={Continuous 3D Perception Model with Persistent State},
 author={Wang, Qianqian and Zhang, Yifei and Holynski, Aleksander and Efros, Alexei A. and Kanazawa, Angjoo},
 journal={arXiv preprint arXiv:2501.12387}, year={2025}
}

@article{maggio2025vggtslam,
 title={{VGGT-SLAM}: Dense {RGB SLAM} Optimized on the {SL(4)} Manifold},
 author={Maggio, Dominic and Lim, Hyungtae and Carlone, Luca},
 journal={arXiv preprint arXiv:2505.12549}, year={2025}
}

@article{chen2026lingbotmap,
 title={{LingBot-Map}: Geometric Context Transformer for Streaming 3D Reconstruction},
 author={Chen, Lin-Zhuo and Gao, Jian and Zhang, Shangzhan and Chen, Yihang and Cheng, Ka Leong and Sun, Yipengjing and Hu, Liangxiao and Xue, Nan and Zhu, Xing and Shen, Yujun and Yao, Yao and Xu, Yinghao},
 journal={arXiv preprint arXiv:2604.14141}, year={2026}
}

% ==============================================================================
% =============================================================================
% 附录（Supplementary Material）——直接拼接自 appendix_standalone/main.tex 的 body 段，
% 内容未改写。bibliography 已在上方出现一次；附录与正文共用同一 references.bib。
% =============================================================================
\clearpage

% ---- 进入 appendix：章节自动 A, B, C, ... ----
\appendix
% ---- 表/图/公式加 "A" 前缀且从 1 顺序编号（与正文 Table 1 / Fig 1 / Eq (1) 区分）----
\setcounter{table}{0}\renewcommand{\thetable}{A\arabic{table}}
\setcounter{figure}{0}\renewcommand{\thefigure}{A\arabic{figure}}
\setcounter{equation}{0}\renewcommand{\theequation}{A\arabic{equation}}
% hyperref 锚点也要同步加 A 前缀，否则附录 Fig. A1/Table A1 的链接会跳到正文 Fig. 1/Table 1（日志 "destination with the same identifier"）
\renewcommand{\theHfigure}{A\arabic{figure}}
\renewcommand{\theHtable}{A\arabic{table}}

% ---- 标题（匿名；无作者）----
\begin{center}
  {\LARGE\bfseries Supplementary Material}\\[4pt]
  {\large G2G: Exploiting Intra-Group Geometry for Inter-Group Pose Estimation}
\end{center}
\vspace{1.2em}

% =============================================================================
% 内容索引（无标题小节）：直观罗列 A–I 各节内容；hyperref 链接，点击/右键可跳转。
% 各 \hyperref 目标为对应 section 的 \label{app:*}；首次编译会有 undefined ref 警告，
% 二次编译后消失（与 \bibliography 同理）。
% =============================================================================
\noindent
This supplementary material is organized into nine self-contained parts.
Each entry below names a section and links to it, so that the corresponding content can be reached directly by clicking the entry.
\begin{itemize}\setlength{\itemsep}{2.5pt}\setlength{\parskip}{0pt}
  \item \hyperref[app:impl]{\textbf{A.\ Implementation Details}}: the complete loss formulation, the three-phase overlap curriculum, the optimizer and task-transfer schedule, the extrinsic-noise augmentation, and the trainable-module parameter breakdown deferred from the main paper.
  \item \hyperref[app:data]{\textbf{B.\ Datasets and Training-Pair Construction}}: the four evaluation datasets with their splits, and the covisibility-driven mining of training pairs.
  \item \hyperref[app:stage1]{\textbf{C.\ Covisibility-Based Window Selection for Deployment}}: an optional front-end that selects each observation window from a long trajectory by predicting inter-frame covisibility from RGB alone, used when the windows are not provided in advance.
  \item \hyperref[app:baselines]{\textbf{D.\ Baseline Methods and Evaluation Protocols}}: per-baseline configuration, scale and metric conventions, the pretrained-versus-finetuned comparison, and two controlled failure diagnostics.
  \item \hyperref[app:metrics]{\textbf{E.\ Scale-Free Metrics and Detailed Errors}}: the scale-free mAA and threshold-accuracy metrics, median errors with complete per-setting tables, and an accuracy breakdown by field-of-view overlap.
  \item \hyperref[app:ablation]{\textbf{F.\ Detailed Ablation Analysis}}: per-design-choice ablations, the ZJH sim-to-real anomaly, component-contribution ranking, extrinsic-noise robustness, the accuracy-efficiency trade-off, intrinsic and attention controls, variable group sizes, and alternative backbones.
  \item \hyperref[app:odometry]{\textbf{G.\ Long-Horizon Odometry}}: accumulated motion error on two NCLT traversals, with input, scale, and bias-correction protocols.
  \item \hyperref[app:qualitative]{\textbf{H.\ Additional Qualitative Examples}}: per-dataset relocalization examples that statically mirror the relocalization videos on the project page, covering synthetic, cross-season, and sim-to-real cases.
  \item \hyperref[app:contributions]{\textbf{I.\ Author Contributions}}: the individual contributions of each author to the formulation, implementation, data collection, experiments, and manuscript.
\end{itemize}
\vspace{0.6em}

% =============================================================================
% 9 节正文已启用（A–I）。引用正文用纯文本，不用 \ref。
% =============================================================================
% =============================================================================
% Appendix A: Implementation Details
% 撰写依据: docs/superpowers/specs/2026-06-01-corl-appendix-spec.md  (Task A)
% 状态: 完成。补全正文 Method 中删除的 Training 段: 损失函数、课程学习、噪声增广、
%       优化器、精度、参数分解。
% =============================================================================
\section{Implementation Details}
\label{app:impl}

This section provides the training details deferred from Sec.~3 of the main paper, including the loss formulation, the three-phase curriculum schedule, the optimizer and task-transfer configuration, the extrinsic noise augmentation, and the trainable parameter breakdown.

% -----------------------------------------------------------------------------
\subsection{Loss Function}
\label{app:impl:loss}

The per-frame loss in Eq.~\eqref{eq:training_loss} combines a chordal rotation distance with a mean absolute translation error:
\begin{equation}
 \mathcal L_\text{frame}=\lambda_R d_R(\mathbf R_\text{pred},\mathbf R_\text{gt})
 +\frac{\lambda_t}{3}\|\mathbf t_\text{pred}-\mathbf t_\text{gt}\|_1,
 \label{eq:appA_loss}
\end{equation}
with $\lambda_R=5$ and $\lambda_t=1$.
Writing $\mathbf\Delta=\mathbf R_\text{pred}^{\top}\mathbf R_\text{gt}-\mathbf I$, the released configurations use two chordal forms:
\begin{equation}
 d_R^{\mathrm{L1}}=\frac{1}{9}\sum_{u,v=1}^{3}|\Delta_{uv}|,
 \qquad d_R^{\mathrm{F}}=\frac{1}{9}\|\mathbf\Delta\|_F^2.
\end{equation}
Relocalization and the HM3D rig configurations use $d_R^{\mathrm{L1}}$; the TartanGround, NCLT, and ZJH rig configurations use $d_R^{\mathrm{F}}$.
The squared Frobenius form is monotonic in the relative rotation angle over $[0,\pi]$.
Both rotation forms average over matrix entries, and the translation term averages over its three coordinates.

The intra-group and inter-group terms are averaged separately over their target frames and then combined:
\begin{equation}
 \mathcal L=\frac{w_\text{intra}}{N_A-1}\sum_{i=1}^{N_A-1}\mathcal L_\text{frame}^{A_i}
 +\frac{w_\text{inter}}{N_B}\sum_{j=0}^{N_B-1}\mathcal L_\text{frame}^{B_j},
 \label{eq:appA_total_loss}
\end{equation}
with $w_\text{intra}=0.5$ and $w_\text{inter}=1$.
The higher inter-group weight prioritizes the unknown alignment, while intra-group supervision trains corrections to noisy input geometry.
The implementation also averages over the batch.

% -----------------------------------------------------------------------------
\begin{figure*}[t]
  \centering
  \includegraphics[width=\linewidth]{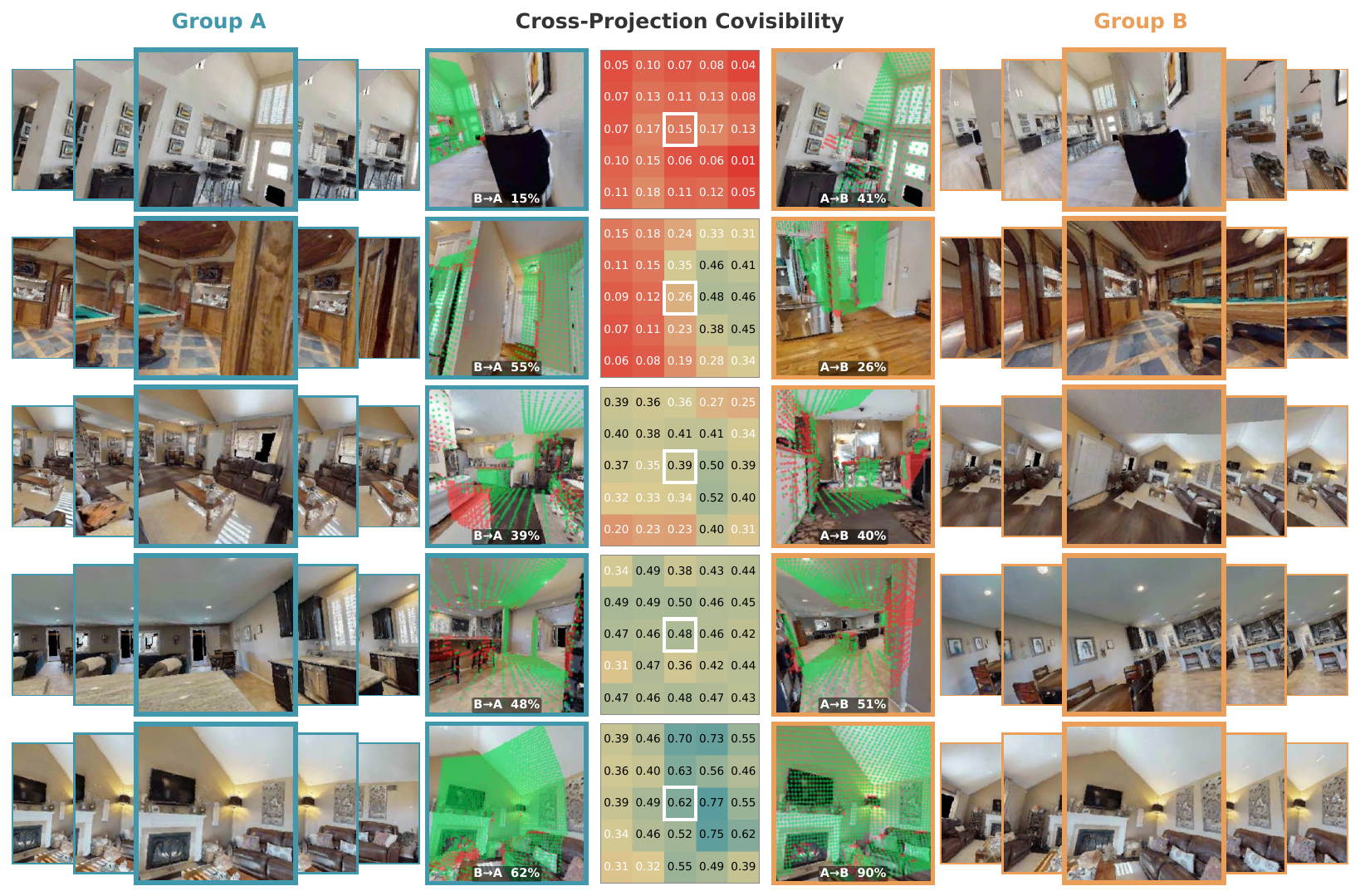}
  \caption{\textbf{Cross-group overlap examples at five difficulty levels.}
  Each row shows a pair of five-frame groups (Group~A in cyan, Group~B in orange) together with the $5{\times}5$ pairwise overlap matrix (center).
  The two images flanking the matrix visualize the cross-projection covisibility of the center frames A2 and B2: pixels from the other center frame are reprojected into the local frame, shown as green points where the depth is consistent and red points where it is not ($\tau{=}0.20$\,m).
  The percentage below each panel reports the fraction of reprojected pixels that pass the depth consistency check.
  As the center-frame covisibility decreases from $0.62$ (bottom) to $0.15$ (top), the co-visible area shrinks substantially, leaving fewer visual cues for the cross-group pose estimation.}
  \label{fig:appA_overlap_examples}
  \vspace{-0.4cm}
\end{figure*}

\begin{figure}[t]
  \centering
  \includegraphics[width=0.95\linewidth]{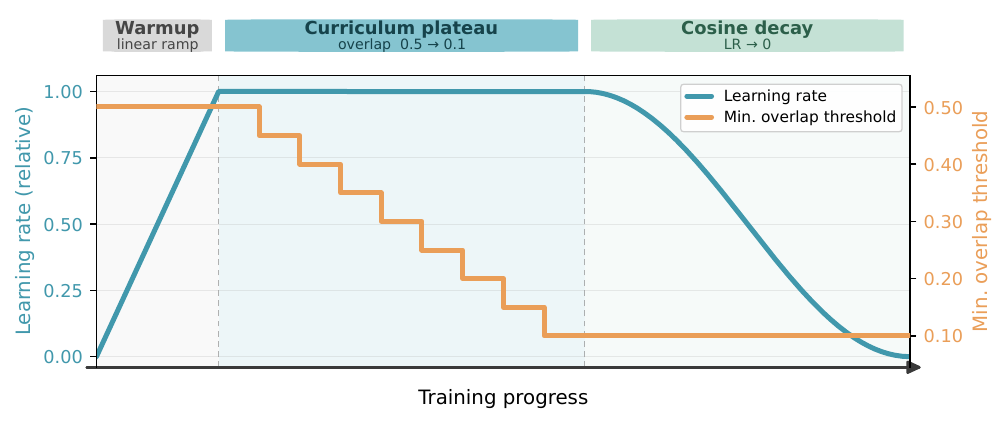}
  \vspace{-0.15cm}
  \caption{\textbf{Three-phase training schedule.}
  The relative learning rate (left axis) and the minimum overlap threshold for sampled training pairs (right axis) are shown against training progress.
  In the warmup phase, the learning rate ramps linearly to its peak while only high-overlap pairs (score $\geq 0.5$) are sampled.
  In the plateau phase, the learning rate is held fixed while the overlap threshold is lowered through nine discrete levels from $0.5$ to $0.1$, progressively introducing harder low-overlap pairs.
  Once the threshold reaches $0.1$, the decay phase applies cosine annealing to drive the learning rate to zero.
  The horizontal axis denotes training progress rather than a fixed step count, because each plateau level advances adaptively once a sliding-window convergence criterion is met.}
  \label{fig:appA_training}
  \vspace{-0.4cm}
\end{figure}

\subsection{Three-Phase Curriculum}
\label{app:impl:curriculum}

The difficulty of a training pair is governed by the visual overlap between the two trajectory groups.
Pairs with low overlap share fewer visual cues across groups, providing weaker constraints for cross-group pose estimation and making the regression harder.
Each entry of the pairwise overlap matrix is a symmetric frame-pair overlap equal to the minimum of the two directional covisibility ratios, where a directional ratio is the fraction of one frame's pixels that remain visible in the other. These ratios are computed offline from depth by cross-projection, as described in Sec.~\ref{app:data:pipeline}.
For a $5{\times}5$ window, the overlap score is the \emph{max-mean} of the corresponding block: within each group, every frame is matched to its best counterpart in the other group, these per-frame maxima are averaged, and the two directional averages are averaged in turn.
This aggregation captures whether every frame in a group has at least one well-matching counterpart, which is more informative than a simple mean over all $25$ frame pairs.
Training on all overlap levels from the outset leads to unstable convergence, so a three-phase curriculum is adopted that introduces progressively harder pairs.
Fig.~\ref{fig:appA_overlap_examples} illustrates five representative pairs at different difficulty levels, showing how the center-frame covisibility (ranging from $0.15$ to $0.62$) diminishes as the window-level overlap score decreases.

\paragraph{Warmup phase.}
The learning rate ramps linearly from $0$ to $10^{-4}$ over $1{,}000$ optimizer steps.
Only easy pairs whose symmetric overlap score is at least $0.5$ are sampled during this phase.

\paragraph{Plateau phase.}
The learning rate is held constant at $10^{-4}$.
The minimum overlap threshold for sampled pairs is progressively lowered through nine discrete levels:
$0.50 \to 0.45 \to 0.40 \to 0.35 \to 0.30 \to 0.25 \to 0.20 \to 0.15 \to 0.10$.
Advancement to the next level is gated by a convergence criterion on the anchor-pair (B$_0$\!$\to$\!A$_0$) prediction: the mean rotation and translation losses over a sliding window of $200$ optimizer steps must both fall below fixed thresholds.
This mechanism ensures that the model stabilizes at each difficulty before harder pairs are introduced.

\paragraph{Decay phase.}
Once the overlap threshold reaches $0.10$, all pairs with valid overlap are used for training and a cosine annealing schedule reduces the learning rate from $10^{-4}$ to $0$ over the remaining epochs.
Fig.~\ref{fig:appA_training} illustrates how the learning rate and the overlap threshold evolve jointly across the three phases.

% -----------------------------------------------------------------------------
\subsection{Training Details}
\label{app:impl:training}

\paragraph{Optimizer and training budget.}
We use AdamW with $\beta_1 = 0.9$, $\beta_2 = 0.999$, weight decay $0.01$, and gradient clipping at $5.0$.
The effective batch size is $128$ (32 per GPU $\times$ 4 GPUs with DDP).
Training runs for two epochs of $100{,}000$ optimizer steps each; within this budget, the curriculum thresholds advance according to convergence (Sec.~\ref{app:impl:curriculum}) rather than at fixed step counts.

\paragraph{Task transfer.}
The rig odometry task (Task~2) benefits from initializing with weights pretrained on the relocalization task (Task~1) on the same dataset.
This initialization accelerates convergence and improves final accuracy, because the cross-group reasoning learned for relocalization transfers directly to the rig setting, where the principal difference is the spatial, rather than temporal, arrangement of intra-group frames.

% -----------------------------------------------------------------------------
\subsection{Extrinsic Noise Augmentation}
\label{app:impl:noise}

Intra-group extrinsics obtained from visual odometry, SLAM, or SfM maps are inherently imprecise, and rig calibrations may drift over time.
To improve robustness, we perturb each non-anchor input extrinsic with a random $\SE$ noise sample during training.
For each frame, the rotation perturbation is drawn from an isotropic Gaussian on the tangent space of $\SO$ with standard deviation $1.5^\circ$, and the translation perturbation is drawn from an isotropic Gaussian with standard deviation $0.1$\,m.
Both groups $A$ and $B$ receive noise, while the anchor frame $A_0$ always remains at identity.
The ground-truth supervision poses are kept noise-free.
The model learns to recover clean geometry from noisy input, directly supporting the role of the intra-group predictions $\{\pose{A_0}{A_i}\}$ in correcting odometry or calibration errors.

% -----------------------------------------------------------------------------
\subsection{Trainable Parameter Breakdown}
\label{app:impl:params}

The frozen backbone accounts for approximately 539M parameters.
The three trainable modules total approximately 32.1M parameters, under 6\% of the full model.
Table~\ref{tab:appA_params} shows the per-module breakdown.

\begin{table}[ht]
  \centering
  \caption{\textbf{Trainable parameter count per module.}
  All numbers are measured on the default \ours{} configuration
  ($L{=}64$ latent tokens, $D{=}768$).
  The frozen MapAnything backbone ($\approx$539M) is excluded.}
  \label{tab:appA_params}
  \vspace{0.4em}
  \begin{tabular}{lrc}
    \toprule
    Module & Parameters & \% of trainable \\
    \midrule
    Perceiver Resampler       & 14.23M & 44.3\% \\
    Cross-Group Bridge        & 14.18M & 44.2\% \\
    Multi-Frame Pose Head     &  3.70M & 11.5\% \\
    \midrule
    \textbf{Total trainable}  & \textbf{32.11M} & 100\% \\
    \bottomrule
  \end{tabular}
\end{table}

The perceiver resampler includes $L{=}64$ learnable query tokens and two cross-attention layers that compress the $P$ patch tokens per frame into the $L$ latent tokens.
The cross-group bridge consists of two merged self-attention layers along with the frame, group, and anchor embeddings (Eq.~\eqref{eq:bridge_token}).
The multi-frame pose head contains a single cross-attention layer shared across all target frames, two learnable pose queries, a set of frame identity embeddings, and two MLPs that decode the rotation and translation features.

\paragraph{What 32M parameters buy.}
The frozen backbone already fuses intra-group geometry into per-frame features, but it cannot relate the two groups because it was never trained on cross-group data.
The 32M trainable modules close this gap: as shown in Table~1 of the main paper, the MA-A baseline reuses the same frozen backbone yet reaches only $1.01$\,m on HM3D, whereas \ours{} reduces this to $0.16$\,m, recovering most of the distance to the MA-AB oracle reference ($0.09$\,m).

Freezing the backbone also serves a representational purpose beyond efficiency.
The pretrained MapAnything encoder carries a visual-geometric representation learned from large-scale diverse data.
\ours{} preserves both the image encoder and the pretrained geometric fusion, confining adaptation to the lightweight cross-group modules.
Our reproduced Rig3R baseline also freezes its DINOv2 image encoder, but trains its DUSt3R-initialized cross-view decoder and prediction modules.
It exhibits a substantial training-to-validation gap on TartanGround (Sec.~\ref{app:baselines:rig3r}); this observation does not by itself isolate the causal effect of freezing the geometric decoder.

A detailed efficiency comparison, including the latent-count trade-off and baseline benchmarks, is given in Sec.~\ref{app:ablation:efficiency}.

% =============================================================================
% Appendix B: Datasets and Training-Pair Construction
% 撰写依据: docs/superpowers/specs/2026-06-01-corl-appendix-spec.md  (Task B)
% 状态: 完成。四个数据集详情 + 共视性驱动的配对管线 + 汇总表。
% =============================================================================
\section{Datasets and Training-Pair Construction}
\label{app:data}

The main paper evaluates \ours{} on four datasets that span indoor simulation, outdoor simulation, real-world cross-season capture, and zero-shot sim-to-real transfer.
This section provides the construction details and splits deferred from Sec.~4 of the main paper.

% -----------------------------------------------------------------------------
\subsection{Dataset Overview}
\label{app:data:overview}

\begin{figure}[t]
  \centering
  \includegraphics[width=\linewidth]{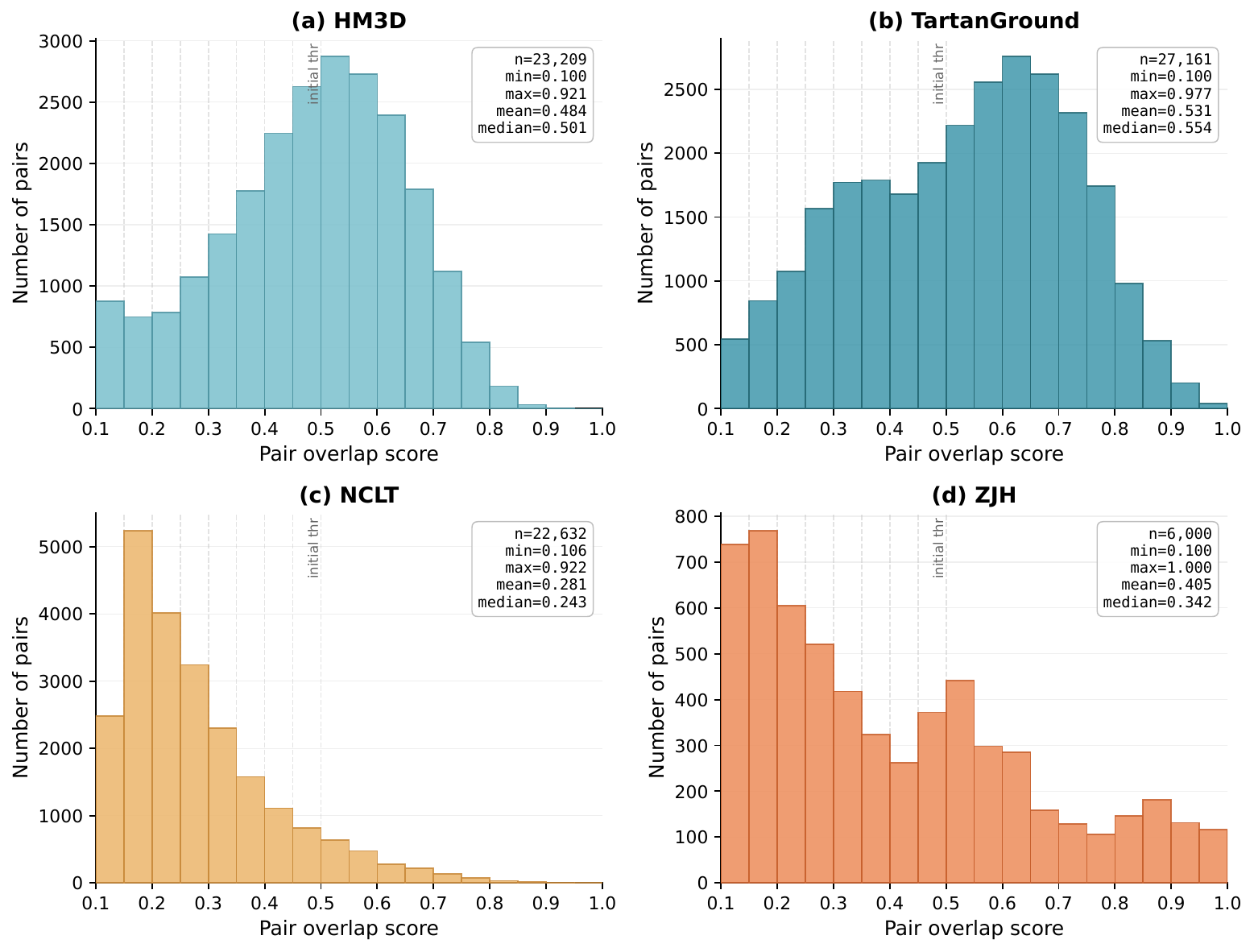}
  \caption{\textbf{Per-pair overlap distribution (overlap $\geq 0.10$).}
  Histograms (bin width $0.05$) of the symmetric overlap score for the evaluation pairs of each dataset, excluding pairs with overlap below $0.10$:
  \emph{(a)} HM3D (mean $0.48$),
  \emph{(b)} TartanGround (mean $0.53$),
  \emph{(c)} NCLT (mean $0.28$), and
  \emph{(d)} ZJH (mean $0.40$).
  Vertical dashed lines mark the nine curriculum-learning thresholds (Sec.~\ref{app:impl:curriculum}).}
  \label{fig:appA_overlap_dist}
\end{figure}

Table~\ref{tab:appA_datasets} summarizes the four datasets.
All images are resized to $224 \times 224$ before entering the model. The four datasets differ markedly in how much field-of-view overlap their evaluation pairs exhibit.
Fig.~\ref{fig:appA_overlap_dist} shows the per-pair overlap distribution for each dataset, computed on the real evaluation pairs rather than on the training distribution.

\begin{table}[h]
  \centering
  \caption{\textbf{Dataset summary.}
  ``Env.''\ counts distinct 3D environments or capture sites;
  ``Seq./env.''\ is the number of sub-sequences collected per environment;
  ``Cams''\ is the number of cameras per rig;
  ``Pano.''\ indicates whether the cameras collectively provide $360^\circ$ azimuthal coverage.}
  \label{tab:appA_datasets}
  \vspace{0.4em}
  \small
  \begin{tabular}{lcccccc}
    \toprule
    Dataset & Domain & Env. & Seq./env. & Split & Cams & Pano. \\
    \midrule
    HM3D         & Indoor sim  & 900     & 25      & 800 / 100  & 8 & \checkmark \\
    TartanGround & Outdoor sim & 63      & 12--160 & 53 / 10    & 4 & \checkmark \\
    NCLT         & Real campus & 1       & 10\rlap{$^\dagger$}  & 8 / 2      & 5 & \checkmark \\
    ZJH (sim)    & Indoor sim  & 25      & 2--18   & 22 / 3     & 4 & -- \\
    ZJH (real)   & Real indoor & 3       & 1\rlap{$^\ddagger$}  & test only   & 4 & -- \\
    \bottomrule
  \end{tabular}
  \vspace{0.2em}

  \raggedright\footnotesize
  $^\dagger$\,Full-length campus traversals on different dates/seasons (cross-season revisits of overlapping areas).\\
  $^\ddagger$\,Full-length indoor traversal per room (multi-lap walks along different routes).
\end{table}

NCLT exhibits the lowest overlap among all four datasets, with a median of $0.24$.
This reflects the inherent difficulty of cross-sequence matching: evaluation pairs span different dates and seasons, during which the campus undergoes substantial changes in vegetation, lighting, building facades, and construction activity, reducing the fraction of mutually visible content even when the vehicle revisits the same physical locations.

% -----------------------------------------------------------------------------
\subsection{HM3D}
\label{app:data:hm3d}

Habitat-Matterport 3D (HM3D)~\cite{ramakrishnan2021habitat} provides 800 training and 100 validation indoor scenes reconstructed from real-world scans.
For each scene, 25 navigable trajectories are sampled with the Habitat simulator.
Along each trajectory, an 8-camera rig renders $224 \times 224$ RGB images together with $224 \times 224$ 16-bit depth maps.
The eight cameras are evenly spaced at $45^\circ$ azimuthal intervals, and each camera covers a $90^\circ$ horizontal field of view, so the rig attains full $360^\circ$ panoramic coverage.
The per-camera intrinsics are randomized over a horizontal field of view from $45^\circ$ to $120^\circ$, and the rig extrinsics receive a random SE(3) perturbation per scene.
These eight cameras are used in two task configurations.
For relocalization, each camera is treated as an independent monocular trajectory.
For rig odometry, a fixed subset of the cameras, such as four, is selected at random to form the rig; within a scene the selected cameras and their extrinsic perturbation are held fixed across all samples.
Training pairs are formed from cross-trajectory windows within the same scene, following the pipeline described in Sec.~\ref{app:data:pipeline}.
The evaluation set contains 22,255 pairs sampled uniformly across the 100 validation scenes.

% -----------------------------------------------------------------------------
\subsection{TartanGround}
\label{app:data:tartanground}

TartanGround~\cite{patel2025tartanground} consists of 53 training and 10 validation outdoor scenes rendered in Unreal Engine with AirSim ground-robot trajectories.
The number of trajectories varies across scenes, from 12 to 160, with a mean of about $40$.
Each scene provides 4 cameras pointing at the cardinal directions, and each camera has a native $90^\circ$ horizontal field of view, so the four cameras together span $360^\circ$ panoramic coverage.
The per-camera intrinsics are augmented by random center-cropping, which simulates a horizontal field of view in $[45^\circ, 90^\circ)$.
Each camera direction can be used as an independent monocular trajectory, or the four cameras at a given timestep can be grouped into a single rig.
The evaluation set contains 26,872 pairs.

% -----------------------------------------------------------------------------
\subsection{NCLT}
\label{app:data:nclt}

The University of Michigan North Campus Long-Term (NCLT) dataset~\cite{carlevaris2016university} records repeated traversals of the same campus.
The data is captured by a Segway-mounted Ladybug3 platform with 5 cameras and a Velodyne HDL-32E LiDAR.
The five cameras together provide approximately $360^\circ$ panoramic azimuthal coverage.
Each camera is center-cropped to a randomized horizontal field of view in $[45^\circ, 105^\circ]$ and resized to $224 \times 224$.
We use 8 dates for training and 2 dates for validation, namely the 2012-02-19 winter traversal and the 2012-08-20 summer traversal.
The training and validation splits sample different dates at the same location, so the evaluation pairs routinely cross seasonal and illumination boundaries.
This cross-season setting isolates the benefit of geometry conditioning over pure visual matching: at a given location, the visual appearance, the vegetation cover, and even the built structures can change substantially between visits.

The 5-camera rig is used in two configurations.
In the single-camera configuration, the rig is decomposed into independent single-camera subsequences, and windows are matched across different cameras or different dates.
In the rig configuration, the five cameras are kept as one group, and rig poses are matched within the same date or across dates.
Both configurations involve cross-session evaluation, which makes NCLT the only benchmark that tests cross-session generalization in both single-view and multi-camera settings.
The evaluation set contains 22,632 pairs.

Unlike HM3D and TartanGround, which use dense rendered depth for overlap computation, NCLT overlap labels are derived from LiDAR-projected sparse depth with approximately 27\% pixel coverage per frame.
The resulting overlap scores are noisier, but they remain well-defined, because the denominator is the number of valid LiDAR pixels rather than the full image area.

% -----------------------------------------------------------------------------
\subsection{ZJH: Sim-to-Real Indoor Dataset}
\label{app:data:zjh}

ZJH is a self-collected dataset built around a humanoid robot platform.
It is designed to test sim-to-real transfer in a setting relevant to emerging legged robotics applications.
Bipedal locomotion introduces periodic gait oscillations and high-frequency vibrations that are absent on wheeled platforms, which makes robust pose estimation particularly challenging.

The simulation component consists of 25 indoor environments reconstructed with 3D Gaussian Splatting from real-world scans.
A total of 131 trajectories are sampled along humanoid walking paths, with 2 to 18 trajectories per environment.
Of the 25 environments, 22 are used for training and 3 are held out for testing.

The 4-camera rig comprises two forward-facing cameras and two oblique cameras.
The two forward-facing cameras form a stereo pair with a 7.26\,cm baseline, and each oblique camera is offset by approximately $70^\circ$ from the forward direction.
Unlike the panoramic rigs of HM3D, TartanGround, and NCLT, the ZJH rig provides only partial forward-biased coverage.
This coverage spans approximately $180^\circ$ in azimuth and leaves small blind spots between adjacent cameras.
In simulation, the rig extrinsics are initialized from the hardware calibration of the physical robot and then receive a random SE(3) perturbation per scene.
This perturbation enriches the diversity of rig configurations and accounts for inter-unit variation across robot instances.
During training, additional SE(3) noise is injected into the input extrinsics, as detailed in Sec.~\ref{app:impl:noise}, to improve robustness against residual calibration errors on the deployed platform.

The real-world component consists of 3 sequences captured by a physical humanoid robot with the same 4-camera rig, walking through previously unseen environments.
The real sequences exhibit the full motion profile of bipedal walking, including periodic vertical oscillation from the gait cycle, abrupt heading changes, and vibrations transmitted through the rigid body.

The same fine-tuned \ours{} checkpoint is evaluated on both the simulated test split and the real captures.
The evaluation is performed in a zero-shot manner, without any adaptation or fine-tuning on real data.
This protocol validates that the proposed architecture can be trained entirely in simulation and deployed directly on a physical humanoid.
It therefore provides an efficient pipeline that avoids costly real-world data collection.
The evaluation pool contains 6,000 pairs across the simulated and real splits.

% -----------------------------------------------------------------------------
\subsection{Covisibility-Driven Pair Mining}
\label{app:data:pipeline}

\begin{figure*}[t]
  \centering
  \includegraphics[width=\linewidth]{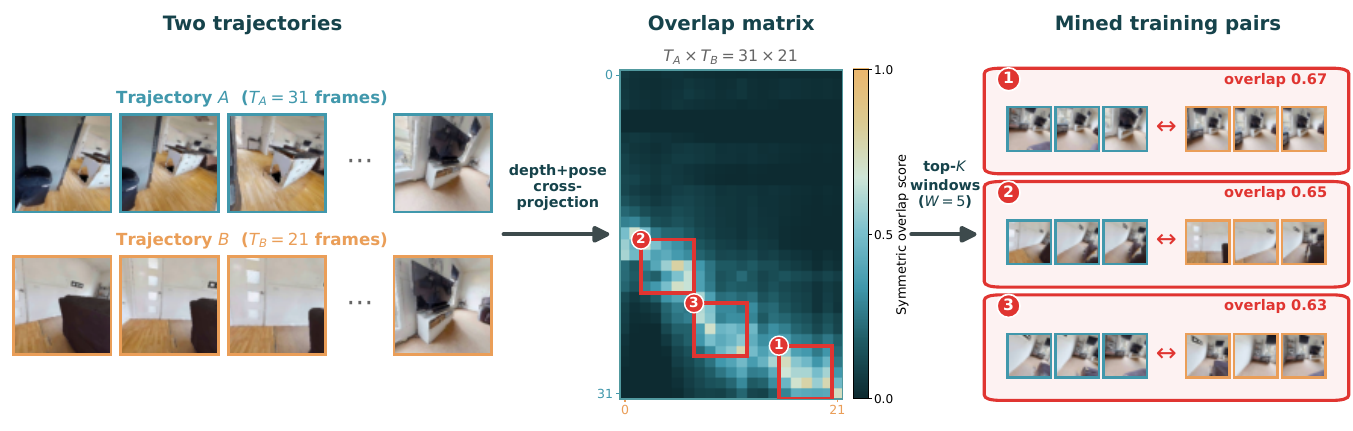}
  \vspace{-0.4cm}
  \caption{\textbf{Covisibility-driven pair-mining pipeline.}
  Left: two complete camera trajectories from the same scene, shown as RGB filmstrips annotated with their lengths $T_A$ and $T_B$.
  Their depth maps and known poses are cross-projected to test depth consistency, which yields the dense $T_A \times T_B$ symmetric overlap matrix in the middle, where rows index trajectory $A$ and columns index trajectory $B$.
  A sliding window of size $W{=}5$ then sweeps the matrix, and the top-$K$ highest-overlap windows are retained as group-to-group training pairs.
  Red boxes mark the three highest-overlap windows on this real HM3D matrix, and the right panel shows their actual frames together with the measured overlap score, numbered to match the boxes.
  Depth maps are used only for offline overlap labeling. The pose estimator receives known intra-group poses, while the ground-truth inter-group transform is used for supervision, not as an input.}
  \label{fig:appA_pair_mining}
  \vspace{-0.4cm}
\end{figure*}

Training pairs are constructed offline using a shared pipeline across all four datasets.
The pipeline proceeds in two steps, illustrated end-to-end in Fig.~\ref{fig:appA_pair_mining}.

\begin{figure}[t]
  \centering
  \includegraphics[width=0.95\linewidth]{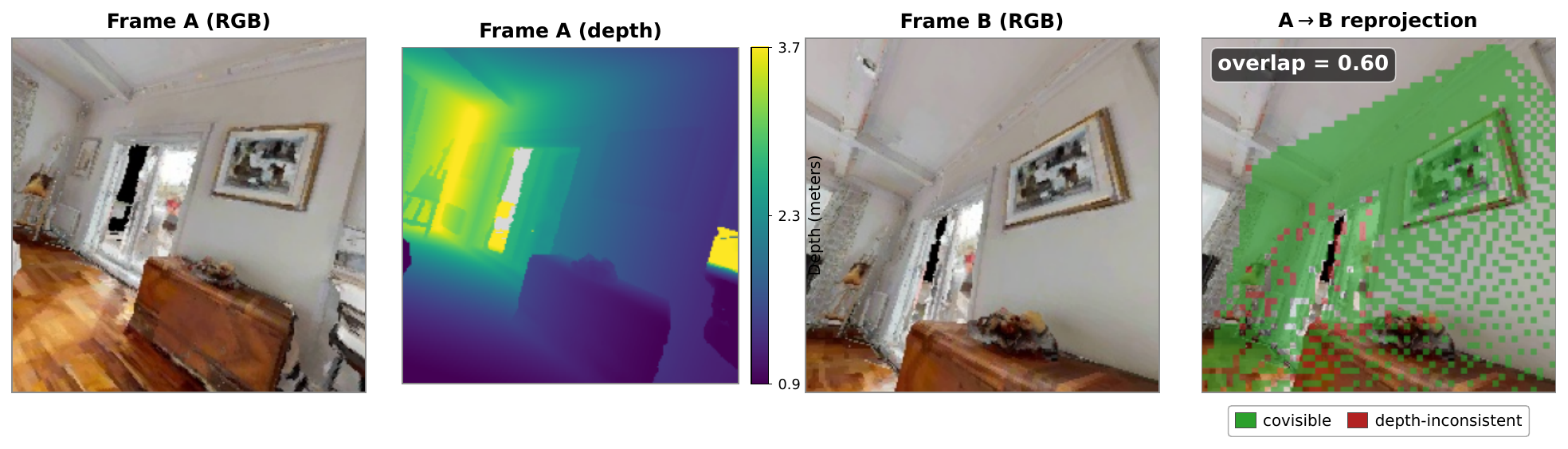}
  \vspace{-0.4cm}
  \caption{\textbf{Covisibility labels from depth reprojection.}
  A real HM3D frame pair with a symmetric overlap of $0.60$.
  From left to right: the RGB image of frame~A; its metric depth map; the RGB image of frame~B, observed from a different viewpoint; and the pixels of frame~A reprojected into frame~B.
  Reprojected pixels are drawn in green and counted as covisible when their depth is consistent, while pixels that land inside frame~B but disagree in depth are drawn in red and rejected.
  These overlap labels are produced offline. Depth is not an input to the pose estimator; known intra-group poses are supplied, but the ground-truth inter-group transform is not.}
  \label{fig:appA_cross_proj}
  \vspace{-0.3cm}
\end{figure}

\paragraph{Overlap matrix computation.}
The overlap score introduced in Sec.~\ref{app:impl:curriculum} is computed offline by depth cross-projection.
For each pair of camera trajectories in the same scene, a dense overlap matrix is formed whose entry $(i,j)$ stores the symmetric overlap between frame $i$ of trajectory $A$ and frame $j$ of trajectory $B$.
The depth map of one frame is projected into the other using the known camera poses and intrinsics, and a pixel is counted as covisible when the projected depth agrees with the observed depth within an absolute tolerance of $0.2$\,m.
The directional ratio is the fraction of covisible pixels in each direction, and the symmetric entry is their minimum.
Fig.~\ref{fig:appA_cross_proj} visualizes this computation on a real HM3D frame pair.

\begin{figure}[t]
  \centering
  \includegraphics[width=0.95\linewidth]{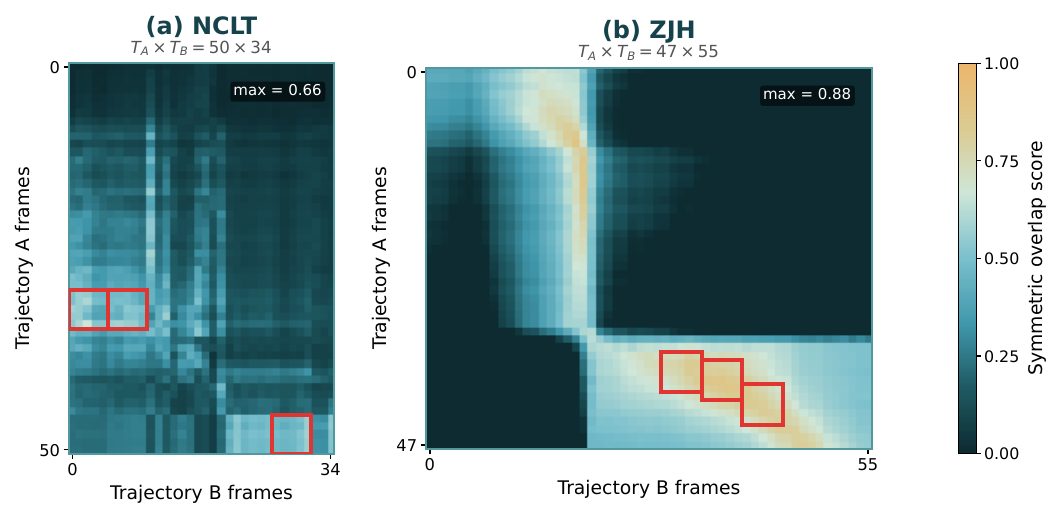}
  \vspace{-0.3cm}
  \caption{\textbf{Real mined overlap matrices and selected windows.}
  Two symmetric overlap matrices, shown with a shared colormap over the range $0$ to $1$:
  \emph{(a)} a cross-season NCLT pair and \emph{(b)} a ZJH pair.
  Red boxes mark example selected $W{=}5$ windows retained as training pairs.
  The diagonal band reflects co-directional traversal, while off-diagonal blocks correspond to loop closures or scene revisits.
  All matrices are computed offline by depth-based cross-projection and are used only to construct training pairs.}
  \label{fig:appA_overlap_matrix}
  \vspace{-0.4cm}
\end{figure}

\paragraph{Window selection.}
Given the overlap matrix, a sliding window of size $W{=}5$ sweeps over both sequences.
Each $W \times W$ window is scored by the average best-match overlap between its frames in the two sequences.
The globally highest-scoring windows are then retained, and non-maximum suppression keeps the selected windows spread across both sequences.
Up to $K{=}100$ windows are kept for each sequence pair, sampled to balance coverage across different overlap ranges rather than selecting purely by rank.
A minimum overlap threshold filters out pairs with insufficient shared content.
During training, this threshold starts at $0.5$ and decreases to $0.1$ following the curriculum described in Sec.~\ref{app:impl:curriculum}.
Fig.~\ref{fig:appA_overlap_matrix} shows two real mined overlap matrices together with a subset of the selected windows for illustration.

\paragraph{Role of depth.}
Depth maps participate exclusively in this offline pair-mining step.
They are never seen by the model and never enter the loss.
Equivalent training pairs can also be constructed from camera poses and field-of-view geometry alone, so the dependency on depth is a property of the data pipeline rather than of the method.
For fair comparison, trainable baselines are retrained on the same datasets with their own original supervision, including dense or sparse depth where applicable.

% =============================================================================
% Appendix C: Covisibility-Based Window Selection for Deployment
% 撰写依据: docs/superpowers/specs/2026-06-01-corl-appendix-spec.md  (Task C)
% 状态: 完成。独立的部署期前端模块(非 "stage")：动机、自研 joint-attention 共视解码器
%       (复用 MA-DINOv2 + 密集监督优化标量预测)、推理选窗、评估、与 GT 选窗等价、推理速度。
%       全文不使用 "Stage 1 / Stage 2" 措辞，亦不依赖任何外部共视方法。
% =============================================================================
\section{Covisibility-Based Window Selection for Deployment}
\label{app:stage1}

\ours{} takes two compact observation groups and estimates their relative pose in a single forward pass.
The main paper evaluates it with the two observation windows given in advance.
In a real deployment, the two groups must instead be selected from continuous video streams or long trajectory logs.
This section describes an optional, self-contained front-end that performs this selection by predicting how much visual content candidate frame windows share.
The module is independent of \ours{} and is needed only when the input windows are not provided in advance.
The pair-mining pipeline of Sec.~\ref{app:data:pipeline} derives this overlap from depth offline; at deployment the module predicts it from RGB alone, so window selection needs neither depth nor camera poses. Pose estimation then uses the selected images and their known intra-group geometry.
Fig.~\ref{fig:appA_stage1_model} illustrates the covisibility prediction model and the resulting overlap matrix.

% -----------------------------------------------------------------------------
\subsection{Motivation}
\label{app:stage1:motivation}

Consider a relocalization query: a robot holds a short segment of its current observations (group~$B$) and must find the best-matching segment in a previously recorded map trajectory (group~$A$).
The map trajectory may contain thousands of frames, yet \ours{} expects a compact window of $W{=}5$ frames per group.
The window-selection module reduces the search space by scoring every candidate window pair according to its predicted visual overlap, so that \ours{} receives only the most informative pair.

A naive alternative would run \ours{} on every candidate pair and rank by prediction confidence.
This is prohibitively expensive, because \ours{} runs the full frozen backbone and the three trainable modules for each pair.
Our predictor instead emits a scalar overlap score per window pair without running the pose-estimation pipeline, at a cost dominated by a single shared encoder pass per frame (Sec.~\ref{app:stage1:retrieval}).

% -----------------------------------------------------------------------------
\subsection{Covisibility Prediction}
\label{app:stage1:model}

\begin{figure}[t]
  \centering
  \includegraphics[width=\linewidth]{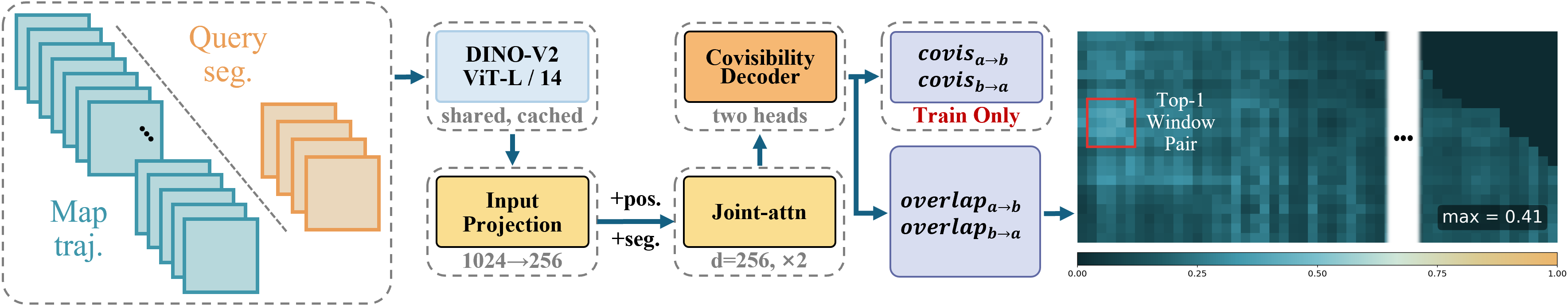}
  \caption{\textbf{Covisibility prediction model.}
  A frozen DINOv2 ViT-L/14 encoder extracts and caches per-frame patch features from the query segment and map trajectory.
  A linear projection reduces the $1024$-dim tokens to $d{=}256$; positional and segment embeddings are added before two layers of joint self-attention.
  Two heads produce dense per-pixel covisibility maps (used only during training) and scalar directional overlap scores.
  The scalar scores populate a pairwise overlap matrix (\emph{right}); at deployment a sliding-window search selects the top-1 window pair (red box) for \ours{}.}
  \label{fig:appA_stage1_model}
\end{figure}

We design a lightweight covisibility predictor that shares the same MapAnything-pretrained DINOv2 encoder that \ours{} keeps frozen, followed by a small trainable decoder (Fig.~\ref{fig:appA_stage1_model}).
Given two frames $I_a$ and $I_b$, the shared encoder produces patch-level features for each frame independently.
A linear projection reduces the $1024$-dim encoder tokens to $d{=}256$. The decoder then concatenates the projected patch tokens from both frames into a single sequence, adds a learned positional embedding and a segment embedding to distinguish the two frames, and applies two layers of joint self-attention.
Reusing the frozen encoder lets per-frame features be computed once and shared with \ours{}, keeping the added cost of this module small.

The decoder has two heads.
A \emph{dense} head predicts, for each pixel, whether it belongs to the covisible region with the other frame.
A \emph{scalar} head directly regresses the two directional covisibility ratios $r_{a \to b}$ and $r_{b \to a}$, where $r_{a \to b}$ is the fraction of pixels in frame~$a$ predicted covisible in frame~$b$.
Their minimum $o(a,b) = \min(r_{a \to b}, r_{b \to a})$ is the same symmetric overlap defined in Sec.~\ref{app:impl:curriculum} and serves as the retrieval score.
The dense per-pixel supervision is the key training signal.
It grounds the scalar prediction in \emph{where} the two views actually overlap, which we find necessary for accurate scalar scores from such a compact decoder.

\paragraph{Training.}
The decoder is trained on the same HM3D scenes used to train \ours{}, with ground-truth covisibility labels derived from the depth-based cross-projection described in Sec.~\ref{app:data:pipeline}.
The loss combines a symmetric per-pixel binary cross-entropy on the dense head with a direct regression on the scalar overlap, in both directions.
The encoder weights remain frozen throughout training, so only the covisibility decoder is updated.

% -----------------------------------------------------------------------------
\subsection{Window Retrieval at Deployment}
\label{app:stage1:retrieval}

At deployment, the module scores candidate window pairs and returns the best match for \ours{}.
The right portion of Fig.~\ref{fig:appA_stage1_model} shows the resulting overlap matrix; the retrieval procedure consists of three steps.

\paragraph{Step 1: Pairwise overlap matrix.}
For a pair of trajectories with $T_A$ and $T_B$ frames, the module evaluates all $T_A \times T_B$ frame pairs into a dense overlap matrix $\mathbf{S} \in [0, 1]^{T_A \times T_B}$ whose entry $(i,j)$ is the symmetric overlap $o(i,j)$ defined above.
The DINOv2 encoder, the most expensive component, runs once per frame and its features are cached. The backbone cost is therefore \emph{linear}, $O(T_A{+}T_B)$, rather than quadratic in the number of pairs.
The decoder consumes cached features only and scores the full matrix in a single batched pass, as benchmarked in Sec.~\ref{app:stage1:eval}.

\paragraph{Step 2: Sliding-window aggregation.}
A sliding window of size $W$ sweeps over both dimensions of $\mathbf{S}$. Each window is scored by the same \emph{max-mean} aggregation introduced in Sec.~\ref{app:impl:curriculum} for training-pair construction.
This aggregation favors regions where every frame finds a well-matching counterpart rather than isolated single-frame matches.

\paragraph{Step 3: Top-$K$ selection.}
The window scores are sorted in descending order, and the top-$K$ non-overlapping windows are retained.
During training-pair construction, up to $K{=}100$ windows are kept per sequence pair (Sec.~\ref{app:data:pipeline}).
At test time, the default is $K{=}1$: the single highest-scoring window pair is forwarded to \ours{}.
Windows whose predicted overlap falls below a threshold are discarded before running \ours{}.

% -----------------------------------------------------------------------------
\subsection{Prediction Quality and Inference Speed}
\label{app:stage1:eval}

Table~\ref{tab:appA_stage1} reports a systematic sweep over decoder depths 
and widths on the HM3D validation set.
Increasing decoder depth from one 
to two layers yields a substantial accuracy improvement, while further 
increasing width beyond $d{=}256$ provides diminishing gains at 
substantially higher parameter counts.
The two-layer, $d{=}256$ 
configuration with $1.84$\,M parameters is the most compact design that reliably stays below the MAE threshold of $0.05$, and we consider it a well-balanced configuration for deployment.
We additionally evaluate downstream frame retrieval for the adopted configuration on HM3D validation trajectory pairs.
The predicted overlap matrix places the ground-truth most-covisible frame in its top-$1$ prediction $64\%$ of the time and within its top-$5$ candidates $94\%$ of the time.
Accordingly, we observe no significant difference in \ours{} pose accuracy between windows selected from the predicted overlap matrix and windows selected from the ground-truth overlap matrix.
The module thus recovers windows of equivalent quality from RGB alone, without requiring the depth maps or known poses that ground-truth overlap computation demands.

\begin{table}[h]
  \centering
  \caption{\textbf{Covisibility decoder sweep on HM3D validation.}
  IoU and accuracy are evaluated on per-pixel covisibility maps; MAE on the scalar overlap score.
  Throughput is measured on a single RTX~4090 in bfloat16 scoring a $300{\times}300$ overlap matrix ($90{,}000$ frame pairs).
  The \textbf{bold} row is the adopted configuration, trained on all $800$ scenes.}
  \label{tab:appA_stage1}
  \vspace{0.4em}
  \begin{tabular}{lrcccr}
    \toprule
    Decoder & Params\,(M) & IoU & Acc. & MAE & Pairs/s \\
    \midrule
    2-layer, $d{=}128$  & 0.73  & 0.702 & 0.887 & 0.051 & 44{,}435 \\
    \textbf{2-layer, $\boldsymbol{d{=}256}$}  & \textbf{1.84}  & \textbf{0.822} & \textbf{0.930} & \textbf{0.030} & \textbf{23{,}771} \\
    2-layer, $d{=}384$  & 3.49  & 0.793 & 0.931 & 0.028 & 14{,}191 \\
    2-layer, $d{=}512$  & 5.65  & 0.805 & 0.934 & 0.027 & 10{,}234 \\
    1-layer, $d{=}512$  & 3.55  & 0.683 & 0.878 & 0.049 & 17{,}728 \\
    \bottomrule
  \end{tabular}
\end{table}

\paragraph{Inference speed.}
Consider a practical deployment scenario: a $T_B{=}30$-frame query segment must be localized against a $T_A{=}3{,}000$-frame map trajectory, producing $90{,}000$ candidate frame pairs.
Extracting DINOv2 features for all $3{,}030$ frames takes approximately $5$\,s at a throughput of $588$ frames/s on a single RTX~4090.
These features are shared with \ours{} and cached, so scoring all candidate pairs then reduces to a single batched decoder pass.
The adopted two-layer, $d{=}256$ decoder completes this pass in under $4$\,s (Table~\ref{tab:appA_stage1}).
The total front-end time from raw frames to a selected window pair is therefore approximately $9$\,s.
Window selection thus dominates the runtime in this scenario, compared with the ${\sim}50$\,ms that \ours{} subsequently spends on the selected window pair.

% =============================================================================
% Appendix D: Baseline Methods and Evaluation Protocols
% 撰写依据: docs/superpowers/specs/2026-06-01-corl-appendix-spec.md  (Task D)
% 状态: 完成。逐方法介绍 + 尺度口径 + pretrained 四数据集对比表 + 两个受控诊断。
% 来源核对: Rig3R_paper.md, LoMa/stage2/{README,CLAUDE}.md,
%           CoViS-Net/hm3d_bev_baseline/README.md, reloc3r/eval_{relpose,seq2seq}.py,
%           README_evaluation_methods.md, table1_reloc.tex, table2_rig.tex,
%           table3_pretrained.md, rig_table_main.md,
%           README_{vggt_nclt,rig3r_tg_overfit}_diagnosis.md
% 注意: \label{app:baselines:vggt} 与 \label{app:baselines:rig3r} 被 E_metrics.tex /
%        A_implementation.tex 交叉引用，重写时必须保留。
% =============================================================================
\section{Baseline Methods and Evaluation Protocols}
\label{app:baselines}

This section details the baselines compared in the main paper, their evaluation protocols, and two controlled diagnostic experiments.
The diagnostics account for the largest error entries in Tables~1 and~2 of the main paper, and show that each entry reflects a property of the evaluation setting or of a baseline's training configuration rather than an artifact of the evaluation pipeline.

% -----------------------------------------------------------------------------
\subsection{Baseline Configurations}
\label{app:baselines:methods}

Table~\ref{tab:appA_baselines} summarizes how each method ingests geometry, the scale of its output and how that scale is recovered, and how it is trained on the target data.
Every learning-based baseline is trained on each target dataset with its full original supervision, including dense or sparse depth where the method uses it, and starts from public pretrained checkpoints whenever the authors release them.
\ours{} instead uses only RGB images and intra-group relative poses, and therefore competes under strictly weaker supervision.
A recurring distinction in the table is \emph{where} the intra-group extrinsics enter a method.
\ours{}, Rig3R, and the MapAnything variants inject them as a conditioning signal inside the forward pass, while LoMa and Reloc3R consult them only at the post-inference geometric stage, and CoViS-Net and VGGT do not receive them at all.

\begin{table}[h]
  \vspace{-0.4cm}
  \centering
  \caption{\textbf{Baseline summary.}
  All methods take RGB images as their only visual input.
  ``Extrinsics'' indicates whether a method receives the intra-group extrinsics and where they enter:
  \emph{cond.}\ denotes a conditioning signal inside the forward pass, and \emph{post}\ denotes use only at the post-inference geometric or aggregation stage.
  ``Scale'' indicates a metric output (directly comparable $t$) or an up-to-scale output (requires alignment).
  MA-AB is an oracle that additionally receives the ground-truth inter-group transform.}
  \label{tab:appA_baselines}
  \vspace{0.4em}
  \small
  \begin{tabular}{l c c l l}
    \toprule
    Method & Extrinsics & Scale & Scale recovery & Target-domain training \\
    \midrule
    \ours{} (Ours)   & $A,B$ (cond.)         & metric      & direct prediction         & 32M modules trained \\
    LoMa             & $A,B$ (post)          & metric      & ray intersection          & matcher pretrained \\
    CoViS-Net        & none                  & metric      & direct regression         & trained from scratch \\
    VGGT             & none                  & up-to-scale & raw\,/\,$\mathrm{Sim}(3)^{\S}$ & finetuned \\
    Reloc3R          & $A,B$ (post)          & metric      & extrinsics aggregation    & finetuned \\
    Rig3R            & $A,B$ (cond.)         & metric      & ray map, closed-form      & DUSt3R-init, finetuned \\
    MA-A             & $A$ (cond.)           & metric      & backbone pose head        & frozen, not trained \\
    MA-AB (oracle)   & $A,B$ (cond.)         & metric      & backbone pose head        & frozen, not trained \\
    \bottomrule
  \end{tabular}

  \vspace{2pt}
  \parbox{\linewidth}{\footnotesize\raggedright%
  $^{\S}$\,VGGT predicts up-to-scale geometry: its localization translation (Table~1 of the main paper) is reported raw without alignment, while its rig translation (Table~2) is reported after a per-pair optimal $\mathrm{Sim}(3)$ alignment to the ground truth.
  ``DUSt3R-init, finetuned'' for Rig3R refers to its decoder initialization and training; the metadata modules and prediction heads are also trained, while the DINOv2 encoder remains frozen.}
  \vspace{-0.4cm}
\end{table}

\paragraph{LoMa.}
LoMa~\cite{nordstrom2026loma} is a state-of-the-art learned local feature matcher.
In the classical sparse-matching and pose-solving paradigm, the matching stage is the dominant source of catastrophic failure and gross error.
To measure how far this paradigm can reach once its weakest link is upgraded, we leave the classical pipeline intact and replace only its matching front-end with LoMa, keeping the geometric solver free of learnable parameters.
For two groups of $W{=}5$ frames, each of the $W^2{=}25$ cross-group frame pairs is matched, and a per-pair essential matrix yields a relative rotation together with an up-to-scale translation direction.
The known metric intra-group extrinsics then express the $25$ pairwise estimates as metric translation rays in the anchor frame, whose closed-form intersection, robustified by RANSAC, recovers the metric inter-group translation, while a Weiszfeld geometric median aggregates the rotations.
Like \ours{}, LoMa reads only RGB images and intra-group extrinsics and never accesses depth.
Because the matcher is pretrained and the solver carries no learnable parameters, this baseline is not retrained per dataset.

\paragraph{CoViS-Net.}
CoViS-Net~\cite{blumenkamp2024covisnet} casts group-to-group estimation as inference over a fully connected graph whose ten nodes are the five frames of each group.
Per-node DINOv2 features are exchanged along the graph edges, and an edge-convolution head regresses a relative pose with predicted uncertainty for every directed edge; the inter-group pose is read from the cross-group edges.
Following the original design, an auxiliary branch additionally predicts a per-group bird's-eye-view occupancy grid at a fixed metric resolution, which anchors the metric scale during training.
CoViS-Net receives no extrinsic input and is trained from scratch on each target dataset.

\paragraph{VGGT.}
VGGT~\cite{wang2025vggt} is a feed-forward multi-view transformer.
The ten frames of the two groups are presented as a single unstructured set, and the model predicts a world-frame pose for every frame, from which the inter-group pose is read.
VGGT takes no extrinsic input and predicts geometry only up to an unknown global scale.
For the localization task (Table~1 of the main paper), we report VGGT's raw translation error without scale alignment, so its $t$ values are larger than those of the metric-scale methods and are not directly comparable, while its rotation and the scale-free RTA remain comparable.
For the rig task (Table~2), a per-pair optimal $\mathrm{Sim}(3)$ alignment is applied so that VGGT's translation column stays comparable, as marked by the $^{\S}$ symbol in that table.
VGGT is evaluated both from its official pretrained checkpoint and after finetuning on each target dataset.

\paragraph{Reloc3R.}
Reloc3R~\cite{dong2025reloc3r} is a relative camera pose regression model.
Each cross-group image pair is processed independently, and the network regresses a relative rotation together with an up-to-scale translation direction.
A consistent, metric inter-group pose is then assembled by aggregating the $25$ pairwise predictions against the known intra-group extrinsics, which also fix the global scale.
The intra-group extrinsics therefore participate only at this post-inference aggregation stage, not as a conditioning signal inside the forward pass.
This distinction becomes important on the cross-season setting analyzed in Sec.~\ref{app:baselines:vggt}.
Reloc3R is evaluated both from its official pretrained checkpoint and after finetuning on each target dataset.

\paragraph{Rig3R.}
Rig3R~\cite{li2026rig3r} extends the DUSt3R family with rig-aware metadata conditioning, and like \ours{} takes the intra-group extrinsics as a forward-pass input.
Our implementation uses precomputed features from a frozen DINOv2 image encoder and a twelve-layer global self-attention decoder with $85.05$M parameters. The decoder is initialized from the self-attention and MLP weights of a DUSt3R checkpoint, omitting its pairwise cross-attention layers.
Before entering the decoder, each frame's image feature is augmented with rig metadata: a camera identifier, a timestamp, and a rig-relative ray map that encodes the camera's pose within the rig.
The decoder then fuses all frames and predicts per-frame pose ray maps, from which metric poses are recovered in closed form.
The decoder, metadata modules, input projection, and prediction heads remain trainable during target-domain training, totaling $88.65$M parameters; the image encoder is not updated. Rig metadata enters after image feature extraction and passes through the trainable decoder.
The consequences of this design under limited training data are examined in Sec.~\ref{app:baselines:rig3r}.
Since no public implementation is available, we reproduce Rig3R following the original paper.

\paragraph{MA-A and MA-AB.}
These two variants reuse the exact frozen MapAnything backbone of \ours{} and read poses from MapAnything's built-in pose head, providing reference conditions for its native camera-prior interface.
MapAnything injects extrinsics relative to a single frame that is fixed as the world origin, so it can natively accept the calibration of only one group.
MA-A uses this native mode: it is given group~$A$'s intra-group extrinsics anchored at $A_0$, while group~$B$ receives no extrinsic injection, so the inter-group relationship must be inferred from visual features alone.
MA-A differs from \ours{} in both the available conditioning and the prediction architecture; Sec.~\ref{app:ablation:conditioning} separately tests cross-group attention with the geometric inputs held fixed.
MA-AB instead uses the ground-truth inter-group transform to express group~$B$'s extrinsics in the $A_0$ frame and injects both groups, simulating a setting in which the inter-group pose is supplied rather than estimated.
MA-AB is an oracle reference with access to the target transform. Its output still depends on the frozen prediction head, so it is not a mathematical bound on the error of another estimator.
Both variants are evaluated only with clean ground-truth extrinsics and have no noisy variant.

% -----------------------------------------------------------------------------
\subsection{Scale and Metric Conventions}
\label{app:baselines:scale}

The main tables report mean translation error $t$ in meters and mean rotation error $r$ in degrees.
We use the mean rather than the median because the median can hide the heavy-tailed failures that determine reliability in deployment.
The translation-direction error RTA is computed as
\begin{equation}
  \text{RTA} = \arccos \frac{\hat{\mathbf{t}} \cdot \mathbf{t}_\text{gt}}{\|\hat{\mathbf{t}}\|\,\|\mathbf{t}_\text{gt}\|},
  \label{eq:appA_rta}
\end{equation}
which measures the angular error of the predicted translation direction without being affected by the inter-group distance.
RTA and RRA are scale-free and comparable across all methods, including VGGT, whose raw $t$ is not.
The localization evaluation uses $22{,}255$ pairs on HM3D, $26{,}872$ on TartanGround, $22{,}632$ on NCLT, and about $3{,}000$ pairs on each of the ZJH simulated and real splits, with all methods sharing the same sampled pairs per dataset for a fair comparison.

% -----------------------------------------------------------------------------
\subsection{Pretrained versus Finetuned Baselines}
\label{app:baselines:pretrained}

The localization tables finetune VGGT and Reloc3R on each target dataset.
To confirm that this is the stronger and fairer setting rather than one that disadvantages the baselines, we additionally evaluate their official pretrained checkpoints without any target-domain training.
Table~\ref{tab:appA_pretrained} reports both settings across all four datasets.

\begin{table}[h]
  \vspace{-0.2cm}
  \centering
  \caption{\textbf{Pretrained versus finetuned baselines on cross-sequence relocalization.}
  Each cell reports mean translation error $t$ (m) and mean rotation error $r$ ($^\circ$); lower is better.
  Pretrained rows use the official checkpoints with no target-domain training; finetuned rows follow each method's published protocol and match the corresponding entries in Table~1 of the main paper.
  \ours{} is shown for reference.
  VGGT predicts up-to-scale geometry, so its $t$ is the raw error without scale alignment and is not comparable across scale conventions, whereas the scale-free $r$ is comparable throughout.
  ZJH~{\scriptsize(S/R)} reports sim/real per cell.}
  \label{tab:appA_pretrained}
  \vspace{0.4em}
  \footnotesize
  \setlength{\tabcolsep}{4pt}
  \begin{tabular}{l cc cc cc cc}
    \toprule
    & \multicolumn{2}{c}{\textbf{HM3D}} & \multicolumn{2}{c}{\textbf{TartanGround}} & \multicolumn{2}{c}{\textbf{NCLT}} & \multicolumn{2}{c}{\textbf{ZJH {\scriptsize(S/R)}}} \\
    \cmidrule(lr){2-3} \cmidrule(lr){4-5} \cmidrule(lr){6-7} \cmidrule(lr){8-9}
    Method & $t${\scriptsize(m)} & $r${\scriptsize($^\circ$)} & $t${\scriptsize(m)} & $r${\scriptsize($^\circ$)} & $t${\scriptsize(m)} & $r${\scriptsize($^\circ$)} & $t${\scriptsize(m)} & $r${\scriptsize($^\circ$)} \\
    \midrule
    VGGT (pretrained)    & 1.532 & 15.80 & 2.916 & 16.92 & 8.829 & 54.44 & 3.009/4.462 & 13.46/13.66 \\
    VGGT (finetuned)     & 1.500 &  8.18 & 3.006 & 16.06 & 8.775 & 43.87 & 3.047/4.497 & 13.09/13.73 \\
    Reloc3R (pretrained) & 1.234 &  9.13 & 2.723 & 22.73 & 8.669 & 33.79 & 2.028/3.436 &  5.59/14.16 \\
    Reloc3R (finetuned)  & 0.725 &  2.66 & 1.649 & 14.88 & 6.918 &  6.34 & 1.701/2.988 &  4.18/ 7.75 \\
    \midrule
    \ours{} (reference)  & \best{0.155} & \best{1.72} & \best{0.526} & \best{5.39} & \best{0.692} & \best{2.47} & \best{0.305/1.206} & \best{1.67/3.46} \\
    \bottomrule
  \end{tabular}
\end{table}

Finetuning consistently improves rotation accuracy on the datasets whose visual domain differs from the pretraining corpus.
Reloc3R's mean rotation drops from $9.13^\circ$ to $2.66^\circ$ on HM3D and from $22.73^\circ$ to $14.88^\circ$ on TartanGround, and VGGT improves from $15.80^\circ$ to $8.18^\circ$ on HM3D.
VGGT gains less than Reloc3R because it accepts no extrinsic input, so additional target-domain visual features are its only avenue for adaptation.
On the self-collected ZJH split, where finetuning is performed only on the simulated trajectories and the real captures are held out for zero-shot evaluation, Reloc3R's mean rotation drops from $5.59^\circ$ to $4.18^\circ$ on the simulated column and from $14.16^\circ$ to $7.75^\circ$ on the real column, while VGGT's rotation moves only marginally on either column.
NCLT behaves differently from the other splits.
There the cross-season setting keeps both methods far from metric accuracy regardless of training.
VGGT's rotation stays above $40^\circ$ in both rows, and although Reloc3R's rotation improves to $6.34^\circ$ after finetuning, its translation direction remains far from accurate, with an RTA of $48.5^\circ$ in Table~1 of the main paper.
This pattern indicates that the NCLT degradation stems from cross-group visual matching under seasonal change rather than from the use of pretrained weights, and it is analyzed in detail in Sec.~\ref{app:baselines:vggt}.
Overall, evaluating only pretrained checkpoints would understate the baselines and overstate the gap to \ours{}, so the main tables report the finetuned models throughout.

% -----------------------------------------------------------------------------
\subsection{Controlled Diagnostic: VGGT on Cross-Season NCLT}
\label{app:baselines:vggt}

NCLT evaluates cross-season relocalization, where group~$A$ and group~$B$ image the same location under different seasons and illumination.
In Table~1 of the main paper, VGGT reaches an RTA of $82.1^\circ$ on NCLT, an essentially random translation direction, together with a $43.9^\circ$ rotation error, far above the $8.2^\circ$ it attains on HM3D.
We designed a controlled experiment to locate the cause and to confirm that it is not an evaluation artifact.

\paragraph{VGGT operates correctly on temporally consistent NCLT video.}
Given a single group of five same-session frames, VGGT estimates the intra-group pose $A_0 \to A_4$ on NCLT with a median rotation error of $2.3^\circ$, on par with HM3D ($2.4^\circ$) and TartanGround ($2.8^\circ$).
The failure is therefore specific to the cross-group, cross-season pairing rather than to the NCLT imagery in general.

\paragraph{A single cross-season frame collapses the cross-group pose.}
We form a six-frame batch from the five same-season frames of group~$A$ plus the single group-$B$ frame with the highest geometric overlap with group~$A$, and we read both the intra-group pose ($A_0 \to A_4$) and the cross-group pose ($A_0 \to b$) from the same inference pass.
On three NCLT pairs the intra-group rotation errors are $3.3^\circ$, $2.1^\circ$, and $1.2^\circ$, while the cross-group errors are $78.0^\circ$, $93.2^\circ$, and $77.1^\circ$.
On HM3D and TartanGround control pairs both errors stay below $5^\circ$.
Because the five same-season frames remain accurate within the very same forward pass, this rules out batch-level explanations such as coordinate conventions, input construction, and preprocessing.

\paragraph{Interpretation.}
VGGT establishes correspondence between frames through visual attention.
When a cross-season frame shares the same geometry but an entirely different appearance, no usable visual correspondence is found and the predicted pose degrades toward random, so geometric overlap measured from depth does not imply visual similarity.
The effect concentrates in the low-overlap regime: $51\%$ of the NCLT evaluation pairs have a covisibility window score below $0.2$, against no more than about $10\%$ on each simulated dataset, and even on the highest-overlap NCLT pairs (score above $0.6$) VGGT's median rotation error is $4.3^\circ$, nearly twice its HM3D value of $2.4^\circ$.
The same mechanism limits Reloc3R: it too must relate the two groups through their appearance, and the known extrinsics enter only afterward, where they cannot repair an estimate already corrupted by the cross-season gap, so its NCLT RTA stays at $48.5^\circ$ even after finetuning.
Both failures share a single cause, a reliance on visual appearance to relate the two groups, which motivates the geometry-grounded design of \ours{}.
Rather than matching the two groups visually, \ours{} has its frozen backbone encode the known intra-group geometry into each group's features while the trainable cross-group bridge reasons about the geometric relationship, so it holds RRA@$5^\circ$ at $95.2\%$ on NCLT.

% -----------------------------------------------------------------------------
\subsection{Controlled Diagnostic: Rig3R on TartanGround}
\label{app:baselines:rig3r}

In Table~2 of the main paper, Rig3R on the TartanGround 4-cam configuration shows mean errors of $12.6$\,m and $37.9^\circ$ with mAA of $4.2$, far from its $0.42$\,m and $5.54^\circ$ on the structurally identical HM3D 4-cam configuration.
A train-versus-validation probe reveals a substantial generalization gap.

\paragraph{The checkpoint overfits the training scenes.}
Evaluating the same checkpoint separately on the $53$ training scenes and the $10$ held-out validation scenes gives a median rotation of $2.0^\circ$ on the training scenes against $21.9^\circ$ on the held-out scenes, the latter reproducing the collapsed Table~2 entry.
The model reproduces the training poses but does not transfer to new scenes.
Earlier checkpoints do not help: validation accuracy is already saturated by the point at which the training error is lowest, so this is not an early-stopping issue.

\paragraph{Interpretation of the generalization gap.}
The frozen DINOv2 encoder preserves its pretrained image features, while the DUSt3R-initialized decoder and the metadata and prediction modules are adapted to the target dataset.
The observed gap is consistent with overfitting in these trainable modules, but does not establish collapse of the frozen encoder or isolate a particular architectural cause.
TartanGround provides 53 training scenes and 10 held-out scenes, whereas HM3D supplies 800 training scenes and NCLT revisits the same campus across dates.
These differences in scene diversity and train/test overlap may contribute to the differing generalization behavior.
A matched experiment varying the decoder's freezing policy would be needed to isolate the effect of decoder adaptation.

\paragraph{The pipeline is sound.}
\label{app:baselines:rig3r:freeze}
On the same validation scenes, \ours{} reaches $1.63^\circ$ and the MA-AB oracle reaches $0.61^\circ$.
The data and the evaluation pipeline are therefore sound, and the large entry reflects a training-configuration limitation of this reproduced baseline rather than its inherent capability.
Unlike this reproduced baseline's trainable cross-view decoder, \ours{} keeps the foundation backbone's geometric fusion and cross-view weights frozen throughout training.
The intra-group representation thus inherits the generalization of large-scale pretraining, and only the lightweight $32$M cross-group modules are learned on top of it.

% =============================================================================
% Appendix E: Scale-Free Metrics and Detailed Errors
% 撰写依据: docs/superpowers/specs/2026-06-01-corl-appendix-spec.md  (Task G → Section E)
% 状态: 完成。mAA rig 表 (双层表头, Ours 置底) + reloc 完整指标表 (含 MA-A/MA-AB, Ours 置底)
%       + per-overlap 两张四面板图 (均值旋转 + 均值平移, 全 7 方法含 Ours) + 分析。
%       两表由 tables/gen_appendix_tables.py 自动生成 (best/second 代码计算)，此处 \input；
%       两图由 figures/figspec_perf_vs_overlap.py 渲染 (rot/trans, 真实逐对数据, 统一选窗 overlap 轴)。
% =============================================================================
\section{Scale-Free Metrics and Detailed Errors}
\label{app:metrics}

The main tables report mean translation and rotation errors.
This section supplements them with scale-free and distributional metrics that provide a more complete picture.

% -----------------------------------------------------------------------------
\subsection{Rig Odometry: mAA Results}
\label{app:metrics:maa}

Table~\ref{tab:appA_maa} reports mAA@$30^\circ$ for the six rig configurations of Table~2, with the ZJH simulated and real rigs listed separately; the protocol is defined in the caption.
Because mAA couples rotation and translation direction within the same threshold, it is stricter than either RRA or RTA alone.
The corresponding mAA@$\tau$ curves for the two NCLT settings are shown in Fig.~\ref{fig:accuracy} of the main paper.

% 由 gen_appendix_tables.py 自动生成，请勿手动编辑数值/着色。
% 数据源: paper_results/rig_table_main.md (mAA 列)。
\begin{table}[h]
  \centering
  \vspace{-0.4cm}
  \caption{\textbf{Rig odometry mAA@$30^\circ$.}
  Following the relative-pose protocol adopted by recent multi-view estimators~\citep{wang2025vggt,dong2025reloc3r,li2026rig3r}, mAA@$30^\circ$ averages the pair accuracy over rotation and translation-direction thresholds swept from $1^\circ$ to $30^\circ$ in $1^\circ$ steps; a pair is counted as correct at threshold $\tau$ when both its rotation error and its scale-free translation-direction error fall below $\tau$. Higher is better.
  \textcolor{cBest}{Cyan} and \textcolor{cSecond}{orange} mark the best and second-best non-oracle result per column; the MA-AB oracle is excluded from the ranking.}
  \label{tab:appA_maa}
  \vspace{0.4em}
  \small
  \setlength{\tabcolsep}{4pt}
  \begin{tabular}{l cc c cc cc}
    \toprule
    & \multicolumn{2}{c}{HM3D} & TG & \multicolumn{2}{c}{NCLT} & \multicolumn{2}{c}{ZJH} \\
    \cmidrule(lr){2-3} \cmidrule(lr){4-4} \cmidrule(lr){5-6} \cmidrule(lr){7-8}
    Method & 8-cam & 4-cam & 4-cam & intra & cross & sim & real \\
    \midrule
    Rig3R & 54.9 & 42.5 & 4.2 & 10.0 & 13.4 & 26.7 & 20.7 \\
    VGGT & 44.5 & 62.5 & 57.6 & 72.2 & 21.0 & \second{75.8} & \second{66.8} \\
    Reloc3R & \second{64.4} & \second{67.0} & \second{72.9} & \second{83.6} & \second{66.1} & 74.3 & 56.6 \\
    \midrule
    MA-A & 45.3 & 41.2 & 49.9 & 56.2 & 33.3 & 75.8 & 66.1 \\
    \rowcolor{gray!8}
    MA-AB~{\scriptsize(oracle)} & 97.1 & 95.6 & 97.8 & 95.8 & 95.8 & 98.5 & 98.1 \\
    \midrule
    \textbf{\ours{}} & \best{75.4} & \best{67.4} & \best{88.5} & \best{91.6} & \best{82.2} & \best{89.1} & \best{77.3} \\
    \bottomrule
  \end{tabular}
\end{table}

\ours{} attains the highest non-oracle mAA on every rig configuration.
The margin widens on the settings that stress cross-group visual matching.
On the NCLT cross-session rig, \ours{} reaches $82.2$ while VGGT drops to $21.0$, and on TartanGround \ours{} reaches $88.5$ while Rig3R collapses to $4.2$.
Both failures are analyzed in Sec.~\ref{app:baselines:vggt} and Sec.~\ref{app:baselines:rig3r}: VGGT takes no extrinsic input and cannot register a cross-season frame to the rest of the group, while the fully trainable Rig3R decoder overfits the small TartanGround training set.

% -----------------------------------------------------------------------------
\subsection{Localization: Median and Extended Metrics}
\label{app:metrics:reloc}

Table~\ref{tab:appA_reloc_full} supplements Table~1 of the main paper with median errors and threshold-based accuracies.
The mean reported in the main paper is sensitive to a small number of gross failures, so further contrasting it with the median exposes the shape of each method's error distribution.

% 由 gen_appendix_tables.py 自动生成，请勿手动编辑数值/着色。
% 数据源: paper_results/table1_trained_clean.md (HM3D / TartanGround / NCLT / ZJH sim+real)。
\begin{table*}[t]
  \centering
  \caption{\textbf{Cross-sequence relocalization: full metric suite.}
  $t$ and $r$ are in meters and degrees; ``med'' and ``mean'' denote median and mean.
  RTA and RRA are reported at both $5^\circ$ and $15^\circ$ thresholds.
  \textcolor{cBest}{Cyan} and \textcolor{cSecond}{orange} mark the best and second-best value per column within each dataset.}
  \label{tab:appA_reloc_full}
  \vspace{0.4em}
  \footnotesize
  \setlength{\tabcolsep}{3.5pt}
  \begin{tabular}{ll cccc cccc}
    \toprule
    & & \multicolumn{4}{c}{Translation} & \multicolumn{4}{c}{Rotation} \\
    \cmidrule(lr){3-6} \cmidrule(lr){7-10}
    Dataset & Method & $t_\text{med}$ & $t_\text{mean}$ & RTA@5$^\circ$ & RTA@15$^\circ$
                     & $r_\text{med}$ & $r_\text{mean}$ & RRA@5$^\circ$ & RRA@15$^\circ$ \\
    \midrule
    \multirow{7}{*}{HM3D} & LoMa & \best{0.069} & 0.802 & \second{76.8} & 82.4 & \best{0.50} & 3.79 & 94.7 & 97.2 \\
     & CoViS-Net & 1.233 & 1.511 & 5.9 & 28.6 & 24.87 & 37.24 & 1.6 & 22.8 \\
     & VGGT & 1.126 & 1.500 & 71.1 & \second{88.6} & 1.85 & 8.18 & 78.8 & 89.0 \\
     & Reloc3R & 0.189 & \second{0.725} & 76.2 & 85.8 & \second{1.09} & \second{2.66} & \second{94.8} & \second{98.3} \\
     & MA-A & 0.740 & 1.008 & 26.4 & 68.3 & 6.36 & 10.08 & 38.5 & 86.1 \\
    \rowcolor{gray!8}
     & MA-AB~{\scriptsize(oracle)} & 0.066 & 0.087 & 99.1 & 100.0 & 0.84 & 1.00 & 99.7 & 100.0 \\
     & \textbf{\ours{}} & \second{0.103} & \best{0.155} & \best{81.4} & \best{97.2} & 1.35 & \best{1.72} & \best{97.7} & \best{99.8} \\
    \midrule
    \multirow{7}{*}{TG} & LoMa & \best{0.074} & \second{1.183} & \best{73.9} & \second{79.2} & \best{0.36} & 18.40 & \second{83.4} & \second{85.8} \\
     & CoViS-Net & 2.580 & 3.025 & 2.4 & 13.2 & 13.84 & 36.18 & 25.6 & 52.2 \\
     & VGGT & 2.459 & 3.006 & 38.8 & 64.6 & 2.70 & 16.06 & 64.8 & 83.2 \\
     & Reloc3R & 0.485 & 1.649 & 54.3 & 71.1 & \second{1.06} & 14.88 & 78.1 & 85.6 \\
     & MA-A & 0.800 & 1.793 & 31.1 & 68.0 & 4.32 & \second{13.58} & 55.2 & 85.4 \\
    \rowcolor{gray!8}
     & MA-AB~{\scriptsize(oracle)} & 0.059 & 0.081 & 98.1 & 99.2 & 0.49 & 0.59 & 99.9 & 100.0 \\
     & \textbf{\ours{}} & \second{0.176} & \best{0.526} & \second{72.3} & \best{90.7} & 1.22 & \best{5.39} & \best{85.7} & \best{94.1} \\
    \midrule
    \multirow{7}{*}{NCLT} & LoMa & 5.486 & 7.858 & 20.9 & 40.4 & 4.41 & 22.13 & 55.1 & 78.6 \\
     & CoViS-Net & 4.162 & \second{5.731} & 16.8 & 43.0 & 6.68 & 17.22 & 37.6 & 75.6 \\
     & VGGT & 6.752 & 8.775 & 7.5 & 19.3 & 27.53 & 43.87 & 18.4 & 39.4 \\
     & Reloc3R & 4.175 & 6.918 & \second{36.5} & \second{60.8} & \second{3.46} & \second{6.34} & \second{70.3} & \second{94.8} \\
     & MA-A & \second{4.062} & 7.021 & 14.5 & 41.4 & 9.42 & 31.69 & 25.9 & 64.8 \\
    \rowcolor{gray!8}
     & MA-AB~{\scriptsize(oracle)} & 0.138 & 0.208 & 98.1 & 99.6 & 0.71 & 0.82 & 99.9 & 100.0 \\
     & \textbf{\ours{}} & \best{0.434} & \best{0.692} & \best{77.4} & \best{93.7} & \best{2.08} & \best{2.47} & \best{95.2} & \best{99.7} \\
    \midrule
    \multirow{7}{*}{ZJH (sim)} & LoMa & 2.246 & 2.946 & 47.1 & 62.7 & \second{1.52} & 19.32 & 65.4 & 75.6 \\
     & CoViS-Net & 1.570 & 2.134 & 12.7 & 45.6 & 15.42 & 32.61 & 12.8 & 49.2 \\
     & VGGT & 2.615 & 3.047 & 65.4 & 81.3 & 2.91 & 13.09 & 72.8 & 87.7 \\
     & Reloc3R & 0.577 & 1.701 & \second{72.2} & 81.7 & 2.51 & \second{4.18} & \second{89.9} & \second{97.4} \\
     & MA-A & \second{0.358} & \second{0.556} & 67.5 & \second{88.3} & 2.76 & 4.68 & 79.9 & 97.2 \\
    \rowcolor{gray!8}
     & MA-AB~{\scriptsize(oracle)} & 0.051 & 0.061 & 94.1 & 96.1 & 0.67 & 0.89 & 100.0 & 100.0 \\
     & \textbf{\ours{}} & \best{0.186} & \best{0.305} & \best{78.2} & \best{90.9} & \best{1.33} & \best{1.67} & \best{96.5} & \best{99.8} \\
    \midrule
    \multirow{7}{*}{ZJH (real)} & LoMa & 1.867 & 3.443 & 49.0 & 66.8 & \best{1.54} & \second{5.75} & \second{86.1} & \second{94.2} \\
     & CoViS-Net & 2.347 & 3.413 & 11.2 & 37.8 & 54.34 & 59.35 & 6.4 & 16.6 \\
     & VGGT & 3.246 & 4.497 & 67.1 & 83.5 & 3.24 & 13.73 & 76.3 & 88.5 \\
     & Reloc3R & 1.096 & 2.988 & 59.2 & 74.4 & 3.00 & 7.75 & 85.0 & 94.0 \\
     & MA-A & \best{0.638} & \second{1.324} & \second{67.9} & \second{90.0} & 3.07 & 9.02 & 78.4 & 94.1 \\
    \rowcolor{gray!8}
     & MA-AB~{\scriptsize(oracle)} & 0.050 & 0.066 & 98.8 & 99.1 & 0.55 & 0.65 & 100.0 & 100.0 \\
     & \textbf{\ours{}} & \second{0.894} & \best{1.206} & \best{77.9} & \best{93.3} & \second{1.64} & \best{3.46} & \best{95.0} & \best{98.7} \\
    \bottomrule
  \end{tabular}
  \vspace{-0.4cm}
\end{table*}

On the simulated datasets, where rendering is geometrically exact and textures remain stable across viewpoints, LoMa's classical sparse matching attains the lowest median translation and rotation: $0.069$\,m / $0.50^\circ$ on HM3D and $0.074$\,m / $0.36^\circ$ on TartanGround, below the $0.103$\,m / $1.35^\circ$ and $0.176$\,m / $1.22^\circ$ of \ours{}.
Its mean translation on HM3D reaches $0.802$\,m, however, over eleven times the median, so a minority of catastrophic match failures dominates the average.
On NCLT, where cross-season appearance change undermines sparse correspondences, the classical advantage disappears: the LoMa median rotation rises to $4.41^\circ$ and its mean to $22.13^\circ$.
Both LoMa and Reloc3R further require processing all $N_A{\times}N_B$ cross-group frame pairs individually, raising LoMa's latency to $3{,}518$\,ms per group pair, whereas \ours{} resolves the group-to-group pose in a single forward pass of under $50$\,ms (Table~\ref{tab:appA_efficiency}).

% -----------------------------------------------------------------------------
\subsection{Performance versus Field-of-View Overlap}
\label{app:metrics:overlap}

\begin{figure*}[t]
  \vspace{-0.4cm}
  \centering
  \includegraphics[width=0.78\linewidth]{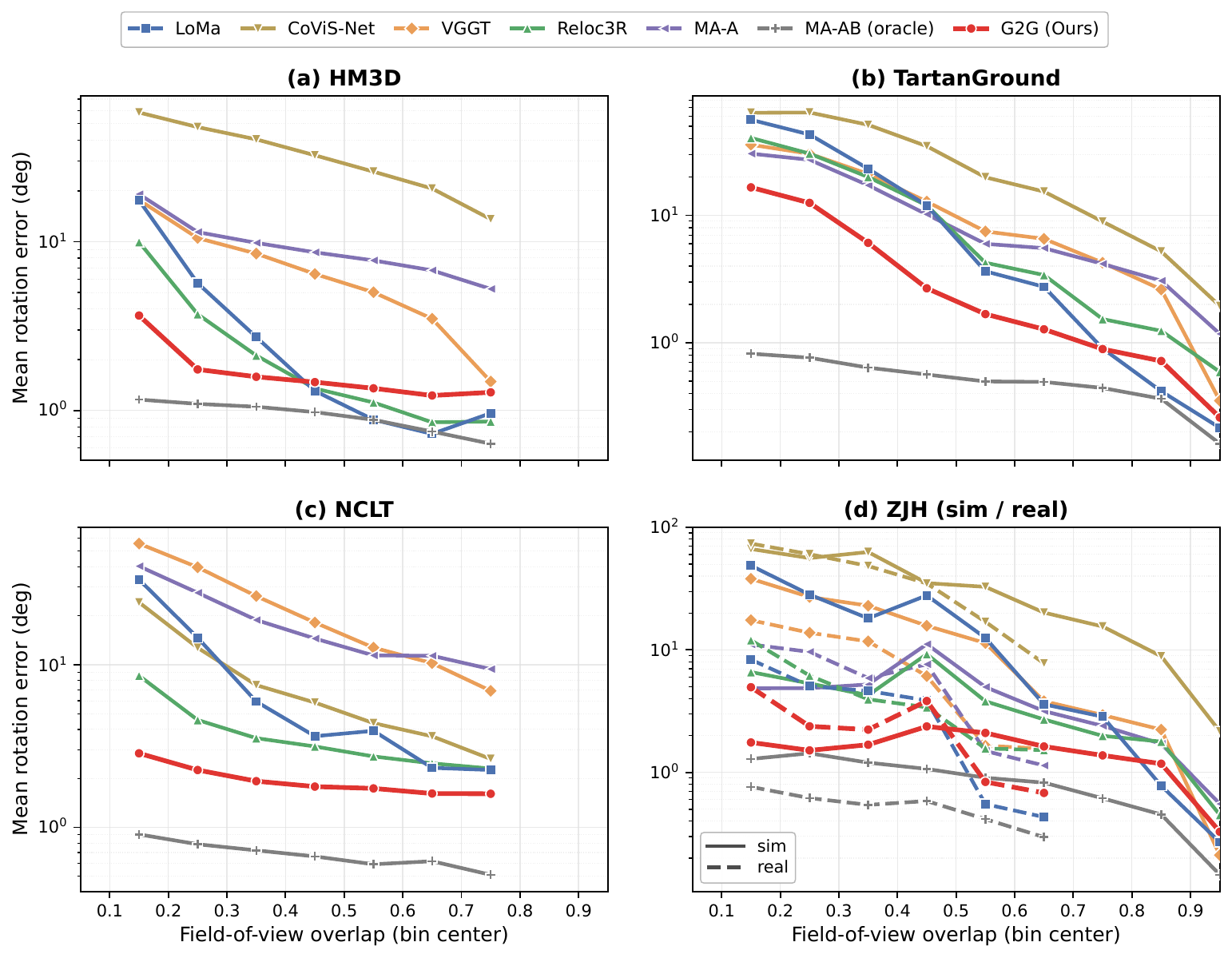}
  \vspace{-0.2cm}
  \caption{\textbf{Mean rotation error vs.\ field-of-view overlap.}
  Panels \emph{(a)}--\emph{(c)}: HM3D, TartanGround, NCLT; panel \emph{(d)}: ZJH sim (solid) and real (dashed).
  The MA-AB oracle (gray) serves as a reference with access to the ground-truth inter-group transform.
  \ours{} (red) degrades most gradually and stays closest to the oracle on the harder settings.}
  \label{fig:appA_perf_overlap_rot}
  \vspace{-0.4cm}
\end{figure*}

\begin{figure*}[t]
  \centering
  \includegraphics[width=0.78\linewidth]{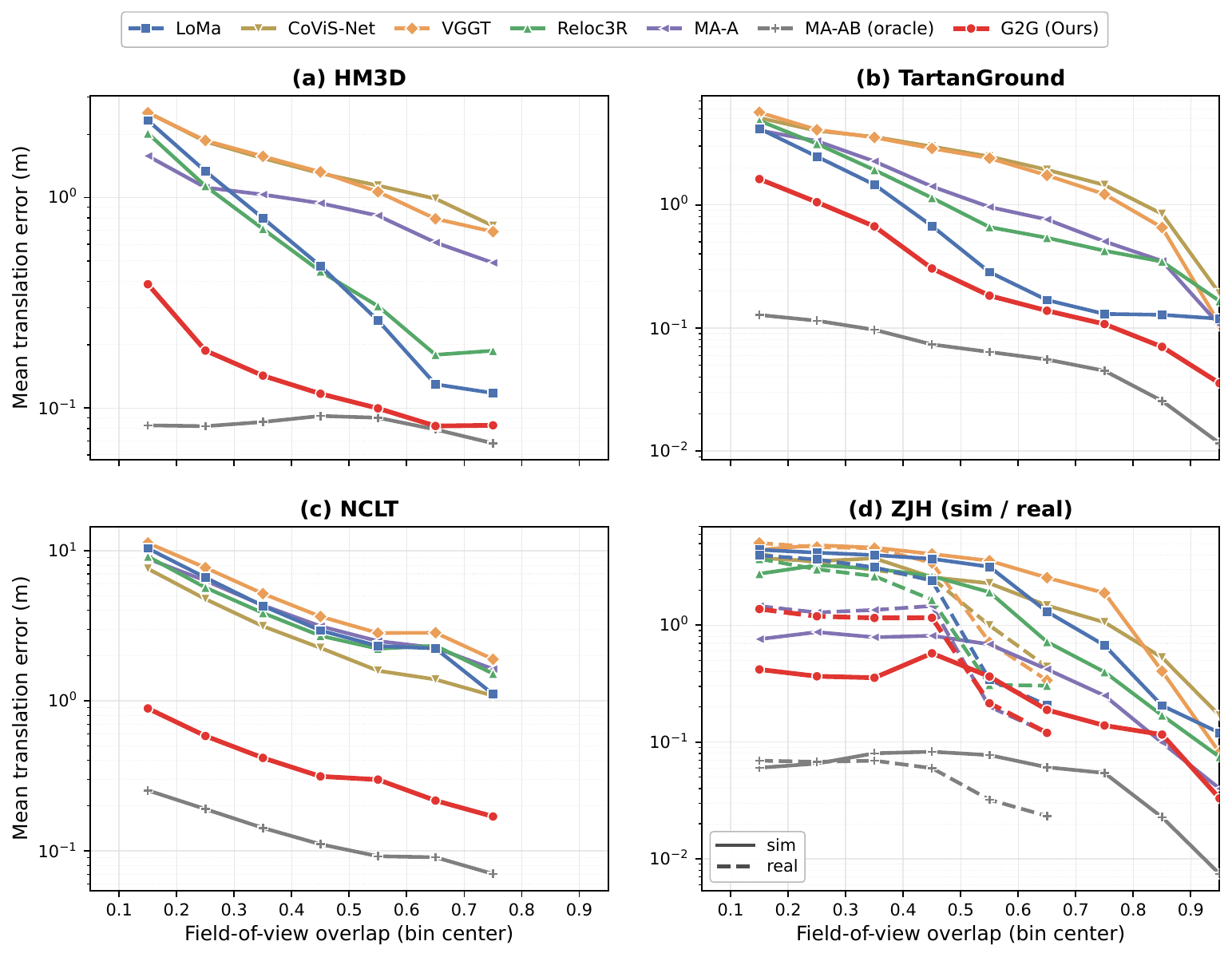}
  \vspace{-0.2cm}
  \caption{\textbf{Mean translation error vs.\ field-of-view overlap.}
  Same panels as Fig.~\ref{fig:appA_perf_overlap_rot}. VGGT reports up-to-scale translation; the remaining methods report metric translation.
  \ours{} tracks the MA-AB oracle closely, whereas baselines grow by an order of magnitude as overlap decreases.}
  \label{fig:appA_perf_overlap_trans}
  \vspace{-0.4cm}
\end{figure*}

The aggregate errors above average over pairs of widely varying difficulty.
Fig.~\ref{fig:appA_perf_overlap_rot} and Fig.~\ref{fig:appA_perf_overlap_trans} resolve accuracy as a function of the field-of-view overlap between the two groups, for rotation and translation respectively.
Evaluation pairs are binned by their overlap in steps of $0.1$, and the mean error is reported within each bin.
All methods are evaluated on the same test pairs, and the overlap bins are constructed from the ground-truth covisibility labels described in Sec.~\ref{app:data:pipeline}.

Every method degrades as overlap decreases, since fewer shared visual cues remain to relate the two groups, and the mean makes the differences in robustness explicit.
On the rotation axis (Fig.~\ref{fig:appA_perf_overlap_rot}), \ours{} is the lowest non-oracle method across almost the entire overlap range on every dataset, and the gap widens as overlap falls.
On HM3D the mean rotation of \ours{} rises only from $1.3^\circ$ at high overlap to $3.7^\circ$ at the lowest bin, whereas LoMa, VGGT, and MA-A rise to $17.6^\circ$, $17.7^\circ$, and $19.1^\circ$.
The contrast is sharper on NCLT, where \ours{} stays between $1.6^\circ$ and $2.9^\circ$ while VGGT and MA-A exceed $40^\circ$ at the lowest-overlap bin.
Methods that depend on pixel-level visual correspondences degrade sharply when seasonal change alters appearance (Sec.~\ref{app:baselines:vggt}).
MA-A shares the same frozen backbone but conditions only on group~$A$'s extrinsics, so group~$B$'s features lack geometric structure; predicting the inter-group pose from one geometrically coherent and one unstructured representation grows increasingly difficult as visual overlap diminishes.
\ours{} avoids both failure modes: the backbone independently fuses each group's intra-group extrinsics into geometry-enhanced features, so that both groups carry coherent 3D structure before the cross-group bridge aligns them (Sec.~3.2 of the main paper).

The translation axis (Fig.~\ref{fig:appA_perf_overlap_trans}) shows the same ordering and adds the effect of metric scale.
\ours{} tracks the MA-AB oracle closely, with a mean translation error as low as $0.08$\,m at high overlap on HM3D and $0.17$\,m on NCLT, against the oracle's $0.07$\,m.
The baselines remain an order of magnitude larger and reach roughly $8$ to $11$\,m at the lowest-overlap bin on NCLT.
The ZJH panels confirm consistent sim-to-real behavior: the simulated and real \ours{} curves stay close together under both metrics.
The advantage of \ours{} therefore lies not in uniformly lower errors on easy pairs but in graceful degradation as overlap diminishes.
Because both groups enter the cross-group bridge as geometrically coherent representations, the bridge can exploit the internal 3D structure of each group to constrain the relative pose even when the shared visual content between them is scarce.

% =============================================================================
% Appendix F: Detailed Ablation Analysis
% 撰写依据: docs/superpowers/specs/2026-06-01-corl-appendix-spec.md  (Task E + F → Section F)
% 状态: 重写。复现消融主表 + 逐设计选择分析(三嵌入/课程引用A章/噪声从动机出发) +
%       ZJH sim-to-real 异常深度分析 + 组件贡献排序(t+r,一致口径,代码校验) +
%       去掉 geometry conditioning + 噪声鲁棒大表(含组内去噪) + 跨方法噪声大表 +
%       效率分析(以训练显存为核心, A6000 实测复现, 真实 run 旁证)。
% 数据核验: /tmp/verify_ranking.py (排序), /tmp/bench_train_latents.py (效率),
%           paper_results/table{1,2,4}_*.md, eval_results_*/overall_summary.txt (A-intra)。
% =============================================================================
\section{Detailed Ablation Analysis}
\label{app:ablation}

Table~3 of the main paper isolates each design choice across five evaluation settings under clean extrinsics.
This section provides a detailed analysis of those results and extends the discussion to robustness to extrinsic noise, the ranking of component contributions, the accuracy-efficiency trade-off, intrinsic conditioning, cross-group attention, variable group sizes, and alternative frozen backbones.
Unless otherwise noted, all numbers below refer to the default configuration and the corresponding ablated variant, reported as mean errors under clean extrinsics.

% -----------------------------------------------------------------------------
\subsection{Per-Design-Choice Analysis}
\label{app:ablation:perchoice}

Table~\ref{tab:appA_ablation} reproduces the main ablation so that the analysis below is self-contained.
The default configuration is a cross-dataset compromise rather than the single-setting optimum, as the resampler discussion makes explicit.

% Reproduced ablation table (mean errors, clean extrinsics).
% 数值与正文 Table 3 (paper/tables/table3_ablation.tex) 完全一致, 仅 relabel 为 tab:appA_ablation
% 供 standalone 补充材料自包含引用。
\begin{table*}[t]
\centering
\caption{\textbf{Ablation of \ours{} design choices across five evaluation settings} (mean errors, clean extrinsics).
  This table is identical to Table~3 of the main paper, reproduced here so that the per-design-choice analysis below is self-contained.
  Each column pair removes one design choice from the default configuration.
  \textcolor{cBest}{Cyan} and \textcolor{cSecond}{orange} mark the best and second-best $t$ and $r$ within each row.}
\label{tab:appA_ablation}
\vspace{0.3em}
\resizebox{\textwidth}{!}{%
\begin{tabular}{l>{\columncolor{gray!8}}cc>{\columncolor{gray!8}}cc>{\columncolor{gray!8}}cc>{\columncolor{gray!8}}cc>{\columncolor{gray!8}}cc>{\columncolor{gray!8}}cc}
\toprule
  & \multicolumn{2}{c}{\textbf{Default}} & \multicolumn{2}{c}{\textbf{w/o Resamp.}} & \multicolumn{2}{c}{\textbf{w/o Anchor}} & \multicolumn{2}{c}{\textbf{w/o Curric.}} & \multicolumn{2}{c}{\textbf{w/o Noise Aug.}} & \multicolumn{2}{c}{\textbf{$3{+}3$ Win.}} \\
\cmidrule(lr){2-3} \cmidrule(lr){4-5} \cmidrule(lr){6-7} \cmidrule(lr){8-9} \cmidrule(lr){10-11} \cmidrule(lr){12-13}
Setting & $t${\scriptsize(m)}$\downarrow$ & $r${\scriptsize($^\circ$)}$\downarrow$ & $t${\scriptsize(m)}$\downarrow$ & $r${\scriptsize($^\circ$)}$\downarrow$ & $t${\scriptsize(m)}$\downarrow$ & $r${\scriptsize($^\circ$)}$\downarrow$ & $t${\scriptsize(m)}$\downarrow$ & $r${\scriptsize($^\circ$)}$\downarrow$ & $t${\scriptsize(m)}$\downarrow$ & $r${\scriptsize($^\circ$)}$\downarrow$ & $t${\scriptsize(m)}$\downarrow$ & $r${\scriptsize($^\circ$)}$\downarrow$ \\
\midrule
HM3D & \second{0.155} & \second{1.72} & \best{0.123} & \best{1.47} & 0.171 & 1.98 & 0.171 & 1.83 & 0.177 & 2.03 & 0.244 & 3.05 \\
TartanGround & \second{0.526} & \second{5.39} & \best{0.441} & \best{4.46} & 0.618 & 5.92 & 0.541 & 5.93 & 0.531 & 5.72 & 0.711 & 10.31 \\
NCLT & \second{0.692} & 2.47 & \best{0.597} & \best{2.29} & 0.793 & 2.64 & 0.742 & 2.50 & 0.744 & \second{2.40} & 0.900 & 3.05 \\
ZJH {\scriptsize(sim)} & \second{0.305} & \best{1.67} & \best{0.249} & \second{1.73} & 0.369 & 2.19 & 0.352 & 1.74 & 0.428 & 2.79 & 0.434 & 2.22 \\
ZJH {\scriptsize(real)} & 1.206 & 3.46 & \second{1.118} & \best{2.81} & 1.122 & \second{3.17} & 1.143 & 3.79 & \best{1.020} & 4.51 & 1.170 & 4.54 \\
\bottomrule
\end{tabular}%
}
\vspace{-0.3cm}
\end{table*}

\paragraph{Anchor embedding.}
The cross-group bridge organizes the merged token sequence with three learnable embeddings, defined in Eq.~\eqref{eq:bridge_token} and listed in Sec.~\ref{app:impl:params}.
A frame embedding encodes each frame's index within its group and is shared between the two groups.
A group embedding distinguishes group~$A$ tokens from group~$B$ tokens once the two sequences are concatenated.
An anchor embedding marks the $A_0$ tokens as the global reference frame.
The ``w/o Anchor'' variant removes the anchor embedding while keeping the other two.
This raises the HM3D mean errors from $0.155$\,m\,/\,$1.72^\circ$ to $0.171$\,m\,/\,$1.98^\circ$, with the same trend on TartanGround, NCLT, and simulated ZJH; the real ZJH split, on which removing any single component can lower the translation error, is analyzed in Sec.~\ref{app:ablation:ranking}.
Without it, the groups are still distinguished, but no token is marked as the fixed origin against which all poses are predicted.
The bridge then has to infer the relations among all frame pairs jointly, rather than focus on aligning each frame to a single designated reference.
The anchor embedding restores this focus by fixing $A_0$ as the common origin, so that every pose is predicted directly with respect to it.

\paragraph{Three-phase curriculum.}
Replacing the three-phase curriculum described in Sec.~\ref{app:impl:curriculum} with single-stage training, in which all overlap levels are sampled from the start, raises the HM3D errors to $0.171$\,m\,/\,$1.83^\circ$, with comparable degradation on TartanGround and NCLT.
Low-overlap pairs provide weak cross-group constraints, so exposing the model to them before it has learned from easier pairs destabilizes early optimization.
The curriculum defers these pairs until the model has converged on high-overlap pairs, which stabilizes training at a modest but consistent accuracy gain.

\paragraph{$\SE$ noise augmentation.}
The intra-group extrinsics available at deployment are estimates produced by visual odometry, SLAM, or SfM, and rig calibrations drift over time, so the inputs the model receives are never exact.
The model should therefore treat the input extrinsics as a noisy measurement to be refined rather than as ground truth to be copied.
We instill this behavior by perturbing the input extrinsics with $\SE$ noise during training (Sec.~\ref{app:impl:noise}), drawn from the error magnitude expected at deployment, while supervising against the clean poses.
This serves two purposes: it makes the intra-group predictions $\{\pose{A_0}{A_i}\}$ meaningful as corrections of the input, and it regularizes training by preventing the model from memorizing perfectly accurate inputs.
Removing the augmentation degrades accuracy even when the test extrinsics are clean, raising HM3D to $0.177$\,m\,/\,$2.03^\circ$.
The benefit is largest on the small ZJH simulated set and underlies the denoising behavior quantified in Sec.~\ref{app:ablation:noise}; its one adverse interaction, on the zero-shot ZJH real split, is analyzed next.

\paragraph{Window size $3{+}3$.}
Reducing the window from the default $5{+}5$ to $3{+}3$ at test time, without retraining, raises HM3D to $0.244$\,m\,/\,$3.05^\circ$, since a smaller window provides less intra-group context for the backbone to fuse.
The purpose of this variant is not peak accuracy but to show that a single trained model accepts a range of group sizes: the rig task already spans four to eight cameras per group, and the same architecture extends to larger windows to cover a wider range of source and target frame counts.
Because the window size is an inference-time setting rather than a training design choice, it is excluded from the component ranking in Sec.~\ref{app:ablation:ranking}.

\paragraph{Perceiver resampler.}
Removing the perceiver resampler, the ``w/o Resamp.'' variant, is slightly more accurate on HM3D ($0.123$\,m\,/\,$1.47^\circ$ versus the default $0.155$\,m\,/\,$1.72^\circ$).
This variant passes all $P{=}256$ patch tokens per frame into the cross-group bridge instead of $L{=}64$ latent tokens, which lengthens the bridge attention sequence by a factor of four.
The accuracy gain is marginal while the training cost is substantial, so the $64$-latent resampler is retained as the default.
Sec.~\ref{app:ablation:efficiency} quantifies this trade-off.

% -----------------------------------------------------------------------------
\subsection{The ZJH Real Anomaly}
\label{app:ablation:zjh}

The ZJH real split is the only setting evaluated entirely as zero-shot sim-to-real transfer, with the model trained on the simulated ZJH environments and tested on real captures without adaptation.
On this split the noise augmentation has opposite effects on translation and rotation.
Removing it lowers the mean translation error from $1.206$\,m to $1.020$\,m, yet raises the mean rotation error from $3.46^\circ$ to $4.51^\circ$.
On the simulated ZJH split the augmentation instead improves both translation and rotation, so the adverse effect is specific to the translation component of the sim-to-real transfer.

Three properties of this split explain the effect.
First, the supervision is uniquely scarce and uniform: the simulated training set spans only $22$ environments captured with a single physical rig geometry, whose two forward-facing cameras form a narrow $7.26$\,cm stereo baseline and whose two remaining cameras point sideways near $70^\circ$ off the forward axis.
Second, the simulated rig is built from the calibration of the real rig, so the nominal intra-group extrinsics transfer faithfully from simulation to the physical platform.
Third, the noise augmentation already injects rig variety along every degree of freedom, and its $0.1$\,m translation standard deviation exceeds the $7.26$\,cm stereo baseline itself.

These properties point to a scale effect rather than an orientation effect.
The degradation from simulation to reality is far larger in translation than in rotation: the mean translation rises by roughly four times ($0.305$\,m to $1.206$\,m) while the mean rotation rises by roughly two times ($1.67^\circ$ to $3.46^\circ$), and the same asymmetry appears in the intra-group corrections of Table~\ref{tab:appA_robustness}.
Metric scale on this rig is anchored by the narrow front stereo baseline.
Perturbing the input extrinsics by more than that baseline trains the model to distrust the precise rig geometry and to rely instead on visual scale cues.
Within the simulation this regularization is beneficial, since it counteracts overfitting to the small training set, which is why the augmentation improves both components on the simulated split.
Under zero-shot transfer the faithful rig calibration would otherwise carry metric scale directly to the real platform.
A model that trusts the rig geometry then recovers translation more accurately, whereas the noise-trained model leans on visual cues that transfer less reliably across the domain gap.
Rotation does not depend on metric scale, so it continues to benefit from the added regularization.
The anomaly is therefore a property of zero-shot sim-to-real transfer under a single-rig, small-data regime, and not a defect of the augmentation, which is beneficial in every in-domain setting.

% -----------------------------------------------------------------------------
\subsection{Component Contribution Ranking}
\label{app:ablation:ranking}

We rank the three training design choices, the anchor embedding, the curriculum, and the noise augmentation, by the mean error increase observed when each is removed.
The window size is excluded as an inference-time setting.
A consistent ranking requires datasets whose error scales are comparable, so that no single setting dominates the average.
HM3D, TartanGround, and NCLT are three in-domain datasets of comparable scale, and Table~\ref{tab:appA_rank} averages over them.
The ZJH split is reported separately, because its training set is far smaller and its real split is evaluated as zero-shot transfer.
On that real split, removing any one component can even lower the translation error, which would distort a contribution average.
The ZJH-specific behavior is analyzed below.

\begin{table}[h]
  \vspace{-0.2cm}
  \centering
  \caption{\textbf{Component contribution, averaged over HM3D, TartanGround, and NCLT.}
  Each entry is the mean increase in error when the component is removed from the default, in both translation and rotation.
  Larger values indicate a larger contribution.}
  \label{tab:appA_rank}
  \vspace{0.3em}
  \small
  \begin{tabular}{lcc}
    \toprule
    Component removed & $\Delta t$ (m) & $\Delta r$ ($^\circ$) \\
    \midrule
    Anchor embedding        & $+0.070$ & $+0.32$ \\
    Three-phase curriculum  & $+0.027$ & $+0.23$ \\
    $\SE$ noise augmentation & $+0.026$ & $+0.19$ \\
    \bottomrule
  \end{tabular}
  \vspace{-0.2cm}
\end{table}

On these three datasets the anchor embedding is the most influential single choice on both axes, since fixing the global reference frame benefits every prediction.
The curriculum and the noise augmentation contribute comparable and smaller amounts.
The ranking changes once the ZJH simulated set is included: there, removing the noise augmentation raises the rotation error by $1.12^\circ$, against $0.19^\circ$ on the three larger datasets, so the augmentation becomes the dominant factor.
This reflects its role as a regularizer under data scarcity, since the ZJH training set is the smallest by an order of magnitude.
The translation behavior of the noise augmentation on the ZJH real split is the anomaly analyzed in Sec.~\ref{app:ablation:zjh}.
Across all settings the intra-group predictions $\{\pose{A_0}{A_i}\}$ remain far easier than the cross-group predictions, as Table~\ref{tab:appA_robustness} shows, so the design choices matter most for the cross-group estimation that is the core of the task.

% -----------------------------------------------------------------------------
\subsection{Robustness to Extrinsic Noise}
\label{app:ablation:noise}

The ablations above vary the training recipe while keeping the test input clean.
This subsection instead perturbs the test input itself, to measure how the model behaves when the intra-group extrinsics are imprecise, as they always are at deployment.
The noisy condition injects the same $\SE$ perturbation used during training, with rotation standard deviation $1.5^\circ$ and translation standard deviation $0.1$\,m, applied to both groups.

\paragraph{End-task robustness and intra-group denoising.}
Table~\ref{tab:appA_robustness} reports \ours{} under clean and noisy extrinsics on all five settings.
Two effects stand out.
The end-task inter-group error barely moves between the two conditions: the mean translation changes by at most $0.015$\,m and the mean rotation by at most $0.05^\circ$, both far below the injected $0.1$\,m and $1.5^\circ$.
The intra-group corrections give an even sharper test, that is, whether the model corrects the extrinsics it is given or merely echoes them.
A baseline that returned the input unchanged would carry the injected noise straight through, so its intra-group rotation error would be on the order of the input $1.5^\circ$.
The model instead re-estimates the within-group poses to a mean rotation error below $0.85^\circ$ on every dataset, well under that level.
It therefore treats the given extrinsics as a noisy measurement to be denoised, not a fixed quantity to reproduce.
This is the payoff of the augmentation introduced in Sec.~\ref{app:impl:noise}: by training against the deployment noise distribution, the model learns to recover clean geometry from imprecise inputs.

% Big robustness table: G2G under clean vs noisy intra-group extrinsics.
% Inter-group (end-task) columns: official numbers from paper_results/table1_trained_clean.md
%   and table2_trained_noisy.md (G2G 64mo). Intra-group columns: per-frame A-intra mean errors
%   from the corresponding eval_results_*_{clean,noisy}/overall_summary.txt of the released models.
% Input noise during the noisy condition: rotation std 1.5 deg, translation std 0.1 m (both groups).
\begin{table*}[t]
\centering
\caption{\textbf{\ours{} is robust to noisy intra-group extrinsics, and recovers clean intra-group geometry from noisy input.}
  Each dataset is evaluated under clean ground-truth extrinsics and under extrinsics corrupted with the training-time $\SE$ noise (rotation std $1.5^\circ$, translation std $0.1$\,m, applied to both groups).
  The left block reports the end-task inter-group pose error, following the protocol of Table~1 of the main paper.
  The right block reports the error of the intra-group corrections $\{\pose{A_0}{A_i}\}$, that is, how accurately the model re-estimates the within-group poses it was given as input.
  Two effects are visible.
  First, the end-task error changes by far less than the injected perturbation: at most $0.015$\,m in mean translation against a $0.1$\,m input noise.
  Second, the intra-group rotation error stays below $0.85^\circ$, well below the $1.5^\circ$ input noise, which indicates that the model treats the input extrinsics as a noisy measurement to be corrected rather than copied.}
\label{tab:appA_robustness}
\vspace{0.3em}
\small
\setlength{\tabcolsep}{4.5pt}
\begin{tabular}{llcccccc cc}
\toprule
& & \multicolumn{6}{c}{\textit{Inter-group pose (end task)}} & \multicolumn{2}{c}{\textit{Intra-group correction}} \\
\cmidrule(lr){3-8} \cmidrule(lr){9-10}
Dataset & Cond. & $t_\text{med}$ & $t_\text{mean}$ & $r_\text{med}$ & $r_\text{mean}$ & RRA@5$^\circ$ & RTA@5$^\circ$ & $r_\text{mean}$ & $t_\text{mean}$ \\
        &       & (m)$\downarrow$ & (m)$\downarrow$ & ($^\circ$)$\downarrow$ & ($^\circ$)$\downarrow$ & (\%)$\uparrow$ & (\%)$\uparrow$ & ($^\circ$)$\downarrow$ & (m)$\downarrow$ \\
\midrule
\multirow{2}{*}{HM3D}
 & clean & 0.103 & 0.155 & 1.35 & 1.72 & 97.7 & 81.4 & 0.57 & 0.043 \\
 & noisy & 0.108 & 0.158 & 1.38 & 1.69 & 98.4 & 81.1 & 0.51 & 0.050 \\
\midrule
\multirow{2}{*}{TartanGround}
 & clean & 0.176 & 0.526 & 1.22 & 5.39 & 85.7 & 72.3 & 0.66 & 0.065 \\
 & noisy & 0.193 & 0.541 & 1.25 & 5.35 & 85.6 & 70.6 & 0.75 & 0.085 \\
\midrule
\multirow{2}{*}{NCLT}
 & clean & 0.434 & 0.692 & 2.08 & 2.47 & 95.2 & 77.4 & 0.58 & 0.064 \\
 & noisy & 0.444 & 0.694 & 2.14 & 2.50 & 95.0 & 77.3 & 0.74 & 0.076 \\
\midrule
\multirow{2}{*}{ZJH {\scriptsize(sim)}}
 & clean & 0.186 & 0.305 & 1.33 & 1.67 & 96.5 & 78.2 & 0.52 & 0.039 \\
 & noisy & 0.186 & 0.311 & 1.33 & 1.70 & 96.5 & 78.3 & 0.56 & 0.041 \\
\midrule
\multirow{2}{*}{ZJH {\scriptsize(real)}}
 & clean & 0.894 & 1.206 & 1.64 & 3.46 & 95.0 & 77.9 & 0.81 & 0.109 \\
 & noisy & 0.885 & 1.193 & 1.68 & 3.51 & 94.9 & 78.3 & 0.84 & 0.108 \\
\bottomrule
\end{tabular}
\vspace{-0.4cm}
\end{table*}

\paragraph{Cross-method comparison.}
The methods that consume intra-group extrinsics differ sharply in how the same noise propagates, as Table~\ref{tab:appA_noise_compare} shows.
On HM3D the scale-free RTA@$5^\circ$ of LoMa falls from $76.8$ to $50.7$ and that of Reloc3R from $76.2$ to $64.4$, while \ours{} changes from $81.4$ to $81.1$.
The mean translation tells the same story: under the $0.1$\,m perturbation LoMa and Reloc3R rise by more than $0.1$\,m on HM3D, exceeding the injected magnitude, whereas \ours{} rises by $0.003$\,m.
The gap follows from where the extrinsics enter each method.
LoMa and Reloc3R consume them only at the post-inference geometric stage, where noise corrupts ray intersection or extrinsics aggregation directly.
\ours{} injects the extrinsics into the frozen backbone before the trainable modules, so the perturbation is absorbed by the same representation that the model was trained to denoise.

% Cross-method sensitivity to extrinsic noise.
% Numbers from paper_results/table1_trained_clean.md (clean) and table2_trained_noisy.md (noisy),
% restricted to the methods that consume intra-group extrinsics and have a noisy variant
% (G2G, Reloc3R, LoMa). Same injected noise as Table~\ref{tab:appA_robustness}.
\begin{table}[t]
\centering
\caption{\textbf{Sensitivity to extrinsic noise across methods that consume intra-group extrinsics.}
  Clean and noisy use the same evaluation pairs; the noisy condition injects the perturbation of Table~\ref{tab:appA_robustness}.
  \ours{} conditions the frozen backbone on the extrinsics before its trainable modules, so the same representation trained to denoise absorbs the perturbation.
  LoMa and Reloc3R instead consume the extrinsics only at the post-inference geometric stage, where the noise propagates directly into ray intersection or extrinsics aggregation.
  The scale-free RTA@5$^\circ$ makes the contrast clearest: under noise it drops by at most $1.7$ points for \ours{}, against up to $26$ points for the post-hoc methods.}
\label{tab:appA_noise_compare}
\vspace{0.3em}
\small
\setlength{\tabcolsep}{5pt}
\begin{tabular}{llcccccc}
\toprule
& & \multicolumn{2}{c}{$t_\text{mean}$ (m)$\downarrow$} & \multicolumn{2}{c}{$r_\text{mean}$ ($^\circ$)$\downarrow$} & \multicolumn{2}{c}{RTA@5$^\circ$ (\%)$\uparrow$} \\
\cmidrule(lr){3-4} \cmidrule(lr){5-6} \cmidrule(lr){7-8}
Dataset & Method & clean & noisy & clean & noisy & clean & noisy \\
\midrule
\multirow{3}{*}{HM3D}
 & \ours{}  & 0.155 & 0.158 & 1.72 & 1.69 & 81.4 & 81.1 \\
 & Reloc3R  & 0.725 & 0.892 & 2.66 & 2.80 & 76.2 & 64.4 \\
 & LoMa     & 0.802 & 0.968 & 3.79 & 4.20 & 76.8 & 50.7 \\
\midrule
\multirow{3}{*}{TartanGround}
 & \ours{}  & 0.526 & 0.541 & 5.39 & 5.35 & 72.3 & 70.6 \\
 & Reloc3R  & 1.649 & 1.723 & 14.88 & 14.99 & 54.3 & 48.7 \\
 & LoMa     & 1.183 & 1.421 & 18.40 & 18.81 & 73.9 & 54.5 \\
\midrule
\multirow{3}{*}{NCLT}
 & \ours{}  & 0.692 & 0.694 & 2.47 & 2.50 & 77.4 & 77.3 \\
 & Reloc3R  & 6.918 & 7.071 & 6.34 & 6.38 & 36.5 & 32.8 \\
 & LoMa     & 7.858 & 7.827 & 22.13 & 22.20 & 20.9 & 17.8 \\
\bottomrule
\end{tabular}
\vspace{-0.4cm}
\end{table}

% -----------------------------------------------------------------------------
\subsection{Efficiency and Latent Count Trade-off}
\label{app:ablation:efficiency}

The freeze-and-bridge design confers practical advantages beyond accuracy.
Because the $539$M backbone parameters store no gradients or optimizer states, peak training memory is far below that of a fully trainable model of comparable size.
At inference, \ours{} produces all target poses in a single forward pass, whereas pairwise methods require $N_A \!\times\! N_B$ independent pair inferences.
Table~\ref{tab:appA_efficiency} quantifies both the inference advantage over the baselines and the training cost of the latent count $L$.

The resampler compresses each frame from $P{=}256$ patch tokens to $L{=}64$ latent tokens before the bridge.
Increasing $L$ from $64$ to $256$, by removing the resampler, yields only a marginal accuracy gain on HM3D ($0.123$\,m versus $0.155$\,m) but lengthens the merged bridge sequence from $640$ to $2{,}560$ tokens.
This longer sequence raises both the training time and the training memory.
Under the same configuration, the training time at $L{=}256$ is about $1.9\times$ that of the $L{=}64$ default.
The training memory grows much more steeply, from $5.44$\,GB at $L{=}64$ to $9.67$\,GB at $L{=}128$ and $21.34$\,GB at $L{=}256$, because the bridge stores activations for the backward pass and self-attention memory scales with the square of the sequence length.

This memory is the binding constraint.
At batch size $16$, the $L{=}256$ variant already consumes nearly the full $24$\,GB of an RTX~4090 and leaves no headroom, whereas $L{=}64$ uses under a quarter of that budget.
Doubling the batch to $32$ makes the contrast decisive: in our controlled measurement the default $L{=}64$ needs $10.3$\,GB, which fits a $24$\,GB GPU with room to spare, while $L{=}256$ needs $39.7$\,GB, which exceeds it and requires a $48$\,GB accelerator.
The released $256$-latent model reflects this directly: it had to be trained at batch size $16$, half the batch size of the $64$-latent default, in order to fit.
At inference with batch size $1$ the frozen backbone dominates and the difference across latent counts is small, between $2.3$ and $2.7$\,GB, so the cost of a large latent count is paid almost entirely during training.
The default $L{=}64$ is therefore a deliberate trade-off that retains competitive accuracy while keeping both the training memory and the training time within practical limits on standard GPUs.

\begin{table}[t]
  \centering
  \caption{\textbf{Efficiency comparison.}
  Inference metrics (left) are measured with batch size $1$ and no gradients.
  The training-time column reports training time under the same configuration, relative to the $L{=}64$ default; the training VRAM column uses batch size $16$ with the AdamW optimizer and includes the backward pass and the optimizer update.
  All runs use a single NVIDIA RTX~4090 GPU ($24$\,GB) in \texttt{bfloat16} precision with $5{+}5$ frames at $224{\times}224$, reporting the median of $100$ iterations after $20$ warmup iterations.
  For pairwise methods (Reloc3R, LoMa), inference latency includes all $5{\times}5{=}25$ pair inferences and geometric aggregation.
  $\dagger$\,VGGT uses $518{\times}518$ input resolution.
  $\ddagger$\,LoMa uses the lighter \texttt{DeDoDe-B} detector.}
  \label{tab:appA_efficiency}
  \vspace{0.3em}
  \small
  \setlength{\tabcolsep}{4pt}
  \begin{tabular}{lcccccc}
    \toprule
    & \multicolumn{2}{c}{\textit{Inference (bs\,=\,1)}} && \multicolumn{2}{c}{\textit{Training (bs\,=\,16)}} & \\
    \cmidrule{2-3} \cmidrule{5-6}
    Method & Latency (ms) & VRAM (GB) && Train time ($\times$) & VRAM (GB) & Trainable (M) \\
    \midrule
    \multicolumn{7}{l}{\textit{\ours{} variants (varying latent count $L$)}} \\
    \quad $L{=}32$                       &  49.70 & 2.33 && $0.92\times$ &  4.18 & 32.09 \\
    \quad $L{=}64$ (default)    &  49.79 & 2.33 && $1.00\times$ & 5.44 & 32.11 \\
    \quad $L{=}128$                      &  50.16 & 2.43 && $1.10\times$ &  9.67 & 32.16 \\
    \quad $L{=}256$ (no resampler)       &  52.10 & 2.74 && $1.90\times$ & 21.34 & 17.88 \\
    \midrule
    \multicolumn{7}{l}{\textit{Baselines}} \\
    Rig3R                                &  54.47 & 2.92 && --- & --- & 88.65 \\
    MA-AB (oracle, frozen)               &  98.48 & 4.00 && --- & --- & 0 \\
    Reloc3R ($25{\times}$ pairwise)      & 461.55 & 1.63 && --- & --- & 426.52 \\
    VGGT$^\dagger$                       & 521.41 & 8.87 && --- & --- & 1125.29 \\
    LoMa$^\ddagger$ ($25{\times}$ pairwise) & 3517.84 & 4.03 && --- & --- & 11.88 \\
    \bottomrule
  \end{tabular}
  \vspace{-0.4cm} % 适度收紧表后间距，保留正文前的留白。
\end{table}

Table~\ref{tab:appA_efficiency} also benchmarks inference against the baselines.
\ours{} completes inference in under $50$\,ms, whereas the pairwise methods Reloc3R and LoMa require $462$\,ms and $3{,}518$\,ms respectively due to the $25$ independent pair inferences, and VGGT reaches $521$\,ms at its native resolution.
\ours{} avoids the $N_A\times N_B$ separate forward passes of pairwise formulations; Sec.~\ref{app:ablation:varwin} measures its inference cost as group size increases.

% 补充消融：数值沿用最终 rebuttal；按实验目的组织，明确覆盖范围及预算。
\subsection{Intrinsic Conditioning and Cross-Group Attention}
\label{app:ablation:conditioning}

We isolate two sources of information in cross-sequence relocalization: camera intrinsics supplied to the frozen backbone and cross-group exchange in the bridge.
Both variants are retrained on HM3D and NCLT with the default data, training schedule, and other hyperparameters.
Table~\ref{tab:additional_controls} reports mean rotation and translation errors under clean extrinsics.

\begin{table}[htbp]
\centering
\caption{\textbf{Controlled localization ablations.} Each retrained variant changes one factor. The default row is the model in Table~\ref{tab:task1}.}
\label{tab:additional_controls}
\begin{tabular}{lrrrr}
\toprule
& \multicolumn{2}{c}{HM3D} & \multicolumn{2}{c}{NCLT} \\
\cmidrule(lr){2-3}\cmidrule(lr){4-5}
Variant & $t$ (m)$\downarrow$ & $r$ ($^\circ$)$\downarrow$ & $t$ (m)$\downarrow$ & $r$ ($^\circ$)$\downarrow$ \\
\midrule
\ours{} (default) & 0.155 & 1.72 & 0.692 & 2.47 \\
w/o intrinsics & 0.176 & 1.88 & 0.939 & 2.73 \\
w/o cross-group bridge attention & 0.562 & 8.45 & 1.502 & 3.76 \\
\bottomrule
\end{tabular}
\vspace{-0.45cm} % 收紧本表与后续正文的间距。
\end{table}

\paragraph{Intrinsic conditioning.}
The no-intrinsics variant omits the ray-direction input from both groups while retaining their intra-group extrinsics.
Translation error increases from $0.155$ to $0.176$\,m on HM3D and from $0.692$ to $0.939$\,m on NCLT; rotation changes more modestly.
Both localization datasets include varying camera fields of view, so providing intrinsics resolves information that would otherwise need to be inferred from the images.
The variant remains more accurate than Reloc3R on both datasets in Table~\ref{tab:task1}.

\paragraph{Cross-group attention.}
A block-diagonal mask disables only cross-group attention in the bridge, retaining backbone conditioning, intra-group attention, the resampler, and the pose head, which still reads anchor and target tokens.
The increased rotation and translation errors support the role of cross-group exchange within the bridge in aligning the independently encoded groups.

\subsection{Variable Group Size and Computational Cost}
\label{app:ablation:varwin}

For each of HM3D and NCLT, we train a variable-window model with $N\in[2,9]$ frames per group and evaluate the same weights at multiple window sizes.
These models use two-thirds of the default training budget.
Table~\ref{tab:variable_window} reports mean errors for the inter-group anchor pose, $\pose{A_0}{B_0}$.
The $N=11$ setting extends beyond the training range.

\begin{table}[htbp]
\centering
\caption{\textbf{Variable-window localization and inference cost.} Accuracy uses one variable-window model per dataset. Timing includes the frozen backbone and trainable modules, measured from raw images on an RTX~4090.}
\label{tab:variable_window}
\small
\begin{tabular}{crrrrrr}
\toprule
& \multicolumn{2}{c}{HM3D} & \multicolumn{2}{c}{NCLT} & Latency & Peak memory \\
\cmidrule(lr){2-3}\cmidrule(lr){4-5}
$N$ & $t$ (m)$\downarrow$ & $r$ ($^\circ$)$\downarrow$ & $t$ (m)$\downarrow$ & $r$ ($^\circ$)$\downarrow$ & (ms) & (GB) \\
\midrule
3 & 0.254 & 2.86 & 0.833 & 3.03 & 41.8 & 2.3 \\
7 & 0.193 & 1.91 & 0.732 & 2.66 & 62.0 & 2.4 \\
11 & 0.207 & 1.95 & 0.717 & 2.44 & 89.6 & 2.5 \\
\bottomrule
\end{tabular}
\vspace{-0.25cm} % 适度收紧表后间距，保留正文前的留白。
\end{table}

Increasing the window from three to seven frames improves both errors on both datasets.
At eleven frames, HM3D changes little relative to seven frames and NCLT improves further, showing that the trained estimator remains usable beyond its training window range.
These results characterize the variable-window models, separately from the fixed-window ablation in Table~\ref{tab:ablation}.

For timing, we use batch size one, $224\times224$ images, and \texttt{bfloat16} inference, and report the median of $50$ iterations after $20$ warmup iterations.
The $N=11$ timing uses interpolation of the learned positional embeddings beyond the nine-frame training range.
The trainable modules take approximately $2$--$4$\,ms; most inference time is spent in the frozen backbone.
A single forward pass avoids enumerating $N^2$ cross-group image pairs, although attention within the model still depends on the number of input tokens.

\subsection{Alternative Frozen Backbones}
\label{app:ablation:backbones}

We replace MapAnything with Depth Anything~3 (DA3-Large)~\cite{lin2025depthanything3} and $\pi^3$X, the camera-prior extension of $\pi^3$~\cite{wang2025pi3}, and retrain the downstream modules on HM3D.
Each backbone remains frozen and independently encodes the two groups with their camera priors.
The resampler, bridge, pose-head design, and loss are retained, with the feature interface adapted to the backbone.
Both experiments use two-thirds of the default training budget.

\begin{table}[htbp]
\centering
\caption{\textbf{Backbone substitution on HM3D localization.} Mean errors under clean extrinsics. The default MapAnything result is included for reference; the alternative backbones use a shorter training budget.}
\label{tab:alternative_backbones}
\begin{tabular}{lccrr}
\toprule
Frozen backbone & Parameters & Relative training budget & $t$ (m)$\downarrow$ & $r$ ($^\circ$)$\downarrow$ \\
\midrule
MapAnything (default) & 539M & $1$ & 0.155 & 1.72 \\
DA3-Large & 411M & $2/3$ & 0.263 & 2.64 \\
$\pi^3$X & 1{,}360M & $2/3$ & 0.151 & 1.42 \\
\bottomrule
\end{tabular}
\vspace{-0.45cm} % 收紧本表与后续正文的间距。
\end{table}

Table~\ref{tab:alternative_backbones} confirms that the group-wise encoding and alignment design supports both alternative backbones.
Despite its shorter training budget, $\pi^3$X improves on MapAnything, while DA3-Large is less accurate.
The observed accuracy ranking matches backbone parameter counts, suggesting that the design can benefit from stronger pretrained representations.

% 长程里程计：沿用最终 rebuttal 定量表，不加入完整轨迹可视化。
\section{Long-Horizon Odometry}
\label{app:odometry}

We evaluate accumulated motion error by composing consecutive relative poses on two held-out NCLT traversals, 2012-02-19 and 2012-08-20 ($\approx$6\,km each).
This complements the local rig-pair evaluation (Sec.~\ref{sec:task2}); no method uses loop closure or global bundle adjustment.

\paragraph{Inputs and sampling.}
Rig methods use the same stride-eight pairs of five-camera observations.
The monocular methods process a single camera stream: CUT3R~\cite{wang2025cut3r} and LingBot-Map~\cite{chen2026lingbotmap} use stride three, and VGGT-SLAM~\cite{maggio2025vggtslam} uses stride five.
Rig3R is retrained on NCLT; CUT3R, Reloc3R, and VGGT are finetuned on NCLT, with VGGT using its cross-date rig checkpoint.
VGGT-SLAM and LingBot-Map use released weights without finetuning.

\paragraph{Metrics and scale.}
We report relative translation error (RTE, percent of segment length) and relative rotation error (RRE, degrees per $100$\,m) over $100$--$800$\,m segments, averaging the two sequence scores.
G2G, Reloc3R, Rig3R, and finetuned CUT3R use metric predictions.
VGGT recovers scale from group-$A$ geometry; the remaining methods use a fixed scale estimated from the first $10$\,m of ground-truth travel.
Rig extrinsics express predictions and ground truth in a common camera frame; VGGT-SLAM additionally uses a hand-eye rotation estimated from ground truth.
These long-horizon calibration conventions differ from VGGT's per-pair alignment in Table~\ref{tab:task2}.

\begin{table}[htbp]
\centering
\caption{\textbf{Long-horizon open-loop odometry on NCLT.} Mean segment errors over two held-out traversals. MA-A and MA-AB include the ground-truth-assisted bias correction described below; MA-AB additionally receives the true inter-group transform.}
\label{tab:long_odometry}
\small
\begin{tabular}{llrr}
\toprule
Method & Use of intra-group extrinsics & RTE (\%)$\downarrow$ & RRE ($^\circ$/100\,m)$\downarrow$ \\
\midrule
\multicolumn{4}{l}{\textit{Monocular input}} \\
VGGT-SLAM & No conditioning interface & 59.6 & 38.5 \\
LingBot-Map & No conditioning interface & 45.3 & 20.7 \\
CUT3R (finetuned) & No conditioning interface & 43.2 & 19.1 \\
\midrule
\multicolumn{4}{l}{\textit{Multi-camera rig input}} \\
Rig3R (retrained) & Forward pass: rig metadata & 48.8 & 24.3 \\
VGGT (finetuned) & Scale from group $A$ & 41.5 & 17.0 \\
Reloc3R (finetuned) & Pose aggregation & 24.3 & 11.0 \\
MA-A (bias-corrected) & Forward pass: group $A$ & 34.1 & 18.0 \\
MA-AB (oracle, bias-corrected) & Forward pass: GT-aligned groups & 9.2 & 4.6 \\
\ours{} & Forward pass: both local groups & \textbf{6.1} & \textbf{3.5} \\
\bottomrule
\end{tabular}
\end{table}

\paragraph{MapAnything diagnostic.}
For each of MA-A and MA-AB (Sec.~\ref{app:baselines}), we estimate one shared rotation correction from pooled prediction residuals of both evaluation sequences and apply it before composing poses.
This uses evaluation ground truth, so the corrected results are diagnostic rather than independently calibrated deployment results.
MA-AB's bias is approximately $0.6^\circ$, predominantly about the vertical axis; correction reduces its mean RTE from $45.4\%$ to $9.2\%$.
G2G receives no bias correction.

\paragraph{Results.}
G2G reaches $6.1\%$ RTE and $3.5^\circ/100$\,m RRE, compared with $24.3\%$ and $11.0^\circ/100$\,m for Reloc3R (Table~\ref{tab:long_odometry}).
The bias-corrected MA-AB result shows that providing the true inter-group transform does not guarantee the smallest accumulated error from a reconstruction model's predicted poses.
This comparison evaluates open-loop composition; full SLAM systems can further reduce drift through loop closure and global optimization.

\raggedbottom % 定性图按整块分页时，避免为齐底而拉伸标题与正文间距。
% =============================================================================
% 附录 H：保留四个数据集的全部案例，每页上下两张图，右侧放图注。
% =============================================================================
\section{Additional Qualitative Examples}
\label{app:qualitative}

Figures~\ref{fig:appG_hm3d}--\ref{fig:appG_zjh} show four cases per dataset; animated views are available on our \href{https://weiyufei0217.github.io/G2G/}{project page}.
Each case shows the ground-truth motion, the naive overlay ($A$: blue; $B$: orange), the G2G alignment with pose errors in red, the ground-truth alignment, and the $2{\times}5$ input frames ($A$ above $B$).
All panels use reconstructed point clouds and predicted poses.

\noindent\begin{minipage}{\linewidth}
  \begin{minipage}[c]{0.71\linewidth}
    \centering
    \includegraphics[width=\linewidth]{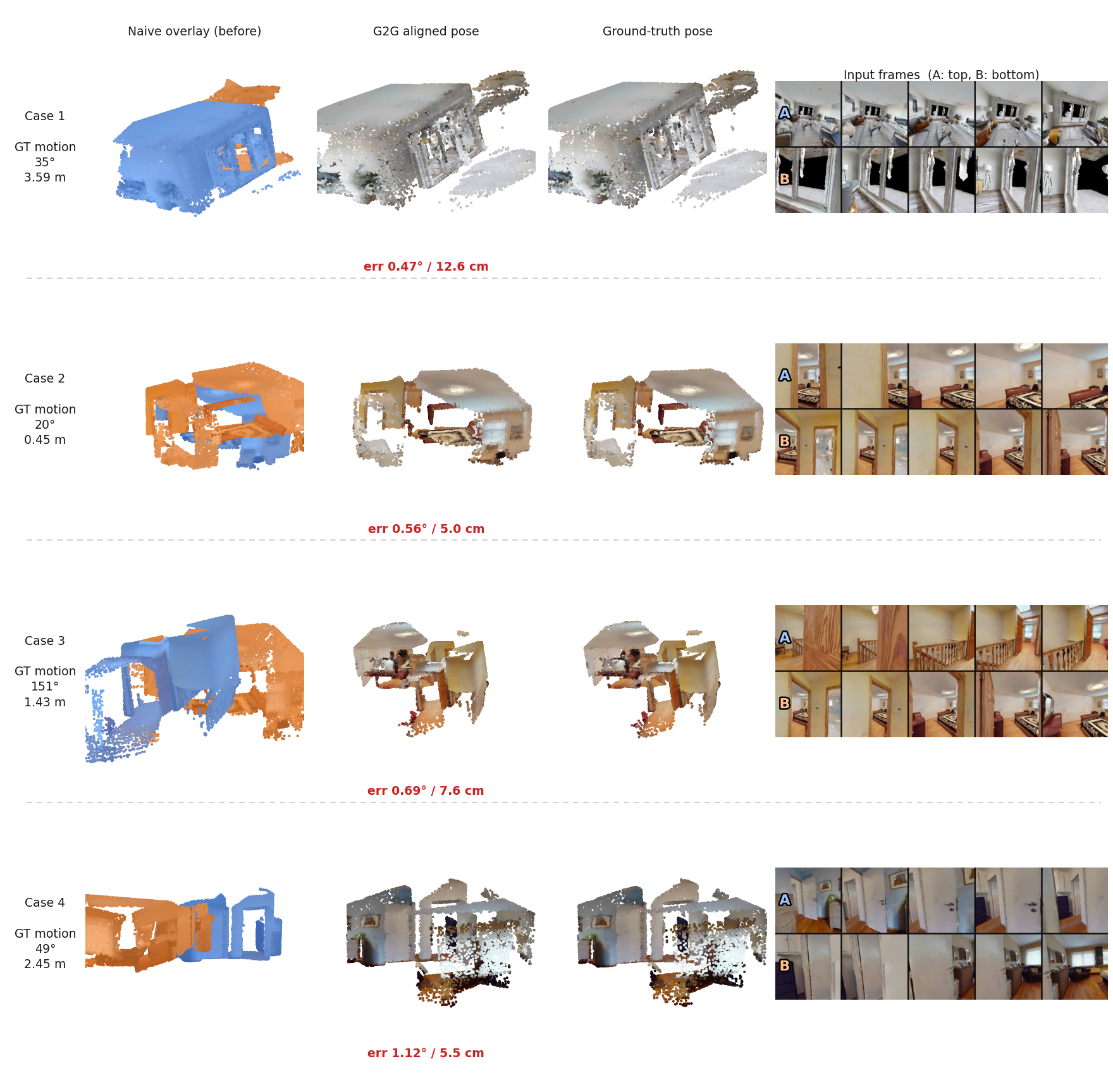}
  \end{minipage}\hfill
  \begin{minipage}[c]{0.27\linewidth}
    \captionsetup{type=figure,justification=raggedright,singlelinecheck=false}
    \caption{\textbf{HM3D relocalization (synthetic indoor, in-distribution).}
  Four cross-sequence cases spanning ground-truth rotations from $20^\circ$ to a near-opposite $151^\circ$.
  \ours{} recovers each pose to within $0.47^\circ$--$1.12^\circ$ and $3$--$13$\,cm.}
  \label{fig:appG_hm3d}
  \end{minipage}
\end{minipage}

\par\vspace{6pt}
\noindent\begin{minipage}{\linewidth}
  \begin{minipage}[c]{0.71\linewidth}
    \centering
    \includegraphics[width=\linewidth]{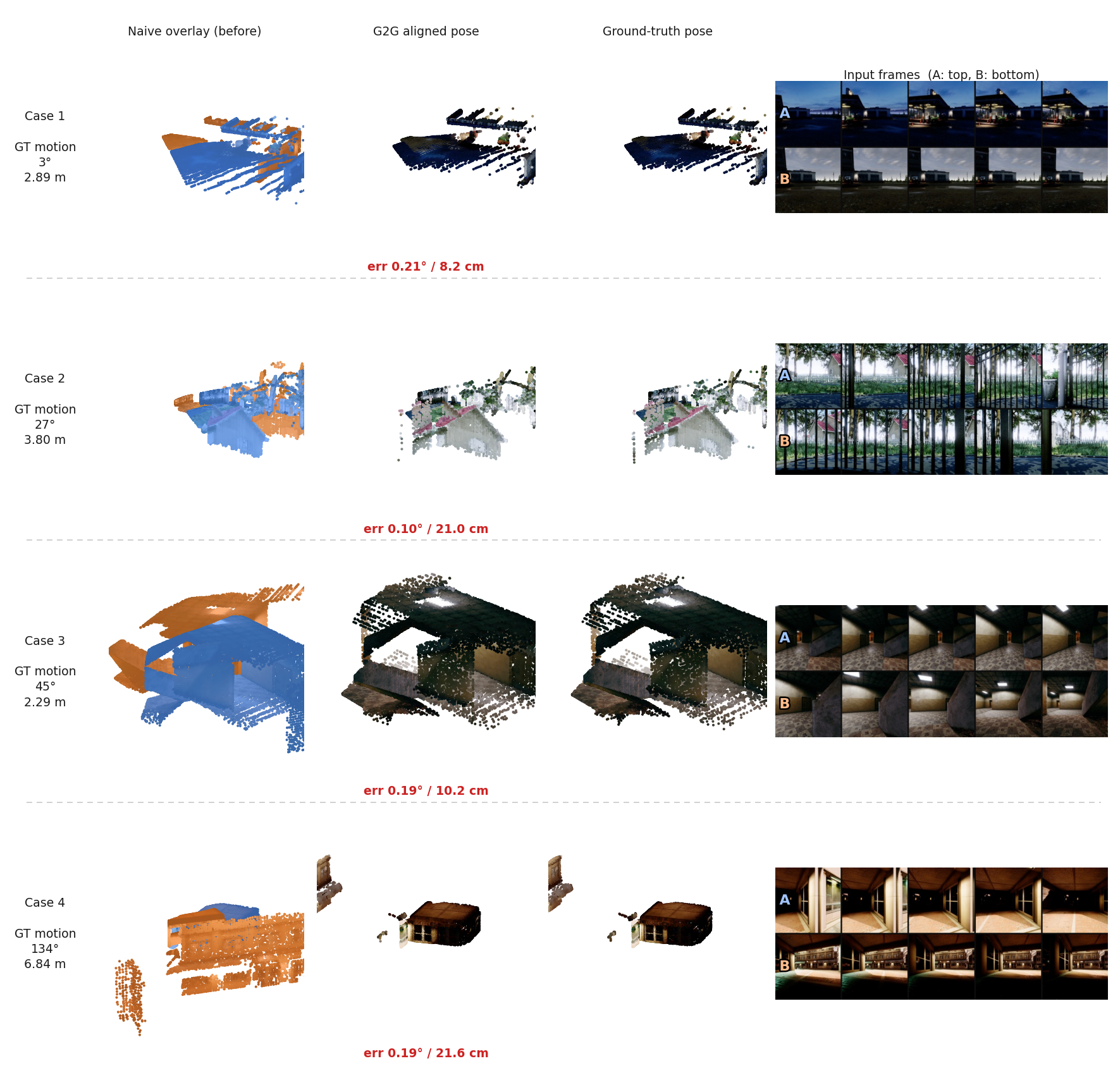}
  \end{minipage}\hfill
  \begin{minipage}[c]{0.27\linewidth}
    \captionsetup{type=figure,justification=raggedright,singlelinecheck=false}
    \caption{\textbf{TartanGround relocalization (synthetic, diverse environments).}
  Four cases spanning low-light outdoor, forest, and indoor scenes, with ground-truth rotations from $3^\circ$ to $134^\circ$.
  \ours{} attains errors of $0.10^\circ$--$0.21^\circ$ and $8$--$22$\,cm, remaining accurate despite the strong appearance variation across environments.}
  \label{fig:appG_tartanground}
  \end{minipage}
\end{minipage}

\clearpage
\noindent\begin{minipage}{\linewidth}
  \begin{minipage}[c]{0.71\linewidth}
    \centering
    \includegraphics[width=\linewidth]{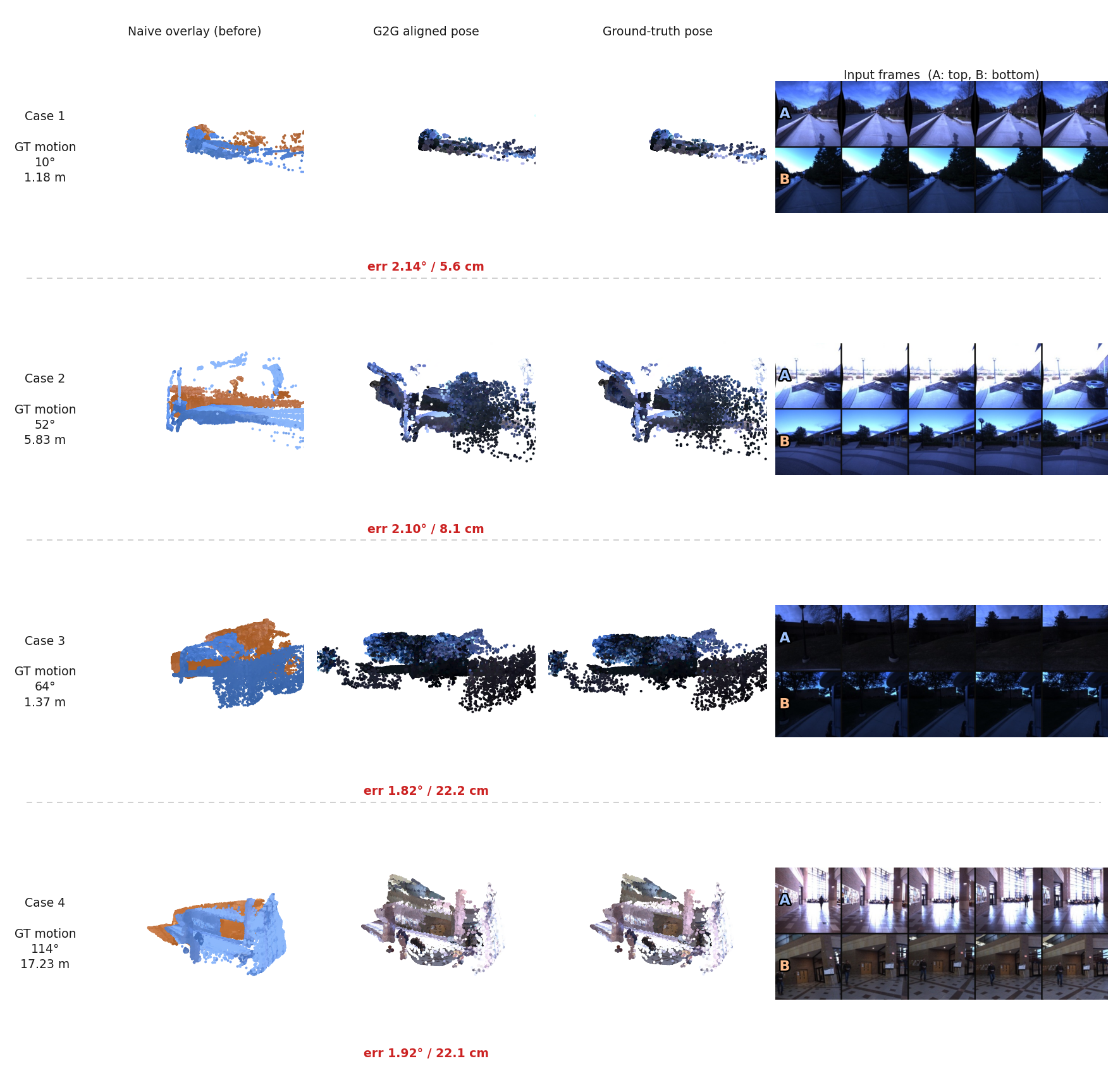}
  \end{minipage}\hfill
  \begin{minipage}[c]{0.27\linewidth}
    \captionsetup{type=figure,justification=raggedright,singlelinecheck=false}
    \caption{\textbf{NCLT cross-season relocalization (real).}
  Group~$A$ is captured in winter (February~19, 2012) and group~$B$ at the same campus location in summer (August~20, 2012), so visual appearance differs substantially.
  Over ground-truth rotations from $10^\circ$ to $114^\circ$ (including a $17.2$\,m baseline in the last case), \ours{} keeps the rotation error near $2^\circ$, relying on the geometry-conditioned features rather than visual similarity.}
  \label{fig:appG_nclt}
  \end{minipage}
\end{minipage}

\par\vspace{12pt}
\noindent\begin{minipage}{\linewidth}
  \begin{minipage}[c]{0.71\linewidth}
    \centering
    \includegraphics[width=\linewidth]{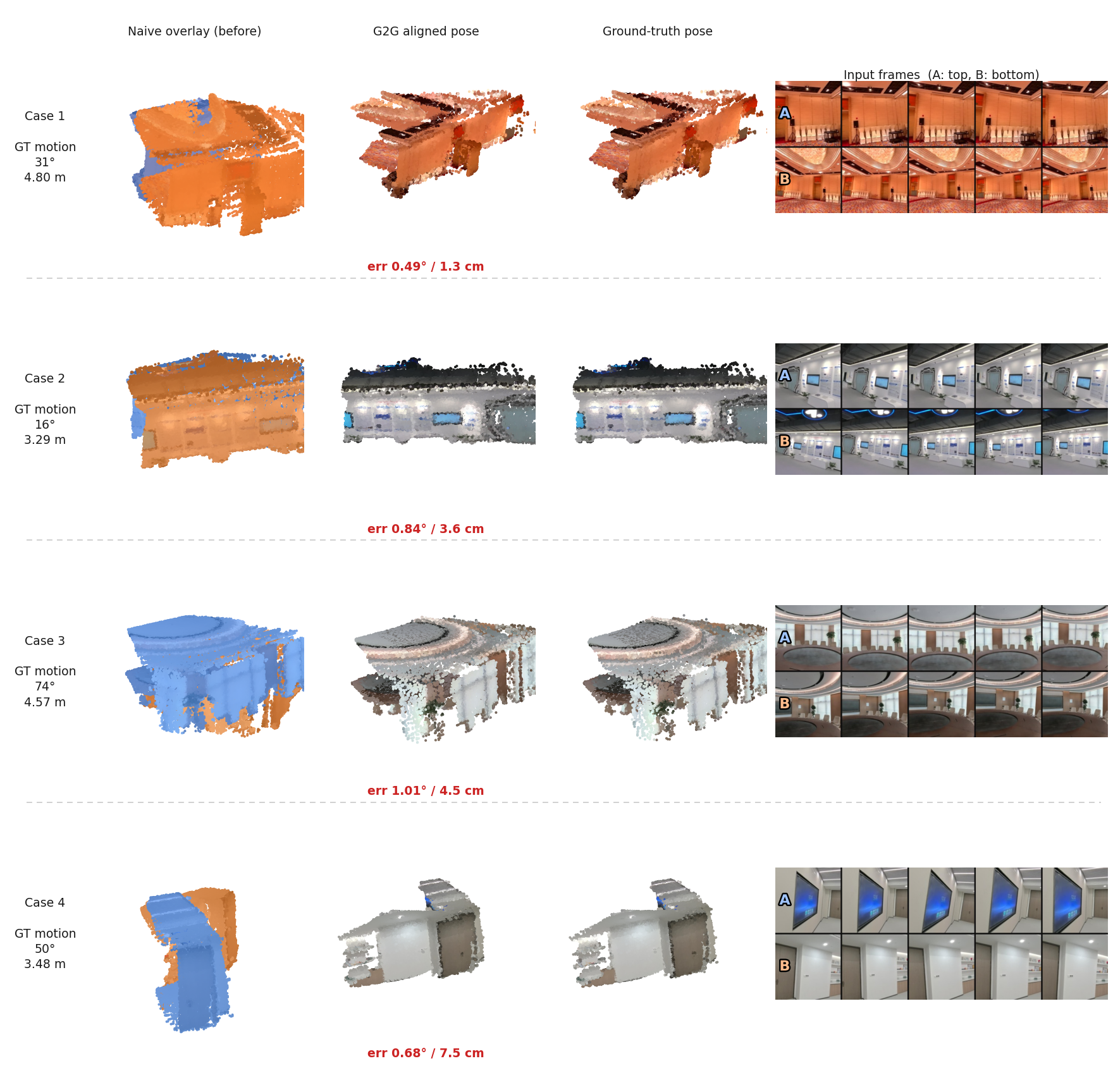}
  \end{minipage}\hfill
  \begin{minipage}[c]{0.27\linewidth}
    \captionsetup{type=figure,justification=raggedright,singlelinecheck=false}
    \caption{\textbf{ZJH sim-to-real relocalization (real).}
  Training images are sampled in a simulator from 3D Gaussian-Splatting scene reconstructions. The examples shown here are real captures, evaluated with the simulation-trained model without adaptation.
  Across ground-truth rotations from $16^\circ$ to $74^\circ$, \ours{} attains errors of $0.49^\circ$--$1.01^\circ$ and $1$--$8$\,cm, confirming transfer from simulation to real captures.}
  \label{fig:appG_zjh}
  \end{minipage}
\end{minipage}

\clearpage
% =============================================================================
% Appendix I: Author Contributions
% 状态: camera-ready 新增。首页脚注只保留一句共同一作说明并用 \ref 指向本节，
%       完整贡献声明放在这里（正文原文一字不改）。
% =============================================================================
\section{Author Contributions}
\label{app:contributions}

Yufei Wei conceived the Group-to-Group formulation and led the project. He designed the full G2G architecture, implemented the training and evaluation codebase, built the data pipeline for all datasets, developed the covisibility-based window-selection module, and designed and conducted all experiments, including the retraining of the learned baselines, the ablation and robustness studies, and the additional control and long-horizon experiments. He wrote the manuscript and the supplementary material. Shuhao Ye ran the LoMa baseline, processed the COLMAP ground truth for the real-world ZJH sequences, and assisted in preparing the architecture figure. Chenxiao Hu implemented the Habitat simulator interface for HM3D. Yiyuan Pan collected the real-world ZJH data, and Dongyu Feng collected the simulated ZJH data. Rong Xiong, Yue Wang, and Yanmei Jiao supervised the project and revised the manuscript.

\end{document}